\documentclass[review,authoryear]{elsarticle}
\usepackage[top=1in, bottom=1in, left=1.25in, right=1.25in]{geometry}
\usepackage{times}
\usepackage{amssymb}
\usepackage{graphicx}
\usepackage{amsmath}
\usepackage{mathrsfs}
\usepackage{epstopdf}
\usepackage{multirow}
\usepackage{subcaption} 
\usepackage{kotex}
\usepackage{tikz}
\usetikzlibrary{matrix}
\usetikzlibrary{positioning}
\usepackage{color}
\usepackage{cite}
\usepackage{pifont}
\usepackage{url}
\usepackage{caption}
\usepackage{sidecap}
\usepackage{rotating}
\usepackage{chngcntr}
\usepackage[flushleft]{threeparttable}
\newcommand{\ws} {\color{black}}

\usepackage{hyperref}
\usepackage{booktabs}
\usetikzlibrary{arrows,decorations.pathmorphing,decorations.markings,backgrounds,positioning,fit,plotmarks}
\usepackage{pgfplots}

\newcommand{\etal}{{\em et al.}}       
\newcommand{\etc}{{\em etc.}}         

\newcommand{\bA}{{\bf A}}
\newcommand{\bff}{{\bf f}}

\newcommand{\bi}{{\bf i}}

\newcommand{\bW}{{\bf W}}
\newcommand{\bc}{{\bf c}}

\newcommand{\bx}{{\bf x}}
\newcommand{\bX}{{\bf X}}
\newcommand{\bu}{{\bf u}}
\newcommand{\by}{{\bf y}}
\newcommand{\bY}{{\bf Y}}

\newcommand{\bo}{{\bf o}}
\newcommand{\bh}{{\bf h}}
\newcommand{\ba}{{\bf a}}
\newcommand{\bb}{{\bf b}}
\newcommand{\bz}{{\bf z}}
\newcommand{\bs}{{\bf s}}

\newcommand{\bM}{{\bf M}}
\newcommand{\bmm}{{\bf m}}

\newcommand{\bdelta}{{\boldsymbol{\delta}}}
\newcommand{\bDelta}{{\boldsymbol{\Delta}}}
\newcommand{\bgamma}{{\boldsymbol{\gamma}}}
\newcommand{\bbeta}{{\boldsymbol{\beta}}}
\newcommand{\ie}{\textit{i.e.}}
\newcommand{\eg}{\textit{e.g.}}

\definecolor{red}{RGB}{255,0,0}
\definecolor{blue}{RGB}{0, 0, 255}
\definecolor{ballblue}{rgb}{0.13, 0.67, 0.8}

\newcommand*{\rom}[1]{\expandafter\@slowromancap\romannumeral #1@}

\newcommand{\eat}[1]{}
\usepackage{url}

\let\today\relax
\makeatletter
\def\ps@pprintTitle{%
    \let\@oddhead\@empty
    \let\@evenhead\@empty
    \def\@oddfoot{\footnotesize\itshape
         {Under Review} \hfill\today}%
    \let\@evenfoot\@oddfoot
    }
\makeatother

\begin{document}

\begin{frontmatter}



\title{Deep Recurrent Model for Individualized Prediction of\\ Alzheimer's Disease Progression}

\author[ku1]{Wonsik Jung}
\author[ku1]{Eunji Jun}
\author[ku1,ku2]{Heung-Il Suk\corref{cor1}}

\author{\\and the Alzheimer's Disease Neuroimaging Initiative\fnref{dataset}}

\address[ku1]{Department of Brain and Cognitive Engineering, Korea University, Seoul 02841, Republic of Korea}
\address[ku2]{Department of Artificial Intelligence, Korea University, Seoul 02841, Republic of Korea}
\cortext[cor1]{Corresponding author: Heung-Il Suk (hisuk@korea.ac.kr)}

\fntext[dataset]{Data used in preparation of this article were obtained from the Alzheimer's Disease Neuroimaging Initiative (ADNI) database (\url{http://www.loni.ucla.edu/ADNI}). As such, the investigators within the ADNI contributed to the design and implementation of ADNI and/or provided data but did not participate in analysis or writing of this report. A complete listing of ADNI investigators can be found at \url{http://adni.loni.ucla.edu/wpcontent/uploads/how_to_apply/ADNI_Authorship_List.pdf}.}

\begin{abstract} 
Alzheimer's disease (AD) is known as one of the major causes of dementia and is characterized by slow progression over several years, with no treatments or available medicines. In this regard, there have been efforts to identify the risk of developing AD in its earliest time. While many of the previous works considered cross-sectional analysis, more recent studies have focused on the diagnosis and prognosis of AD with longitudinal or time series data in a way of disease progression modeling (DPM). Under the same problem settings, in this work, we propose a novel computational framework that can predict the phenotypic measurements of MRI biomarkers and trajectories of clinical status along with cognitive scores at multiple future time points.
However, in handling time series data, it generally faces with many unexpected missing observations. In regard to such an unfavorable situation, we define a secondary problem of estimating those missing values and tackle it in a systematic way by taking account of temporal and multivariate relations inherent in time series data. Concretely, we propose a deep recurrent network that jointly tackles the four problems of (i) missing value imputation, (ii) phenotypic measurements forecasting, {(iii) trajectory estimation of cognitive score,} and (iv) clinical status prediction of a subject based on his/her longitudinal imaging biomarkers. Notably, the learnable model parameters of our network are trained in an end-to-end manner with our circumspectly defined loss function. In our experiments over The Alzheimer’s Disease Prediction Of Longitudinal Evolution (TADPOLE) challenge cohort, we measured performance for various metrics and compared our method to competing methods in the literature. Exhaustive analyses and ablation studies were also conducted to better confirm the effectiveness of our method.

\end{abstract}


\end{frontmatter}

\section{Introduction}
\label{sec:introduction}
Alzheimer's disease (AD) is a catastrophic and progressive neurodegenerative disease that is clinically characterized by the impairment of cognitive and functional abilities along with behavioral symptoms. It is known that AD is the most common cause of dementia, accounting for about $70\%$ of age-related dementia~\citep{alzheimer20192019}. It is estimated that there are approximately $50$ million AD patients around the world, and the figure will double by 2030~\citep{patterson2018world}, thus causing great social and economic burdens~\citep{brookmeyer2007forecasting, alzheimer20192019}. Unfortunately, there is currently no pharmaceutical or clinical treatment available to reverse or cure the progression of AD~\citep{marinescu2018tadpole,gaugler20192019}.
When considering the characteristics of patients with AD, the clinical symptoms that disturb the activities of daily living, such as memory loss, language impediment, and other malfunctions, become apparent only after several years from the time that the brain is initially affected by the disease~\citep{braak1996development, morris1996cerebral}. For this reason, it is of great importance to identify the emergence of the disease at its earliest stage or predict the risk of emergence as early as possible.

In recent decades, leveraging the advances in machine learning, especially deep learning, there have been efforts to formulate the task of neuroimaging-based early AD diagnosis or prognosis as regression and classification problems \citep{zhang2011multimodal,suk2014hierarchical,suk2016deep,liu2018joint}. On the one hand, many of the existing studies focused on cross-sectional data analysis for AD diagnosis {\citep{suk2013deep,cheng2017classification,liu2018classification}}. These studies showed reasonable performance in the clinical status prediction task, even though only baseline features were used. However, they have limitations that did not properly model the progressive deterioration of the disease and different occurrence time for each individual, which are characteristics of AD.

On the other hand, in order to better utilize the available historical data, \ie, longitudinal data, disease progression modeling (DPM) has drawn researchers' attention \citep{sukkar2012disease,peterson2017personalized,lee2019predicting,lorenzi2019probabilistic,ghazi2019training}, especially for prediction of the time-to-conversion to the next clinical status in the AD spectrum.
The AD progression generally involves not only the acceleration of regional volume atrophy (\eg, the hippocampus, which is presumed to show the earliest morphological changes due to AD) but also increased enlargement of the ventricle over time \citep{nestor2008ventricular}. These types of changes can be better captured from longitudinal data, which allows for learning the underlying temporal characteristics in the disease. {\ws By discovering such characteristics, a DPM may be possible to forecast the future changes in phenotypic measurements and predict the time of conversion in clinical statuses that are symptomatically categorized as cognitively normal (CN), mild cognitive impairment (MCI), and dementia \citep{mills2014methods}. In particular, a better understanding of the individual disease progression contributes to more accurate diagnosis, monitoring, and prognosis, which are respectively beneficial for early diagnosis, intervention, and personalized care \citep{oxtoby2017imaging}. In this work, we focus on neuroimaging-based DPM for AD prognosis and time-to-conversion prediction in the AD spectrum.}

For the DPM, conversion time prediction is critical, as it gives us meaningful information about the disease progression and the severity of the disease. It is much more challenging than just predicting whether the patient will progress to AD. Conversion time prediction in this study is analogous to survival analysis~\citep{miller2011survival,oulhaj2009predicting,thung2018conversion}. Survival analysis is the study of time-to-event data, modeling the expected period until event occurrence (\eg, disease status conversion) at a future time. Although conversion time prediction resembles survival analysis, they actually tackle the problems with different approaches. First, survival analysis focuses on predicting the probability of AD conversion at different future time points. However, conversion time prediction forecasts when the conversion of the disease will occur. 
For example, in general, probability methods (\eg, Cox regression model) are used to access the survival problems, while conversion time prediction is based on typical regression methods (\eg, least-squares regression model)~\citep{thung2018conversion}. 
In this work, we investigates both approaches in devising the DPM for AD progression.

In the literature on survival analysis and related fields, DPM has been widely applied when the causes of diseases were not well-known or various factors were involved. Although working with longitudinal data is useful for improving our knowledge of the disease, adding a temporal dimension entails different forms of difficulties in data analysis, such as increasing dimensionality, missing values, and time alignment problems \citep{ibrahim2009missing}.
In a longitudinal dataset, there could be samples of subjects with one or two follow-ups (`short-term') and those of subjects with more frequent follow-ups (`long-term') after the first visit. Both scenarios have it owns issues to be tackled. With short-term data, we might not suffer from missing measurements and data imbalance problems, but we are restricted from learning patterns of long-range dependence by nature, thus limited in building a reliable model for DPM \citep{ardekani2017prediction,fiot2014longitudinal,gray2012multi,shi2017nonlinear}. Meanwhile, long-term longitudinal data are potentially better suited for DPM by allowing us to get better insights into the global patterns in disease progression~\citep{bilgel2015temporal,guerrero2016instantiated,bilgel2016multivariate,aghili2018predictive}, but generally have issues of many missing measurements and time alignment, due to research or patients drop out or unexpected accidents, \etc~\citep{petersen2010alzheimer}.

Some previous studies~\citep{wei2016prediction,zhang2012predicting,misra2009baseline} have been developed for MCI-to-AD conversion, which identified patients who will progress to AD. \citep{wei2016prediction} used MRI thickness measures as features to predict MCI-to-AD conversion. They used data at the baseline and up to 18 months (short-term samples). Meanwhile, \citep{misra2009baseline} used longitudinal MRI data to extract the changes of brain atrophy for detecting MCI-to-AD conversion. However, this study employed a short period follow-up data (up to 15 months). In addition, Zhang \etal\ also utilized longer period of longitudinal MRI data up to 24 months to predict cognitive scores as well as the conversion of MCI patients by longitudinal feature selection method that extracts the brain regions across multiple time points~\citep{zhang2012predicting}.

More recently, we have witnessed the potential of deep learning methods \citep{wang2018predictive,lee2019predicting,ghazi2019training,jung2019unified} for DPM thanks to their favorable characteristics of learning feature representations from data, rather than engineering feature manually. Especially, recurrent neural networks (RNNs) and their variants have been widely used because of their methodological ability to handle the time alignment issue among intra- and inter-subject trajectories. In particular, long short-term memory (LSTM)~\citep{hochreiter1997long} is one of the successful techniques to encode temporal patterns   \citep{wang2018predictive,yoon2018estimating,ghazi2019training,jung2019unified} by capturing long-term dependencies among multivariate measurements over time with efficient gating-based operations \citep{gers1999learning,gers2001lstm}.
However, its use in DPM remains challenging due to the incomplete samples, \ie, missing measurements in the data. For that reason, the previous works~\citep{lipton2016rnn,wang2018predictive,ghazi2019training,yoon2018estimating} have mostly involved the step of missing value imputation either as a means of preprocessing prior to training RNNs \citep{lipton2016rnn,wang2018predictive} or by regarding it as a sub-task of the target model training for prediction \citep{yoon2018estimating}.

More insight for these approach, Lipton \etal~exploited RNNs with LSTM cells to diagnose the pediatric intensive care unit (PICU) from clinical time series data~\citep{lipton2016rnn}. Wang \etal~also used similar RNNs variant to predict future cognitive scores given 11 yearly observations~\citep{wang2018predictive}. When training their method, all missing values of the patients' historical observations due to varying number of visits and uneven time intervals, were first imputed by simple mean or forward imputation methods~\citep{lipton2016rnn,wang2018predictive}, disregarding both the temporal relations and multivariate relations of observations. Meanwhile, Ghazi \etal~formulated the task of AD progression modeling by means of predicting changes in volumetric MRI biomarkers in a vanilla LSTM with peephole connections~\citep{gers2000} added to all the gates~\citep{ghazi2019training}. In contrast, all the missing values of longitudinal samples were initialized with zeros, and no further imputation update was conducted. Here, we named this model as PLSTM-Z. It is also noteworthy that for the clinical status prognosis, they built a separate classifier with linear discriminant analysis (LDA) by taking the predicted biomarker values from the trained PLSTM-Z as input features.

Since the performance of predictive models with preprocessing-based missing value imputation are highly dependent on the static values, \eg, zeros and means, determined manually or statistically, some researchers have devoted their efforts to constructing adaptive and dynamic imputation models. For example, Yoon \etal~focused on the task of missing value imputation and utilized a bi-directional RNN to better take into account temporal relations, but no relations among multivariate features, for mortality prediction with electronic health records~\citep{yoon2018estimating}. More recently, Cao \etal~gave special attention to the task of missing values imputation in time series data~\citep{cao2018brits}. In particular, given a sequence of observations in a fixed time interval, they designed the LSTM model, the one in the original time order and the other in the reversed time order, to maximally utilize the inherent temporal relations along with multivariate relations. However, due to the use of a time-reversed sequence, their imputation method is not applicable for disease progression modeling, \ie, long-term prognosis utilize current and more recently available data, but not future data.

Inspired by the recent work of dynamic imputation methods described above, in this work, we hypothesize that both multivariate and temporal relations inherent in longitudinal data could be informative to estimate missing values in devising DPM for AD. It is also believed that the appropriately imputed values help enhance the predictive power of a model for downstream tasks, \ie, in our work, volumetric changes of MRI biomarkers, trajectories of cognitive score, and clinical status prognosis over multiple time points ahead, thus making it possible to infer time-to-conversion in the AD spectrum.
In these regards, we propose a unified framework that can (i) adaptively impute missing values (`\emph{imputation module}'), (ii) encode imputed observations into latent representations via LSTM (`\emph{encoding module}'), and (iii) {\ws predict future MRI volumetric measurements, trajectories of cognitive score,} and clinical statuses (`\emph{prediction module}') over time in terms of the AD progression from a subject's historical measurements. Methodologically, we devise an imputation architecture that takes advantage of the encoded latent representations from an LSTM cell, \ie, temporal relations, and utilizes multivariate relations among variables to infer missing values. For the temporal encoding in LSTM-Z (LSTM with zero imputation), we feed the \emph{complete} observation, \ie, replace the missing values with imputed ones, a mask vector that indicates which values are imputed, and the information of time-delay from the last observed times. All of these greatly help to better encode and capture temporal characteristics in a longitudinal sample for modeling disease progression. Further, we also design an objective function such that the network parameters of all the three modules can be trained jointly in an end-to-end manner.

We evaluated the performance of our proposed method using the longitudinal cohorts from `The Alzheimer’s Disease Prediction Of Longitudinal Evolution (TADPOLE)' Challenge\footnote{\url{https://tadpole.grand-challenge.org/}} in terms of (i) future volumetric biomarker regression, {\ws(ii) cognitive score regression,} and (iii) clinical status classification, respectively.
Furthermore, we conducted exhaustive analyses to verify the effectiveness of the proposed framework for AD progression in three aspects, namely, model perspective analysis, longitudinal stability analysis, and personalized analysis.

The main contributions of our work can be summarized as follows:
\begin{itemize}
    \item To the best of our knowledge, our work is the first to use a novel deep recurrent network that jointly performs missing value imputation, {\ws forecasting of future MRI biomarker and cognitive score,} and clinical status prognosis over multiple time points in a unified framework.
    \item A multi-objective function for the proposed imputation-encoding-prediction network is devised and optimized in an end-to-end manner.
    \item Compared to the competing methods considered in our experiments, our proposed method outperformed them in all metrics for imputation, regression, and classification.
\end{itemize}

This work extends the preliminary version published by~\citep{jung2019unified} by revising the network architecture and performing further extensive experiments and analyses. {\ws Specifically, compared to the earlier work, the following extension and improvements have been made: first, we utilized the cognitive scores of Mini-Mental State Examination (MMSE) and Alzheimer's Disease Assessment Scale-Cognitive Subscale (ADAS-cog11 and ADAS-cog13) as input to the model in addition to volumetric MRI features, and predicted the cognitive scores along with the clinical labels over time. Moreover, we performed detailed analyses of the experimental results and the model perspective.} 
\section{Materials and Preprocessing}
\label{sec:materials}
In this work, we used the TADPOLE longitudinal cohort\footnote{\href{https://tadpole.grand-challenge.org/Data/}{Available at `\url{https://tadpole.grand-challenge.org/Data/}'}} from the ADNI database{,} including ADNI-1, ADNI-2, and ADNI-GO. The main objective of the challenge was to track the progression of subjects with AD using MRI biomarkers, {\ws cognitive test scores,} and clinical measurements. The dataset includes $1,500$ kinds of longitudinal MRI biomarkers, \eg, cortical thickness {and} cortical volume, from $1,737$ patients (aged from $54.40$ up to $91.40$) with $12,741$ visits at $22$ different time points between 2003 and 2017 \citep{marinescu2018tadpole}. 
The summary of the TADPOLE dataset is presented in Table \ref{appendix:tab_demographic} in the appendix A. 

As our main objective in this work is to predict the yearly progression, we considered 11 regular visits out of 22 total visits. In the TADPOLE dataset, the clinical groups were labeled as 342 AD, 417 CN, 310 EMCI (early MCI), 562 LMCI (late MCI), and 106 SMC (significant memory concern). Like the previous work~\citep{ghazi2019training,jung2019unified}, we merged the groups of CN and SMC into CN, and EMCI and LMCI into MCI, thus resulting in three categories: CN, MCI, and AD. 
Moreover, from the original TADPOLE dataset, we excluded subjects with no visit on the baseline or having less than 3 visits, resulting in a total of 691 subjects in the end. The number of longitudinal measurements ranges from 4 to 11 with $5.37\text{(mean)}\pm2.46\text{(std)}$. The duration of sequences ranges from 36 to 120 with $64.40\text{(mean)}\pm29.48\text{(std)}$. The detailed information of demographic and cognitive scores for samples used in our work is summarized in Table~\ref{append:demograpic_info} over 11 yearly time points (see appendix C). Note that there are many missing observations in the MRI biomarkers and clinical labels over the selected dataset.

\begin{figure}[!t]
\centering
\includegraphics[width=\textwidth]{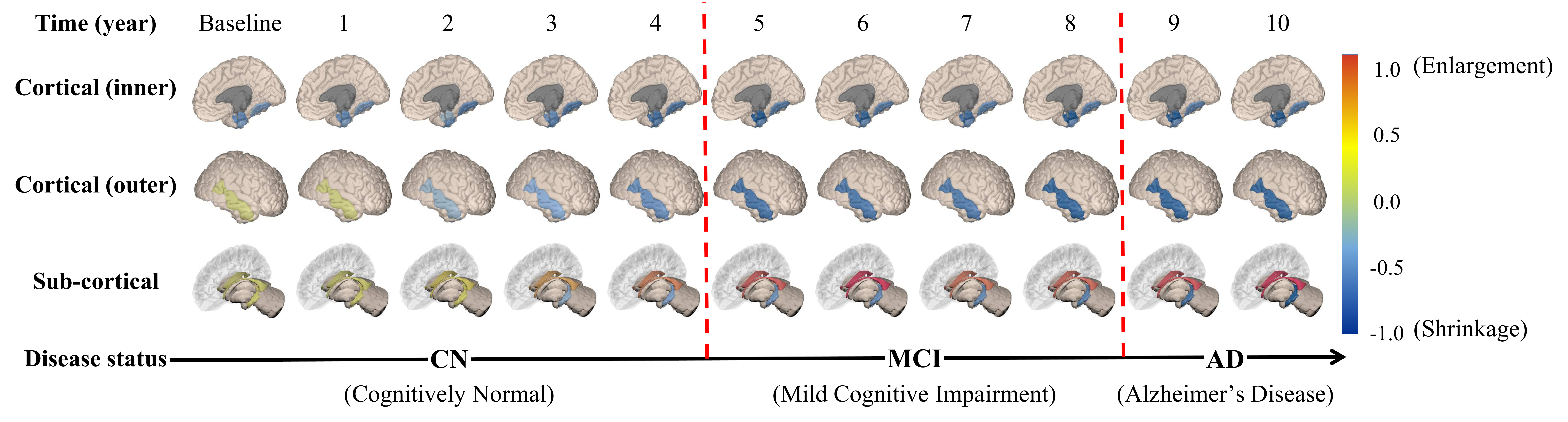}
\caption{An example of morphological changes of six regions in a subject's brain. The color-coded values indicate the amount of volume changes from a baseline time point. The darker color of either blue or red, the more shrunk or enlarged volume from a baseline time point.} 
\label{fig:sample}
\end{figure}
Although the TADPOLE dataset offers numerous kinds of biomarkers to forecast the AD status~\citep{marinescu2018tadpole}, {\ws by following previous studies~\citep{oxtoby2018data,ghazi2019training,albright2019forecasting,lu2019statistical,iddi2019predicting,marinescu2019dive}} in this paper, we considered the six volumetric features of ventricles, hippocampus, fusiform gyrus, middle temporal gyrus, entorhinal cortex, and whole-brain, extracted from the T1-weighted MRI that are shown in Fig. \ref{fig:mri_biomarkers} and cognitive tests such as MMSE, ADAS-cog11, and ADAS-cog13. In the case of volumetric MRI features, we normalized the original values of the biomarkers with each subject's intracranial volume (ICV) to compensate for the inter-subject variability in brain size \citep{davis1977new,ghazi2019training} and further conducted feature-wise linear normalization based on the min/max values to be in the range of $[-1,1]$. For a reference, we illustrated an example of morphological changes of six regions in a subject's brain, who experienced three clinical statuses during the 11-year period, in Fig. \ref{fig:sample}. {\ws In the case of cognitive scores, we normalized their values to be in the range of $[0, 1]$.}

\section{Proposed Method}
\begin{figure}[!t]
\centering
\begin{subfigure}[]{0.9\textwidth}
\centering
    \includegraphics[width=\textwidth]{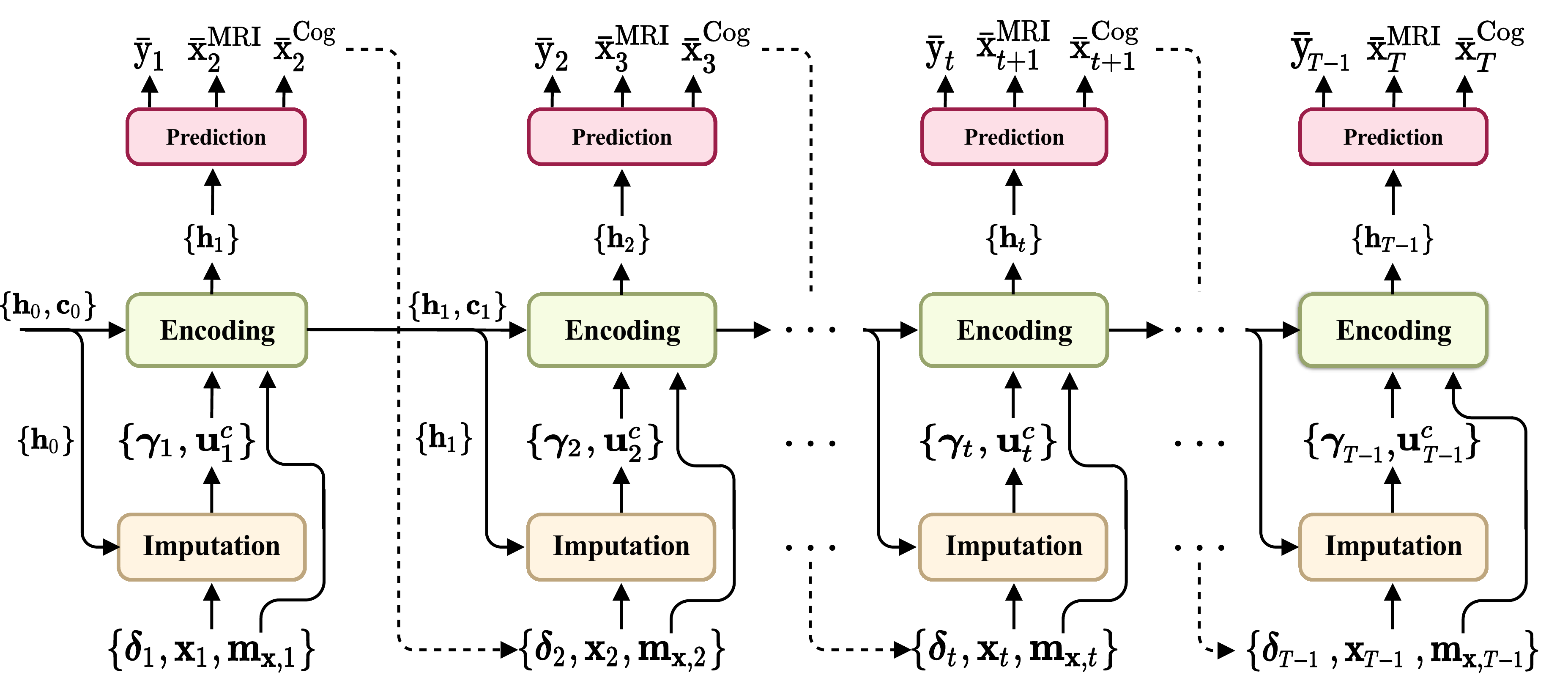}
    \caption{A proposed deep recurrent network architecture}
    \label{architecture}
\end{subfigure}
\begin{subfigure}[]{0.48\textwidth}
\centering
     \includegraphics[width=\textwidth]{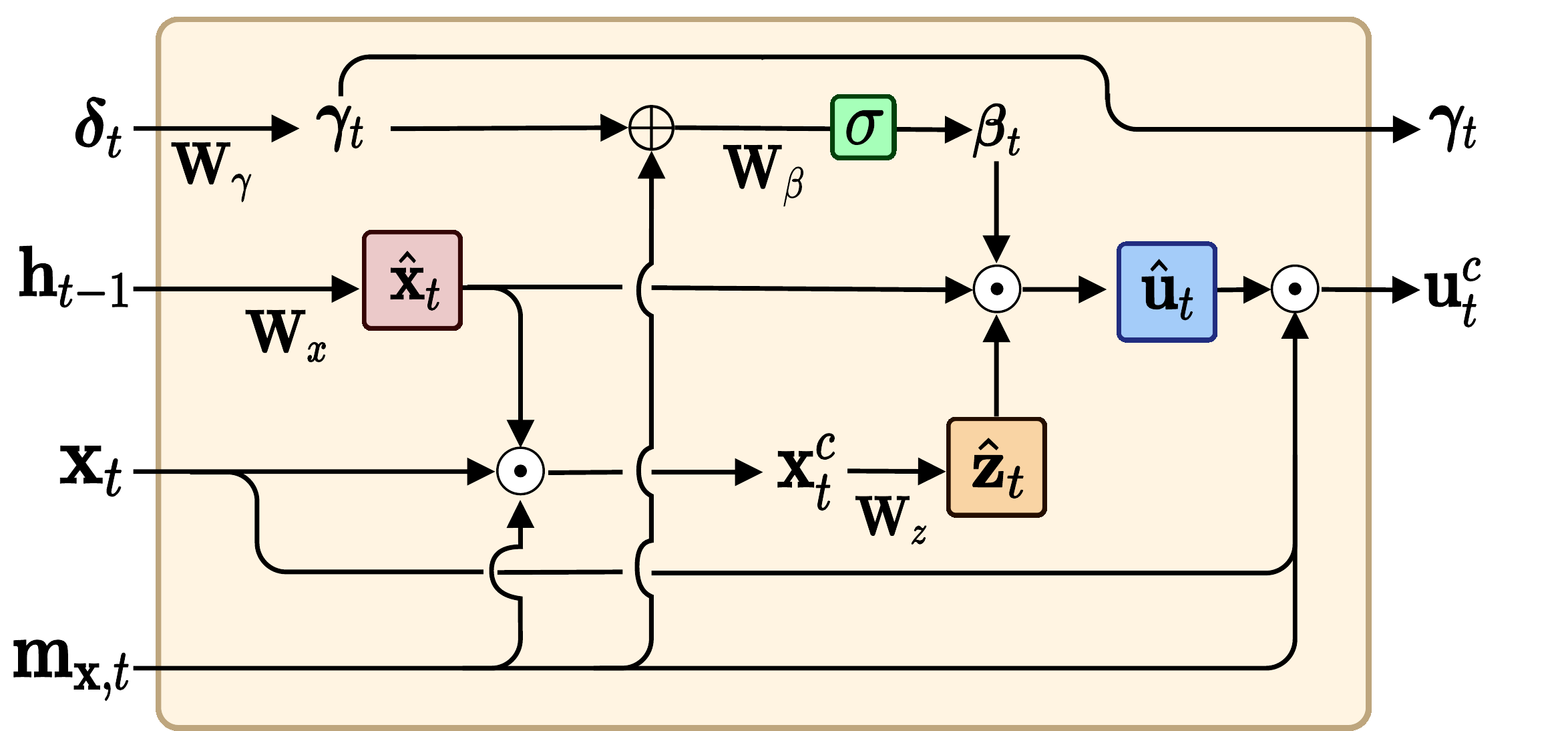}
    \caption{An architecture of our imputation module}
    \label{gate_rits}
\end{subfigure}
\begin{subfigure}[]{0.48\textwidth}
\centering
    \includegraphics[width=\textwidth]{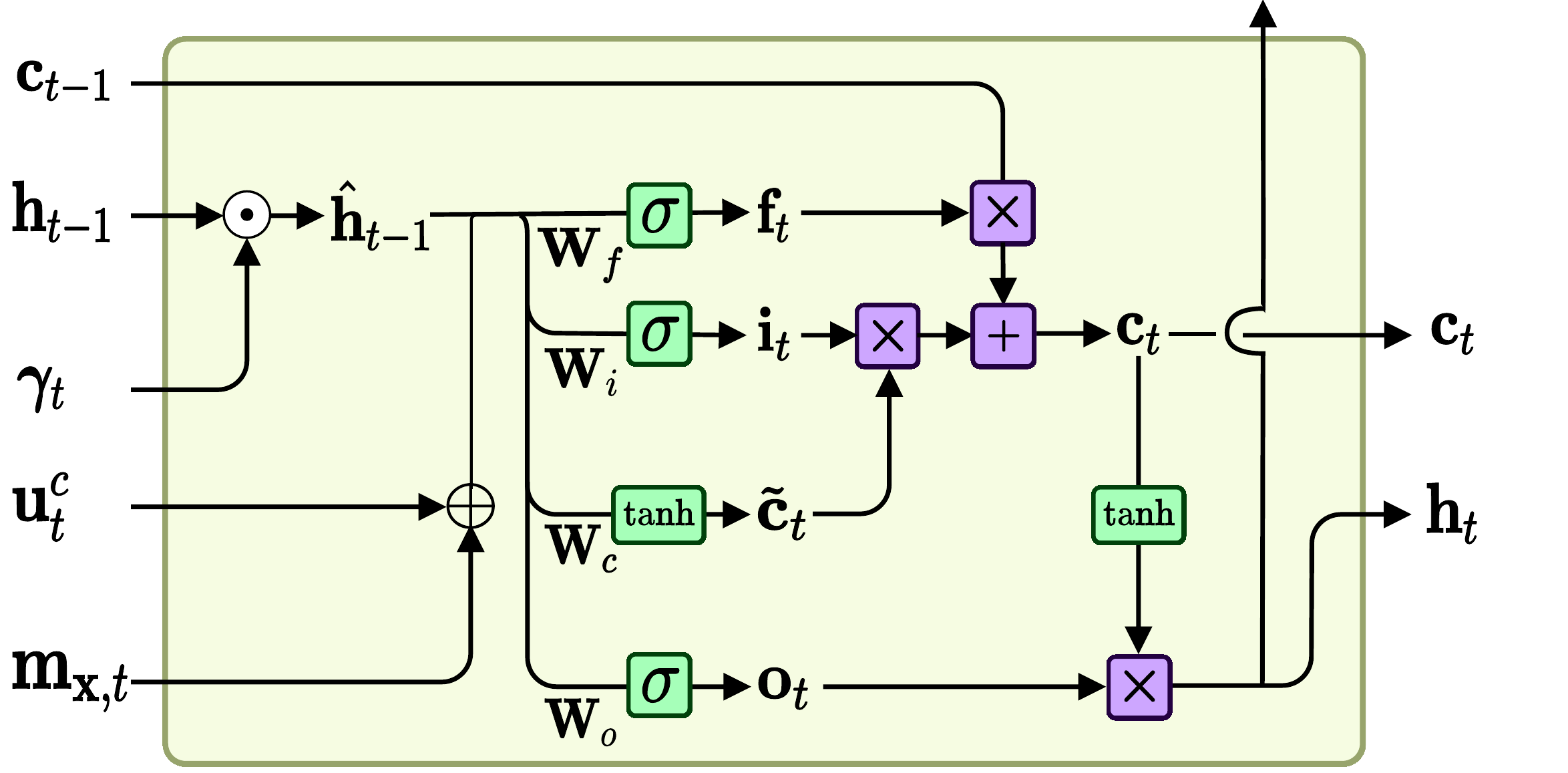}
    \caption{An architecture of our encoding module}
    \label{}
\end{subfigure}
\caption{(a) The architecture of the proposed method combines three modules, \ie, imputation, encoding, and prediction. The modules impute the missing values of the current input, forecast both the biomarker values and cognitive scores, and predict the clinical status at the next time sequence.
(b) Operations to consider the feature-based relations and temporal relations in estimating missing values. 
(c) A variant of a vanilla LSTM unit with $\bgamma$ and $\mathbf{m}_{\bx,t}$. For simplicity, the biases are excluded. If unobserved, the predicted values are used as the input of the next time step, which is denoted as dotted arrows. Solid arrows indicate the process of architecture.}
\end{figure}

In this work, we propose a novel framework for AD progression modeling with incomplete longitudinal data. Specifically, our proposed method is composed of three modules, namely, an imputation module, encoding module, and prediction module, as illustrated in Fig. \ref{architecture}.
At a certain time point, the imputation module first estimates missing values using the encoded latent representations from the previous time point in a deep recurrent network, as well as the multivariate relations among the observed values of the current time point. {\ws Second, the encoding module receives the \emph{complete} observation with the missing values replaced by imputed ones, a mask vector that indicates which values are imputed, and the time-delay information from time interval which represents a duration from the last observed time points. Through this process,} the proposed network encodes and captures underlying temporal characteristics in the given longitudinal data. Lastly, the encoded representations are further transformed to forecast {\ws both the volumetric measurements and cognitive scores of the next time point} and predict the clinical status of the current time point.

\subsection{Notations}
\label{notations}
Throughout the paper, we denote matrices as boldface uppercase letters, vectors as boldface lowercase letters, and scalars as normal italic letters. For a matrix $\bA$, its $(i,j)$-th element and the $i$-th column are denoted as $A(i,j)$ and $\ba_{i}$, respectively. For a vector $\ba$, its $i$-th element is denoted as $a(i)$.

Assume that a set of multivariate time series of $D$ variables over $T$ times from $N$ number of subjects are given $\left\{\bX^{(n)}, \bY^{(n)}\right\}_{n=1}^{N}$, where $\bX^{(n)}=\left[\bx_{1}^{(n)},\dots,\bx_{t}^{(n)},\dots,\bx_{T}^{(n)}\right]\in\mathbb{R}^{D\times T}$, $\bY^{(n)} = \left[\by_1^{(n)},\dots,\by_t^{(n)},\dots,\by_{T}^{(n)}\right]\in\left\{0,1\right\}^{K\times T}$, and $K$ denotes the number of classes considered in a target task. In our case, it corresponds to the clinical state of a patient, \ie, CN, MCI, and AD. To limit clutter, hereafter we omit the superscript $(n)$ unless it is unclear in the context. {\ws Additionally, to clearly define two types of features, we employ input features $\bx_t = \{\bx^{\text{MRI}}_t,\bx^{\text{Cog}}_t\}$, where $\bx^{\text{MRI}}_t$ is MR volumetric features and $\bx^{\text{Cog}}_t$ is cognitive tests features.}

{\ws To inform the model which input features or labels are observed (or missing), we utilized mask vector as two auxiliary variables, \ie, input mask vector and label mask vector that are denoted as $M_{\bx}(d,t)$ and $\bmm_{\by}(t)$.}
To better use the temporal information regarding both missing value imputation and latent feature representation, we further consider a time delay for each observation from the last observation time. For this, we define an auxiliary variable $\bs\in\mathbb{R}^T$ that indicates when the observations were acquired, and a time delay matrix $\bDelta\in \mathbb{R}^{D \times T}$, whose elements \ie, $\Delta(d,t)$ or $\boldsymbol{\delta}_t$, are set as follows:
\begin{equation*}
  \Delta(d,t) = 
  \begin{cases} 
  \text{s}(t) - \text{s}(t-1) +   \Delta(d, t-1) & \text{if } t>1, M_{\bx}(d, t-1)=0,  \\
  \text{s}(t) - \text{s}(t-1)  & \text{if } t>1, M_{\bx}(d,t-1)=1, \\
  1       & \text{if } t=1.
  \end{cases}
\end{equation*}

\subsection{Missing Values Imputation}
\label{subsec:missing_imputation}
We design an architecture that jointly exploits temporal and multivariate relations for imputation. 
Our imputation module is based on a deep recurrent neural network, which becomes an encoding module as described in Section \ref{subsec:encoding_prediction}. Specifically, we take the hidden state values of a recurrent network for the context of temporal relations in a given sequence.

At time $t$, the previous temporal context information up to $(t-1)$ is encoded in the hidden state $\bh_{t-1}$ of an encoding module. 
From the hidden state $\bh_{t-1}$, we first use the temporal relations for the estimation of {\ws between the MRI measurements and cognitive scores} of the current time $t$ as follows:
\begin{equation}
    \hat{\mathbf{x}}_t = \mathbf{W}_x\mathbf{h}_{t-1}+\mathbf{b}_x
    \label{eq:xhat}
\end{equation}
where $\bW_{x}$ and $\bb_{x}$ are learnable parameters. Then, by exploiting the masking vector $\bmm_{\mathbf{x},t}$, we obtain a temporary vector $\mathbf{x}_t^c$ by replacing the missing values with those estimated from the temporal relations, while maintaining the observed both MR features and cognitive scores as follows:
\begin{equation}
    \mathbf{x}_t^c = \bmm_{\bx,t} \odot {\bx}_t + (\mathbf{1}-\bmm_{\bx,t}) \odot \hat{\mathbf{x}}_t.
    \label{eq:xtc}
\end{equation}

We also consider multivariate relations among the MRI features and cognitive scores of the imputed temporal vector $\mathbf{x}_t^c$ in Eq. (\ref{eq:xtc}) by introducing learnable parameters $\bW_{z}$ and $\bb_{z}$ as follows: 
\begin{equation}
    \hat{\bz}_t = \bW_z \bx_t^c + \bb_z.
    \label{eq:zhat}
\end{equation}
Here, it should be noted that we constrain the diagonal elements in $\bW_z$ to be zero, such that the values of elements in $\hat{\bz}_{t}$ are obtained from those of the other variables in $\bx_t^c$ only, thereby focusing purely on the multivariate relations.

Using temporal and multivariate relations, two imputed vectors, $\bx_t^c$ and $\hat{\bz}_t$, are obtained as described above. By considering robustness of the missing value estimates, we further consider combining these two vectors by introducing a weighting coefficient vector that is dynamically determined, based on the missing patterns. In relation to the missing patterns, there are two sources of information, the masking vector $\bmm_{\bx,t}$ and the time delay from the last observation in $\bdelta_{t}$.
We use the time delay information by defining a time decay factor $\bgamma_t \in (\textbf{0},\textbf{1}]^{D}$ as follows:
\begin{equation}
    \bgamma_t = \text{exp}\{-\text{max}(\textbf{0}, \mathbf{W}_\gamma\bdelta_{t}+\mathbf{b}_\gamma)\}
    \label{eq:gamma}
\end{equation}
where $\bW_{\gamma}$ and $\bb_{\gamma}$ are tunable parameters. Combining this time decay factor with a masking vector via concatenation, we compute the weighting coefficient vector $\bbeta_t$ as follows:

\begin{eqnarray}
    \bbeta_t &=& \sigma(\bW_\beta\cdot[\bgamma_t \oplus \bmm_{\bx,t}] + \bb_\beta)
\end{eqnarray}
where $\sigma$ denotes a logistic sigmoid function to guarantee that the coefficients are in the range of $(0,1)$ and $\oplus$ denotes the concatenation operation. Thus, $\bbeta_t$ carries the integrated information about the time decay factors $\bgamma_t$ and the missing patterns at the current time step. With this weighting coefficient vector $\bbeta_t$, we finally estimate the missing values using an interpolation between $\hat{\bx}_{t}$ and $\hat{\bz}_{t}$
\begin{equation}
    \hat{\bu}_t = \bbeta_t \odot \hat{\bz}_t + ({\bf 1}-\bbeta_t) \odot \hat{\bx}_t.
    \label{eq:chat}
\end{equation}
Consequently, we obtain the `\emph{complete}' observation vector $\bu_t^c$ by replacing the missing values with the estimates in Eq. (\ref{eq:chat})
\begin{equation}
    \bu_t^c = \bmm_{\bx,t} \odot {\bx}_t + ({\bf1} - \bmm_{\bx,t}) \odot \hat{\bu}_t.
    \label{eq:complete}
\end{equation}

\subsection{Recurrent Temporal Encoding and Multi-task Prediction}
\label{subsec:encoding_prediction}
After imputing the missing values, we exploit deep recurrent network for temporal encoding. Here, to better reflect the existence of missing values in an observation, we devise a novel computational mechanism, \ie, a variant of a vanilla LSTM cell with $\bgamma$ and $\bmm_{\bx,t}$. In other words, the complete value $\bu_t^c$ in Eq. (\ref{eq:complete}) is fed into the recurrent model.

First, as in the imputation module, we use the time delay information in $\bdelta_{t}$, which is one of the important sources for temporal patterns \citep{che2018grud}. That is, although the temporal context information up to the previous time $(t-1)$ is represented in the hidden state $\bh_{t-1}$ of a recurrent network, their influence in encoding with the current observation $\bu_t^c$ should be treated in different ways depending on when the `real' last observation was acquired. Therefore, we use this crucial information with a temporal decaying factor $\bgamma_t$ in Eq. (\ref{eq:gamma}) before embedding with the current observation $\bu_t^c$ as follows:
\begin{equation}
    \hat{\bh}_{t-1} = {\bh}_{t-1}\odot\bgamma_t.
    \label{eq:hath_t}
\end{equation}
Meanwhile, we also feed the masking vector $\bmm_{\bx,t}$ to the encoding module to let the model know which values are the imputed ones estimated from our imputation module.
Concretely, the internal operations in our encoding module are as follows:
\begin{eqnarray}
    \bff_{t} &=& \sigma\left(\bW_f\cdot[\hat{\bh}_{t-1},\bu_t^c \oplus \bmm_{\bx,t}]+\bb_{f}\right)\nonumber\\
    \bi_{t} &=& \sigma\left(\bW_i\cdot[\hat{\bh}_{t-1},\bu_t^c \oplus \bmm_{\bx,t}]+\bb_{i}\right)\nonumber\\
    \widetilde{\bc}_{t} &=& \textnormal{tanh}\left(\bW_c\cdot[\hat{\bh}_{t-1},\bu_t^c \oplus \bmm_{\bx,t}]+\bb_{c}\right)\nonumber\\
    \bc_{t} &=& \bff_{t}\cdot \bc_{t-1} + \bi_{t}\cdot \widetilde{\bc}_{t}\nonumber\\
    \bo_{t} &=& \sigma\left(\bW_o\cdot[\hat{\bh}_{t-1},\bu_t^c \oplus \bmm_{\bx,t}]+\bb_{o}\right)\nonumber\\
    \bh_t &=& \bo_{t} \cdot \textnormal{tanh}\left(\bc_{t}\right) \nonumber
\end{eqnarray}
where $\left\{\bW_f,\bW_i,\bW_c,\bW_o,\bb_f,\bb_i,\bb_c,\bb_o\right\}$ are the learnable parameters of our modified LSTM cell. $\bff_t, \bi_t, \widetilde{\bc}_t$, and $\bo_t$ represent the outputs of the forget, input, update, output gates, respectively.

Based on the hidden state representation $\bh_{t}$ from the encoding module, our prediction module produces three outcomes, \ie, the MRI measurements $\bar{\bx}^{\text{MRI}}_{t+1}$, the cognitive tests $\bar{\bx}^{\text{Cog}}_{t+1}$ of the next time point, and the clinical state $\bar{\by}_{t}$ of the current time point. We use simple linear and logistic regression models for each of the outcomes as follows:
\begin{eqnarray}
    \bar{\bx}^{\text{MRI}}_{t+1} &=& \bW_{\text{MRI}}\bh_t+\bb_{\text{MRI}} \\
    \bar{\bx}^{\text{Cog}}_{t+1} &=& \bW_{\text{Cog}}\bh_t+\bb_{\text{Cog}} \\
    \bar{\by}_{t} &=& \textnormal{softmax}(\bW_\text{y}\bh_t+\bb_\text{y})
\end{eqnarray}
where $\bW_{\text{MRI}}, \bW_{\text{Cog}}, \bW_\text{y}, \bb_{\text{MRI}}$, $\bb_{\text{Cog}}$, and $\bb_\text{y}$ are parameters. Thus, our prediction module is connected with the encoding module, and the parameters can be tuned jointly with those of the encoding module. From a learning standpoint, this prediction module takes the advantage of a multi-task learning strategy for more predictive representations, comparable to the work of \citep{ghazi2019training}, where they first conducted MRI measurements forecasting and then built an independent classifier for clinical status prediction. {\ws In contrast to \citep{ghazi2019training} which utilized MRI measurements only, we employed both MRI measurements and cognitive scores.} In Section \ref{subsec:single_task}, we present the validity of our multi-task learning by comparing it with the counterpart single-task learning.

{\subsection{Learning}
We defined a composite loss function for the joint training of the three modules, \ie, missing value imputation, temporal encoding, and output prediction.
Specifically, for the imputation loss $\mathcal{L}_\text{imputation}$, we measure the similarity between the observed data and the imputed data by the mean absolute error (MAE) from three different perspectives, \ie, temporal relations in Eq. (\ref{eq:xhat}), multivariate relations in Eq. (\ref{eq:zhat}), and composite relations in Eq. (\ref{eq:chat}).
\begin{equation}
    \mathcal{L}_{\text{imputation}} = \sum_{t=1}^{T}\left(\left|\widetilde{\bx}_t -\hat{\bx}_t\right|+ \left|\widetilde{\bx}_t-\hat{\bz}_t\right| + \left|\widetilde{\bx}_t-\hat{\bu}_t\right|\right)\odot \left(1-\widetilde{\bmm}_{\bx,t}\right)
    \label{eq:imputation_loss}
\end{equation}
where $\widetilde{\bx}_t={\bx}_t\odot \bmm_{\bx,t}$ and $\widetilde{\bM}_{\bx,t}$ is a masking matrix of random removal from observations for imputation module training only.
Empirically, when comparing with a loss defined only with the final imputed values $\bu_{\bx}^{c}$ in Eq. (\ref{eq:complete}), \ie, $\left|\widetilde{\bx}_t-\bu_{\bx}^{c}\right|$, the joint loss in Eq. (\ref{eq:imputation_loss}) enhanced the stability and speed in training. 

As for the MRI biomarker prediction loss $\mathcal{L}_\text{MRI}$, we assessed the correspondence between the model prediction $\bar{\bx}^{\text{MRI}}_{t+1}$ and the future real MRI measurements ${\bx}^{\text{MRI}}_{t+1}$ {\ws by using the mean square error (MSE) as follows:}
\begin{equation}
    \mathcal{L}_{\text{MRI}} = \sum_{t=1}^{T-1}\left(\bx^{\text{MRI}}_{t+1} \odot \bmm_{\bx,t+1} - \bar{\bx}^{\text{MRI}}_{t+1}\odot \bmm_{\bx,t+1}\right)^{2}.
\end{equation}
{\ws Likewise, as for the cognitive scores prediction loss $\mathcal{L}_\text{Cog}$, we evaluated the correspondence between the model prediction $\bar{\bx}^{\text{Cog}}_{t+1}$ and the real cognitive test measurements ${\bx}^{\text{Cog}}_{t+1}$ as follows:
\begin{equation}
    \mathcal{L}_{\text{Cog}} = \sum_{t=1}^{T-1}\left(\bx^{\text{Cog}}_{t+1} \odot \bmm_{\bx,t+1} - \bar{\bx}^{\text{Cog}}_{t+1}\odot \bmm_{\bx,t+1}\right)^{2}.
\end{equation}}
Lastly, for the clinical prediction, we exploited the focal cross-entropy loss $\mathcal{L}_\text{y}$ \citep{lin2017focal} due to the imbalance in samples among the class labels:
\begin{equation}
    \mathcal{L}_{\text{y}} =  -\sum_{t=1}^{T-1}{\bmm_{\by,t}}\left[\sum_{k=1}^{K}\by_t(k)(1-\bar{\by}_{t}(k))^\epsilon\log(\bar{\by}_{t}(k))\right]
    \label{eq:focal_loss}
\end{equation}
where $\epsilon$ is a hyperparameter $(\epsilon\ge0)$. Therefore, we define the overall loss function $\mathcal{L}_\text{total}$ as follows:
\begin{equation}
        \mathcal{L}_{\text{total}} = \alpha\mathcal{L}_{\text{imputation}} + \zeta \left(\mathcal{L}_{\text{MRI}} + \mathcal{L}_{\text{Cog}}\right) + \xi \mathcal{L}_{\text{y}}
        \label{eq:multi_loss}
\end{equation}
where $\alpha$, $\zeta$, and $\xi$ are the hyperparameters to weight the corresponding losses. The optimization of this loss function allows us to train all the parameters of the three modules, \ie, imputation, encoding, and prediction, via stochastic gradient descent in an end-to-end manner.} {\ws Note that the objective functions are chosen differently depending on tasks from the experimental results, \ie, MAE for imputation and MSE for regression.}

\section{Experimental Results and Analysis}
\subsection{Experimental Settings}
We validated the effectiveness of the proposed framework that systematically unifies data-driven missing value imputation and multi-task learning for AD progression modeling in tasks of missing value imputation at the current time and both MRI biomarker and clinical test scores forecasting, as well as the clinical status prediction, we made predictions for every time point (with a one-year interval between consecutive time points) up to 11 time points, thus the next 10 years from the baseline. For each time point, the prediction was conducted based on all the observed historical values and the predicted or imputed values, if unobserved or missing.

We reported the average test results from 5-fold cross-validation.
Specifically, we partitioned the dataset described in Section \ref{sec:materials} into three non-duplicated subsets for training, validation, and testing. For rigorous evaluation, we randomly selected $10\%$ of the subjects in each class as the validation set and another $10\%$ as the test set from the baseline time point. {\ws To train our model, we use the early stopping strategy based on the highest mAUC in validation dataset. On the other hand, we used a different strategy for evaluating performance in flawless imputation task and other tasks such as both MRI biomarker and cognitive test scores forecasting, and clinical status prediction. For the imputation task, we randomly removed $10\%$ of the `true' observation values in samples from all train, validation, and test sets and then used them as the ground truth. However, we utilized full observations for other tasks. In other words, for other tasks, we first loaded the optimized model and evaluate task-specific performance, simultaneously.}
For quantitative evaluation, we used the metrics of MAE and mean relative error (MRE) for the imputation task, MAE for the MRI biomarker prediction task, {\ws root mean square error (RMSE) for the cognitive test scores forecasting task,} and multi-class area under the receiver operating characteristic curve (mAUC) \citep{hand2001simple} for the clinical status prediction task.
We also conducted statistical significance tests in comparison with the competing methods. Briefly, we used the paired, two-sided Wilcoxon signed-rank test~\citep{wilcoxon1992individual} for regression tasks, \ie, imputation and MRI biomarkers forecast, on MAE and MRE, {\ws while RMSE was used for forecasting cognitive tests.} Meanwhile, McNemar's test \citep{mcnemar1947note} on mAUC was applied for the clinical status prediction task.

We compared our proposed method with the following closely-related methods that deal with missing values imputation and prediction tasks:
\begin{itemize}
    \item {Mean imputation with LSTM (LSTM-M)} \citep{wang2018predictive}: Missing values were imputed with the mean values of the respective variables in a training set, and a standard LSTM network was used for MRI biomarker and cognitive test scores forecasting, independently, which were then used for classification with LDA.
    \item {Forward imputation with LSTM (LSTM-F)} \citep{lipton2016rnn}: This is similar to the LSTM-M, except that the missing values were imputed with the most recent observed values.
    \item {Zero imputation with Peephole LSTM (PLSTM-Z)} \citep{ghazi2019training}: A peephole LSTM \citep{gers2000} was used for MRI biomarker and cognitive test scores forecasting with missing values imputed by zeros as input. The predicted MRI biomarkers and cognitive scores were then used for classification via LDA.
    \item {Multi-directional RNN (MRNN)} \citep{yoon2018estimating}: This operates forward and backward within the intra-stream directions, \ie, for each variable, and in the inter-stream directions, \ie, among variables. Unlike a typical bidirectional-RNN, the timing of inputs into the hidden layers was lagged in the forward direction and advanced in the backward direction. The network outputted the MRI biomarkers and cognitive scores measure for the next time point, which were then fed into LDA for clinical status prediction.
\end{itemize}
For all the competing methods, we set $64$ hidden units and trained them using an Adam optimizer~\citep{kingma2014adam} with a learning rate of $5\times10^{-2}$ (LSTM-M and LSTM-F) and $5\times10^{-3}$ (PLSTM-Z and MRNN), a mini-batch size of $64$, and an epoch of $300$. To avoid an overfitting problem and make the training curve converge considerably, we applied an $\ell_2$-regularization by setting the corresponding coefficient to $10^{-4}$.
Regarding the hyperparameters of the composite loss function in Eq. (\ref{eq:multi_loss}), we set them as $\alpha=0.1$, $\zeta=0.5$, and $\xi=0.5$. In Eq. (\ref{eq:focal_loss}), we set $\epsilon = 5$. {\ws For the competing methods, for fairly comparison, we tried to search for optimal hyperparameters in the composite loss function in Eq. (\ref{eq:multi_loss}) excluding $\xi$ term. Specifically, $(\alpha=1.0, \zeta=0.1)$, $(\alpha=1.0, \zeta=0.1)$, $(\alpha=0.25, \zeta=0.5)$, and $(\alpha=1.0, \zeta=0.25)$ were chosen for LSTM-M, LSTM-F, MRNN, and PLSTM-Z, respectively.} 
For evaluation on the test set, we chose the optimal parameters of the comparative networks that achieved the highest mAUC over the validation set. All codes used in our experiments are available at {`\url{https://github.com/ssikjeong1/Deep_Recurrent_AD}'}.

\subsection{Performances}
\begin{table}[!t]
\caption{Performance for an imputation task in terms of MAE and MRE (mean$\pm$std). The best performance is indicated in boldface.}
\label{imputation_result}
\centering{
\begin{threeparttable}
\scriptsize\begin{tabular}{lcc}
\toprule
Models         & MAE                & MRE                \\\toprule
Mean imputation     & ${\color{black}{0.235\pm0.009}^{*}}$                & ${\color{black}{0.788\pm0.062}^{*}}$          \\\midrule
Forward imputation       & ${\color{black}{0.062\pm0.011}^{*}}$       & ${\color{black}{0.204\pm0.043}^{*}}$ \\\midrule
MRNN~\citep{yoon2018estimating}           & ${0.143\pm0.014}^{*}$                & ${0.474\pm0.056}^{*}$          \\\midrule
PLSTM-Z~\citep{ghazi2019training}       & ${0.305\pm0.015}^{*}$                & ${1.000\pm0.000}^{*}$          \\\midrule
\textbf{Ours}  & $\mathbf{0.059\pm0.005}$ & $\mathbf{0.101\pm0.010}$\\\bottomrule
\end{tabular}
\begin{tablenotes}
\scriptsize
    \item ($*: p<0.05$)
\end{tablenotes}
\end{threeparttable}
}
\end{table}


\subsubsection{Missing Value Imputation}
\label{sec:imputation}
We presented the experimental results of the missing values imputation in Table \ref{imputation_result}. 
{\ws First, it is noteworthy that our proposed method achieved the lowest MAE and MRE scores, outperforming all the competing methods under our consideration with a statistical significance of $p<0.05$ for all the competing methods.
Interestingly, Mean and Forward imputation, which took the simple mean or the last observed values for imputation, achieved relatively better performance than their counterpart method of PLSTM-Z that imputed with zeros and applied a peephole LSTM. 
Further, as for the MRNN, even though it used the multi-directional information of temporal relations, it reported the highest MAE and MRE scores.
Moreover, although both our proposed method and MRNN utilized information sources, \ie, temporal relations, our method further considered missing patterns in $\bM_{\bx}$ and time delay in $\bDelta$ systematically through a series of operations \ie, multivariate relations, as described in Section \ref{subsec:missing_imputation}.
We believe that this systematic imputation is beneficial and lead to a significantly enhanced performance. 
We conducted extensive ablation studies on the time delay and missing pattern information and described the results in Section \ref{sec:ablation_study}.}

\begin{table}[!t]
\caption{Performance of forecasting MRI biomarkers one-year later in terms of MAE (mean$\pm$std). The smallest MAE for each MRI biomarker is highlighted in boldface.}
\label{results_regression}

\scriptsize\centering{
\begin{threeparttable}
{\begin{tabular}{lccccccc}
\toprule
\multirow{2}{*}{Models} & \multicolumn{6}{c}{MAE $(\times 10^{-3})$}\\\cline{2-7}
 & \multicolumn{1}{c}{Ventricles} & \multicolumn{1}{c}{Hippocampus} & \multicolumn{1}{c}{Whole Brain} & \multicolumn{1}{c}{Entorhinal Cortex} & \multicolumn{1}{c}{Fusiform Gyrus} & \multicolumn{1}{c}{Middle Temporal Gyrus} \\ \toprule
LSTM-M   & \multirow{2}{*}{8.20$\pm$1.37}$^{*}$                     & \multirow{2}{*}{0.51$\pm$0.24}$^{*}$ & \multirow{2}{*}{28.98$\pm$16.17}$^{*}$ & \multirow{2}{*}{0.26$\pm$0.03}$^{*}$ & \multirow{2}{*}{0.87$\pm$0.36}$^{*}$ & \multirow{2}{*}{0.81$\pm$0.06}$^{*}$ \\
\citep{wang2018predictive}&&&&&&\\\midrule
LSTM-F   & \multirow{2}{*}{8.09$\pm$1.15}$^{*}$                     & \multirow{2}{*}{0.42$\pm$0.08}$^{*}$ & \multirow{2}{*}{26.85$\pm$7.22}$^{*}$ & \multirow{2}{*}{0.25$\pm$0.07}$^{\dagger}$ & \multirow{2}{*}{0.78$\pm$0.13}$^{*}$ & \multirow{2}{*}{0.89$\pm$0.09}$^{*}$ \\
\citep{lipton2016rnn}&&&&&&\\ \midrule
MRNN       & \multirow{2}{*}{6.25$\pm$1.13}$^{*}$& \multirow{2}{*}{0.41$\pm$0.05}$^{*}$& \multirow{2}{*}{26.58$\pm$2.35}$^{*}$ & \multirow{2}{*}{0.31$\pm$0.03}$^{*}$ & \multirow{2}{*}{0.90$\pm$0.10}$^{*}$ & \multirow{2}{*}{0.96$\pm$0.13}$^{*}$ \\
\citep{yoon2018estimating}&&&&&&\\ \midrule
PLSTM-Z    & \multirow{2}{*}{6.04$\pm$1.19}$^{*}$ & \multirow{2}{*}{0.32$\pm$0.05}$^{*}$ & \multirow{2}{*}{23.07$\pm$5.96}$^{\dagger}$& \multirow{2}{*}{0.24$\pm$0.01}$^{*}$ & \multirow{2}{*}{0.74$\pm$0.12}$^{*}$ & \multirow{2}{*}{0.82$\pm$0.11}$^{*}$ \\
\citep{ghazi2019training}&&&&&&\\ \midrule
\textbf{Ours}& \textbf{1.55$\pm$0.16} & \textbf{0.14$\pm$0.00} & \textbf{11.01$\pm$1.22} & \textbf{0.19$\pm$0.01} & \textbf{0.41$\pm$0.02} & \textbf{0.43$\pm$0.05}\\\bottomrule
\end{tabular}}
\begin{tablenotes}
\scriptsize
    \item ($*: p<0.05$, $\dagger:$ no statistical difference)
\end{tablenotes}
\end{threeparttable}
}
\end{table}

\subsubsection{MRI Biomarker Modeling}
The forecast errors in MAE over six volumetric MRI biomarkers are presented in Table \ref{results_regression}.
Unlike the imputation task, LSTM-M and LSTM-F, which used simple imputation, and typical LSTM resulted in relatively higher errors across the MRI biomarkers. 
{\ws However, modeling temporal relations for hidden state representation in MRNN helped boosting performance in Ventricles, Hippocampus, and Whole Brain compared to LSTM-M and LSTM-F. 
From these findings, it could be hypothesized that the MRI biomarker forecasting was not necessarily helpful for downstream task.
However, our understanding of this phenomenon is that because the amount of longitudinal changes of the MRI biomarker values had been subtle, the simple static values were somehow close to the ground truth on average, but it was not helpful for those simple static values to capture the temporal and multivariate relations inherent in the data. 
As for PLSTM-Z, whose major difference from LSTM-M, LSTM-F, and MRNN was the use of a peephole LSTM, it achieved much better MAE scores for all the biomarkers.
However, our proposed method\footnote{The yearly detailed performance of our method is presented in Table \ref{tb:longitudinal_performance_time_points_cog}.} was even superior to PLSTM-Z, with margins of {4.49} (ventricles), {0.18} (hippocampus), {12.06} (whole brain), {0.05} (entorhinal cortex), {0.33} (fusiform gyrus), and 0.39 (middle temporal gyrus).}
From those results, we argue that the piece of masking pattern information played a key roles, as discussed in Section \ref{sec:ablation_study}.

\begin{table}[!t]
\caption{Performance of predicting cognitive scores in terms of RMSE. The smallest RMSE value for each cognitive score is highlighted in boldface. ($\rho$: correlation coefficient)}
\label{results_cognitive_prediction}

\centering{
\begin{threeparttable}
\scriptsize\begin{tabular}{lccccccc}
\toprule
\multirow{2}{*}{Models} & \multicolumn{2}{c}{MMSE (Baseline)} & \multicolumn{2}{c}{ADAS-cog11 (Baseline)} & \multicolumn{2}{c}{ADAS-cog13 (Baseline)} \\
                        & RMSE & $\rho$ & RMSE & $\rho$ & RMSE & $\rho$ \\\toprule
\multirow{2}{*}{LSTM-M~\citep{wang2018predictive}} & 6.825$\pm$1.522$^*$   & \multirow{2}{*}{0.432}  & {8.902$\pm$2.603}$^*$ & \multirow{2}{*}{0.651}& 11.544$\pm$2.146$^*$ &\multirow{2}{*}{0.675}\\
&(8.854$\pm$3.898)&& (8.235$\pm$4.080) && (12.542$\pm$4.038) &\\\midrule
\multirow{2}{*}{LSTM-F~\citep{lipton2016rnn}}      & 6.078$\pm$3.018$^*$   & \multirow{2}{*}{0.382}   & {7.510$\pm$0.676}$^*$ & \multirow{2}{*}{0.645}& 10.730$\pm$1.888$^*$ & \multirow{2}{*}{0.722}\\
&(8.943$\pm$4.079)&& (8.154$\pm$3.315) && (11.946$\pm$5.001) &\\\midrule
\multirow{2}{*}{MRNN~\citep{yoon2018estimating}}   & 3.758$\pm$0.614$^*$   & \multirow{2}{*}{0.616}  & {7.208$\pm$1.191}$^*$ & \multirow{2}{*}{0.674}& 9.514$\pm$1.551$^*$ & \multirow{2}{*}{0.685}\\ 
&(3.951$\pm$0.559)&& (8.344$\pm$2.191) && (11.529$\pm$2.690) &\\\midrule

\multirow{2}{*}{PLSTM-Z~\citep{ghazi2019training}} & {3.490$\pm$0.621}$^*$ & \multirow{2}{*}{0.698}  & {5.449$\pm$0.913}$^*$ & \multirow{2}{*}{0.833}& {6.986$\pm$0.983}$^*$ & \multirow{2}{*}{0.847}\\
& (2.929$\pm$0.505)& &(5.269$\pm$1.682)& &(6.971$\pm$1.997) &\\\midrule
\multirow{2}{*}{\textbf{Ours}}& \textbf{2.319$\pm$0.123}  & \multirow{2}{*}{0.863} & \textbf{4.310$\pm$0.210} &\multirow{2}{*}{0.900}&\textbf{5.307$\pm$0.202} & \multirow{2}{*}{0.916}\\
                                  & (2.181$\pm$0.096)     & & (4.293$\pm$0.825) & & (5.448$\pm$0.706) &\\
\bottomrule
\end{tabular}
\begin{tablenotes}
{\scriptsize
    \item ($*: p<0.05$)}
\end{tablenotes}
\end{threeparttable}
}
\end{table}

{\ws\subsubsection{Cognitive Scores Prediction}
\label{subsec:Cognitive_Scores_Prediction}
The forecast errors in RMSE over three cognitive test scores are presented in Table \ref{results_cognitive_prediction}.
Similar to the MRI biomarker modeling task, LSTM-M and LSTM-F that combined typical LSTM with the simple imputation method resulted in relatively higher errors across cognitive scores. On the other hand, MRNNs that employed temporal relations outperformed the simple imputation methods such as LSTM-M and LSTM-F. From these results, we argue that the cognitive score forecast is associated with MRI biomarker modeling, and utilizing these features are beneficial for downstream work. It can be inferred that the longitudinal changes of the MRI biomarker values represent morphological changes, and in real clinical practice, the trajectories of the cognitive scores are, reflected in the clinician's opinions for judging the patient's disease progression. As for PLSTM-Z, it achieved much better RMSE scores for all the biomarkers than others (LSTM-M, LSTM-F, and MRNN). However, our proposed method\footnote{The yearly detailed performance of our method is presented in Table \ref{tb:longitudinal_performance_time_points_cog}.} outperformed PLSTM-Z, with margins of {1.171 (0.748)}, MMSE, {1.139 (0.976)}, ADAS-cog11, and 1.679 (1.523), ADAS-cog13. 

}

\begin{table}[!t]
\caption{Performance on a multi-class (AD vs. MCI vs. CN) classification task (mean$\pm$std). The best performance in each metric is highlighted in boldface.}
\label{results_multi_classification}
\centering{
\begin{threeparttable}
\scriptsize\begin{tabular}{lccc}
\toprule
Models & mAUC & Recall & Precision \\
                       \toprule
LSTM-M~\citep{wang2018predictive} & 0.758$\pm$0.034$^{***}$    & {0.501$\pm$0.201} & 0.577$\pm$0.218 \\ \midrule
LSTM-F~\citep{lipton2016rnn}& 0.741$\pm$0.018$^{***}$    & 0.549$\pm$0.162       & 0.521$\pm$0.107 \\ \midrule
MRNN~\citep{yoon2018estimating}& 0.774$\pm$0.045$^{***}$    & {0.611$\pm$0.045}               & 0.580$\pm$0.092                \\ \midrule
PLSTM-Z~\citep{ghazi2019training}& 0.842$\pm$0.035$^{***}$    & {0.706$\pm$0.092}               & 0.636$\pm$0.093
\\ \midrule
\textbf{Ours}      & \textbf{0.878$\pm$0.022} & \textbf{0.723$\pm$0.071}&\textbf{0.710$\pm$0.071}\\
\bottomrule
\end{tabular}
\begin{tablenotes}
{\scriptsize
    \item ($***: p<0.001$)}
\end{tablenotes}
\end{threeparttable}
}
\end{table}

\subsubsection{Clinical Status Prediction}
In the clinical status prediction task, due to high imbalance in sample sizes among the three classes of AD, MCI, and CN, we additionally reported precision and recall, along with mAUC, in Table \ref{results_multi_classification}.
Notably, like the MRI biomarker and cognitive test score forecast, LSTM-M and LSTM-F based on the standard LSTM architecture were inferior to the other competing methods. Among the comparative methods, PLSTM-Z~\citep{ghazi2019training} resulted in boosted performance in all metrics of mAUC, recall, and precision. Meanwhile, our proposed method\footnote{The yearly computed performance of our method and the respective confusion matrices are presented in Table \ref{tb:longitudinal_performance_time_points_cog} and Fig. \ref{confusion_matrix_cog}, respectively.} outperformed all the competing methods by achieving the highest scores in mAUC, recall, and precision with margins of 0.036, 0.017, and 0.074, respectively, compared to PLSTM-Z.


\section{Discussion}
\label{sec:discussion}
\subsection{Effect of Multi-task Learning}
\label{subsec:single_task}
It is noteworthy that in the clinical status prediction task, one of the major differences of our method from the other comparative methods was to jointly optimize the parameters of the MRI biomarker regressor, {\ws cognitive scores regressor,} and the clinical status prediction classifier. Thus, we also conducted an experiment by modifying our method such that it was trained with a single task of MRI biomarker forecasting and {\ws cognitive test scores forecasting}, and then independently trained an LDA-based classifier, under the same condition as the comparative models. That is, we trained our model by setting the hyperparameters of $\xi=0$ ($\alpha=1.0$, $\zeta=0.75$) in the composite loss function. We called this variant of the model as single-task learning.

{\ws In single-task learning, the results of our method were $0.869\pm0.019$ in mAUC, $0.732\pm0.060$ in recall, and $0.672\pm0.071$ in precision, respectively, which are higher than the other comparative methods with a margin of 0.111 (vs. LSTM-M), 0.128 (vs. LSTM-F), 0.095 (vs. MRNN), and 0.027 (vs. PLSTM-Z) in terms of mAUC. Compared to multi-task learning in our proposed method, the joint learning of MRI biomarker forecast, cognitive test scores forecast, and clinical status prediction was helpful at enhancing the performance by 0.009 in mAUC and 0.038 in precision.}


\begin{figure}[!t]
\centering
    \includegraphics[width=0.8\textwidth]{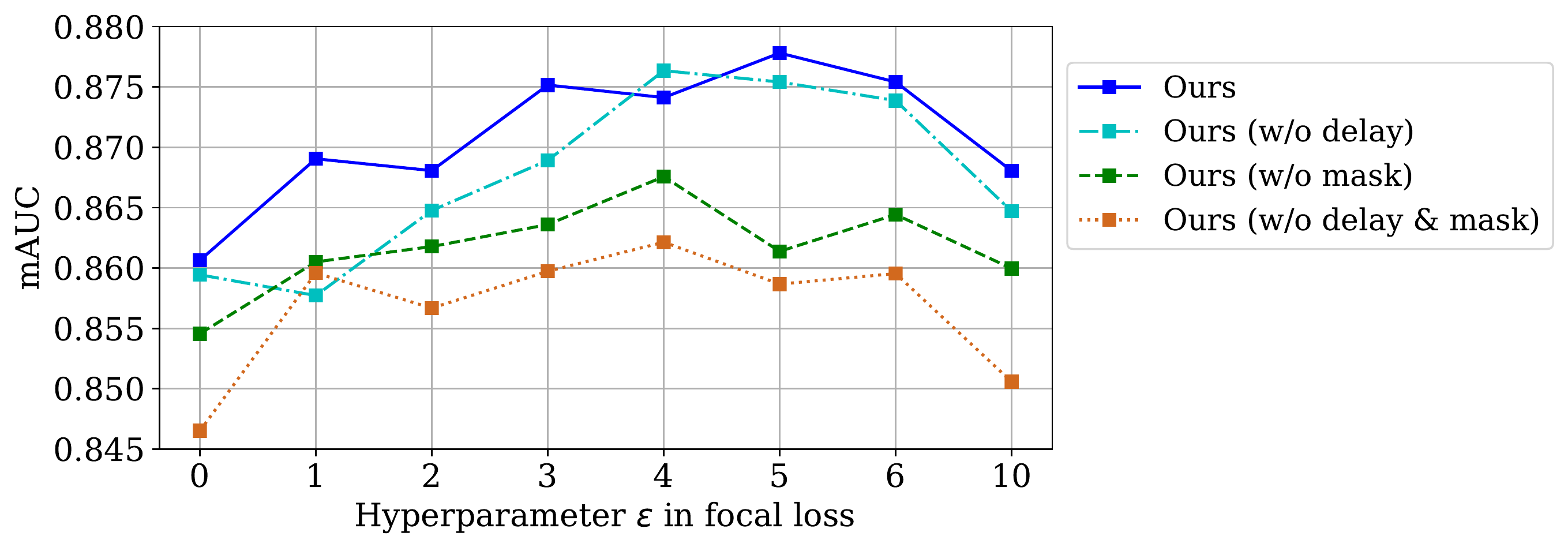}
    \caption{\ws Result of ablation studies on the effectiveness of the time delay and missingness information for the proposed method. In this case, we selected the classification results from the setting of the $\epsilon$ value, which is the hyperparameter in the focal loss function.}
    \label{Ablation_imputation}
\end{figure}


\subsection{Ablation Studies}
\label{sec:ablation_study}
{\ws We first investigated the effect of varying a pair of hyperparameters $\{\alpha,\zeta,\xi\}$ in Eq.~(\ref{eq:multi_loss}). We considered these parameters as a ratio to weigh the missing values and future biomarker regression and clinical status classification to achieve optimal performance. For each parameter, we defined a set of range values for these hyperparameters as $\{0.1,0.25,0.5,0.75,1.0\}$. Based on the highest average AUC score of $0.878$ with $\alpha=0.1$, $\zeta=0.5$, and $\xi=0.5$ are chosen.

Our proposed method used a focal loss that considers the imbalance of samples among classes. We varied the value of the hyperparameter $\epsilon$ in Eq. (\ref{eq:focal_loss}) from 0 to 10 with an interval of 1. The result is presented in Fig. \ref{Ablation_imputation} with a blue solid line. Note that when $\epsilon=0$, which corresponds to the conventional cross-entropy loss, the mAUC was $0.861\pm0.018$ still outperforming the competing methods. Clearly, due to the high imbalance in the number of samples in our training set, it was useful for performance improvement to apply non-zero $\epsilon$ values, achieving the highest performance when $\epsilon=5$.

To validate the effectiveness of using the missingness information, \ie, mask vectors and the time-delay information between observations, we also conducted experiments with and without them. The results are shown in Fig. \ref{Ablation_imputation}. With the information on the missingness, our model performed better than with the delay information. Moreover, when ignoring both time-delay and missingness information, we achieved the lowest performance over the different experimental settings. As a result, we confirmed not only model performance was affected in the order of information on the time delay and missingness, but also it showed justification for the proposed method.}

\begin{figure}[!t]
\begin{subfigure}[]{\textwidth}
\centering
    \includegraphics[width=0.65\textwidth]{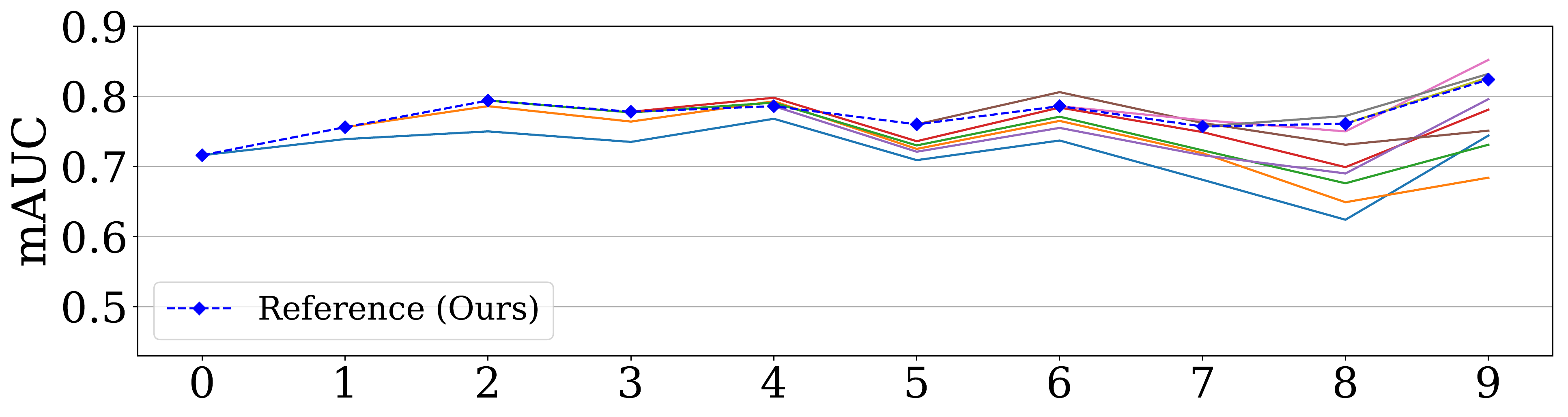}
    \caption{{Ours}}
    \label{fig:Prognosis_result_cog_a}
\end{subfigure}
\begin{subfigure}[]{\textwidth}
\centering
    \includegraphics[width=0.65\textwidth]{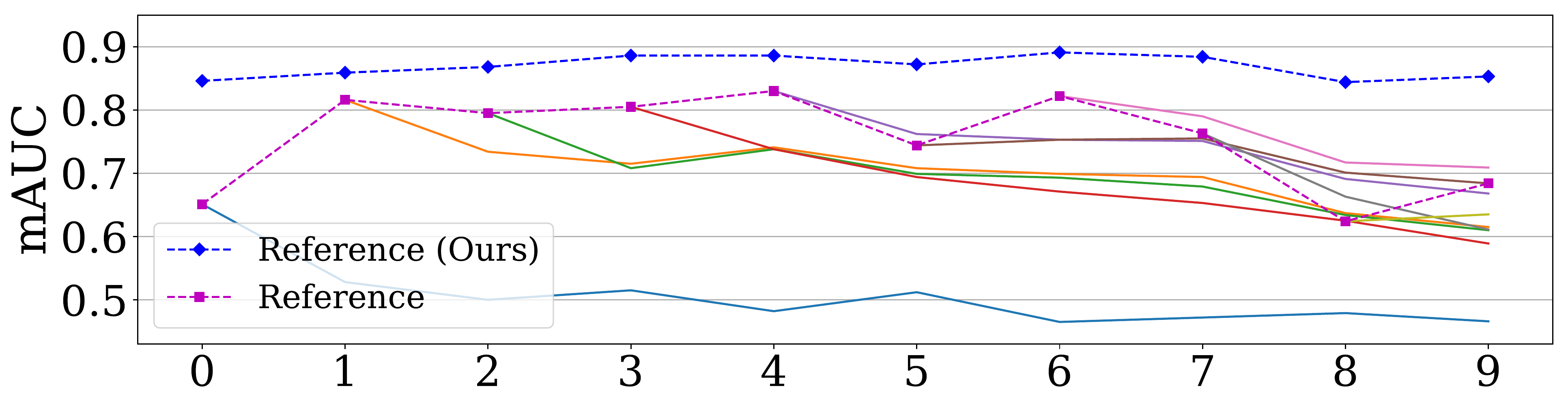}
    \caption{MRNN~\citep{yoon2018estimating}}
    \label{fig:Prognosis_result_cog_c}
\end{subfigure}
\begin{subfigure}[]{\textwidth}
\centering
    \includegraphics[width=0.65\textwidth]{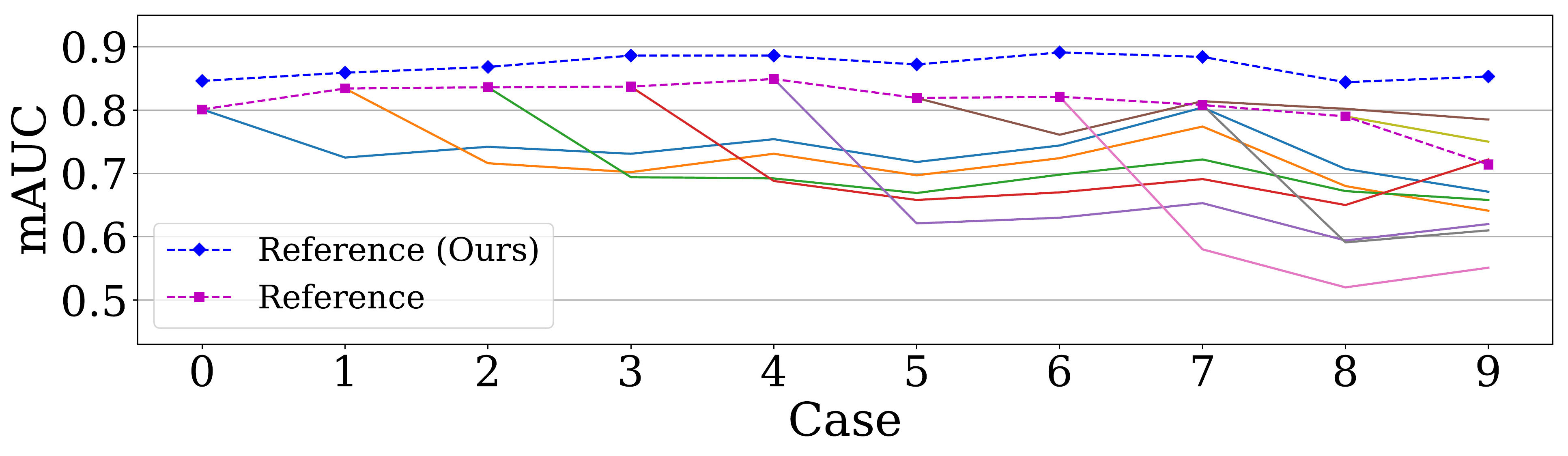}
    \caption{PLSTM-Z~\citep{ghazi2019training}}
    \label{fig:Prognosis_result_cog_b}
\end{subfigure}
\caption{Disease prognostic results from historical observations between the proposed method, MRNN, and PLSTM-Z. From top to bottom, the panels display the disease prognostic performance over time using our method, MRNN, and PLSTM-Z. Specifically, the colored solid lines in each panel mean that the patient's disease prognosis results are represented using accumulated data up to that point. In addition, we plotted the comparative method's performance in one-time-point prediction with a dotted magenta line.}
\label{fig:Prognosis_result}
\end{figure}


\subsection{Prediction at Multiple Time Points within a 10-Year Range}
\label{ghazi_impute_prognosis}
{\ws To verify the validity of our proposed method for multiple-time-points prediction, we compared the prognostic performance between MRNN, PLSTM-Z, and our method, which ranked in the top three in mAUCs for single next-time-point prediction in Table \ref{results_multi_classification}.
At each time point, we made clinical status predictions for multiple subsequent time points up to 10 years beyond the baseline with the historical observations and the imputed or predicted values for missing ones.
To do this, we divided 10 cases based on the number of historical time points used for clinical status prediction and then measured the mAUC for each case. As an example, for case 5, when having used 5 historical time points, from 0 (baseline) to 4 (4 years), then the clinical statuses over the following 5 time points (5 years) were predicted, along with the MRI biomarker and cognitive test score forecast. The mAUCs of those methods over time are presented with line graphs in Fig. \ref{fig:Prognosis_result}. In the figure, we also plotted our proposed method's performance in one-time-point prediction with a dotted blue line for reference. First, from a bias-variance point of view, our proposed method notably showed high (\ie, low-bias) and stable (\ie, low-variance) mAUC scores across the cases. Meanwhile, PLSTM-Z and MRNN showed low mAUC (\ie, high-bias) and unstable mAUC (\ie, high-variance), respectively. Second, the data-driven imputation methods of MRNN and ours were superior to the counterpart method of PLSTM-Z excluding baseline (case 0). As a result, the data-driven imputation methods are relatively robust to predict multiple time points than the method ignoring the missing value, such as PLSTM-Z.}

\begin{figure}[!thp]
    \centering
    \includegraphics[width=0.8\textwidth]{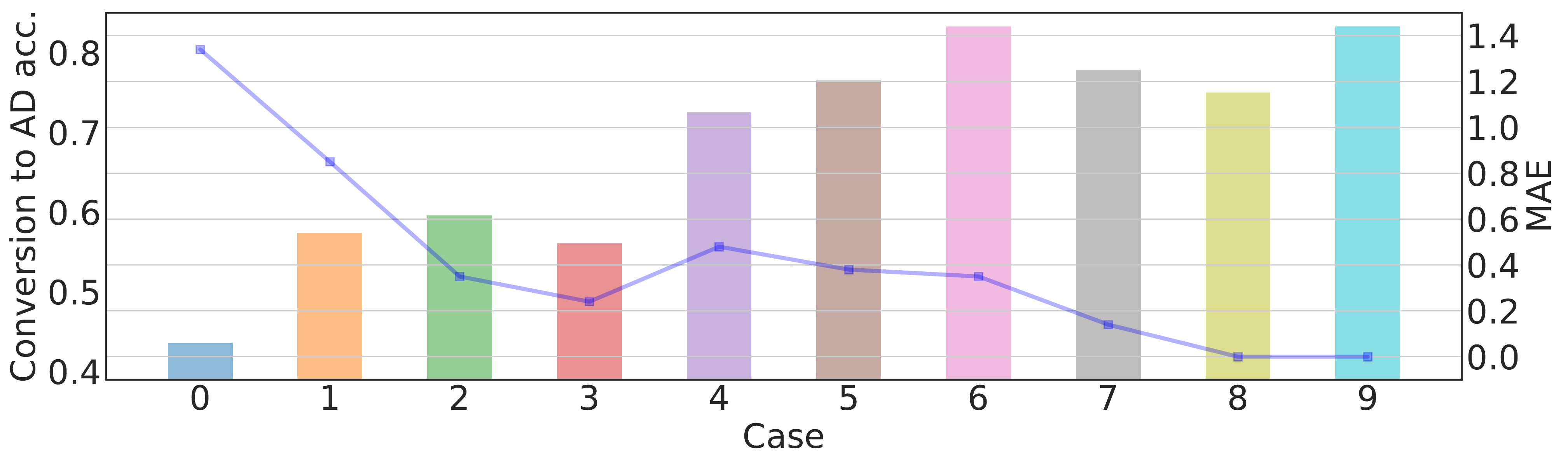}
    \caption{Conversion-to-AD accuracy for our proposed method. Each box indicates the ability to capture changes in a patient's disease status. Furthermore, the solid line represents the absolute values of the difference between the actual patient label transition time and the model's predictive label conversion time. Note that each case denotes that conversion-to-AD predictions are performed using data accumulated up to that time.}
    \label{fig:prognosis_historical_conversion_label_prediction_MTL}
\end{figure}

{\ws For an early identification of AD progression risk, we also considered the predictive accuracy of our method in terms of conversion-to-AD within the period of consideration. Based on our 10-year range prognosis, we measured the accuracy and MAE for the conversion-to-AD, and the results are illustrated in Fig.~\ref{fig:prognosis_historical_conversion_label_prediction_MTL}. In the figure, we observed improved accuracies and decreased MAE across cases.}

\begin{figure}[!t]
\begin{subfigure}[]{0.49\textwidth}
\centering
    \includegraphics[width=\textwidth]{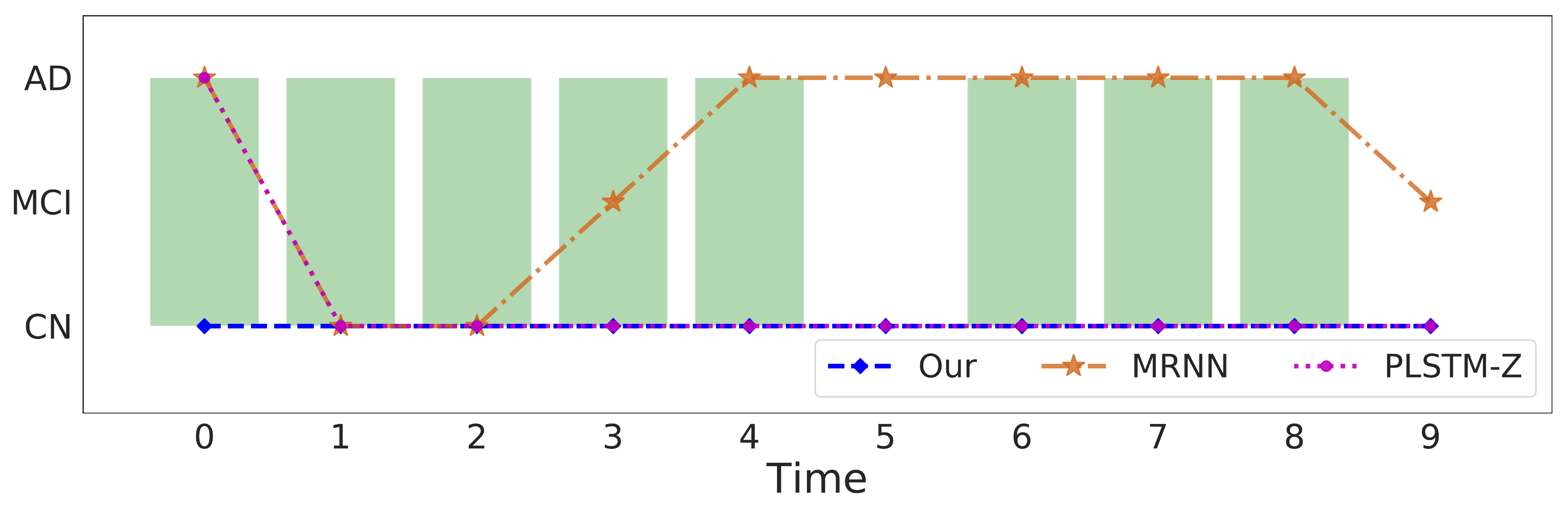}
    \caption{CN}
    \label{fig:Diagnosis_CN}
\end{subfigure}
\begin{subfigure}[]{0.49\textwidth}
\centering
    \includegraphics[width=\textwidth]{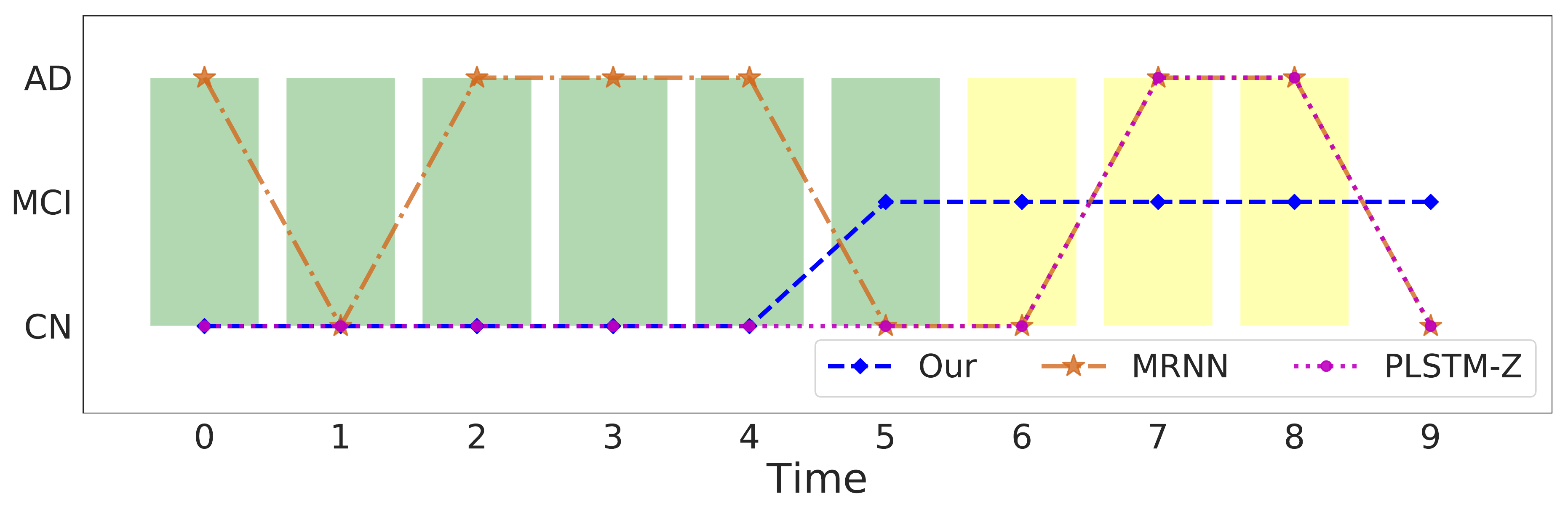}
    \caption{CN-MCI}
    \label{fig:Diagnosis_CN_MCI}
\end{subfigure}
\begin{subfigure}[]{0.49\textwidth}
\centering
    \includegraphics[width=\textwidth]{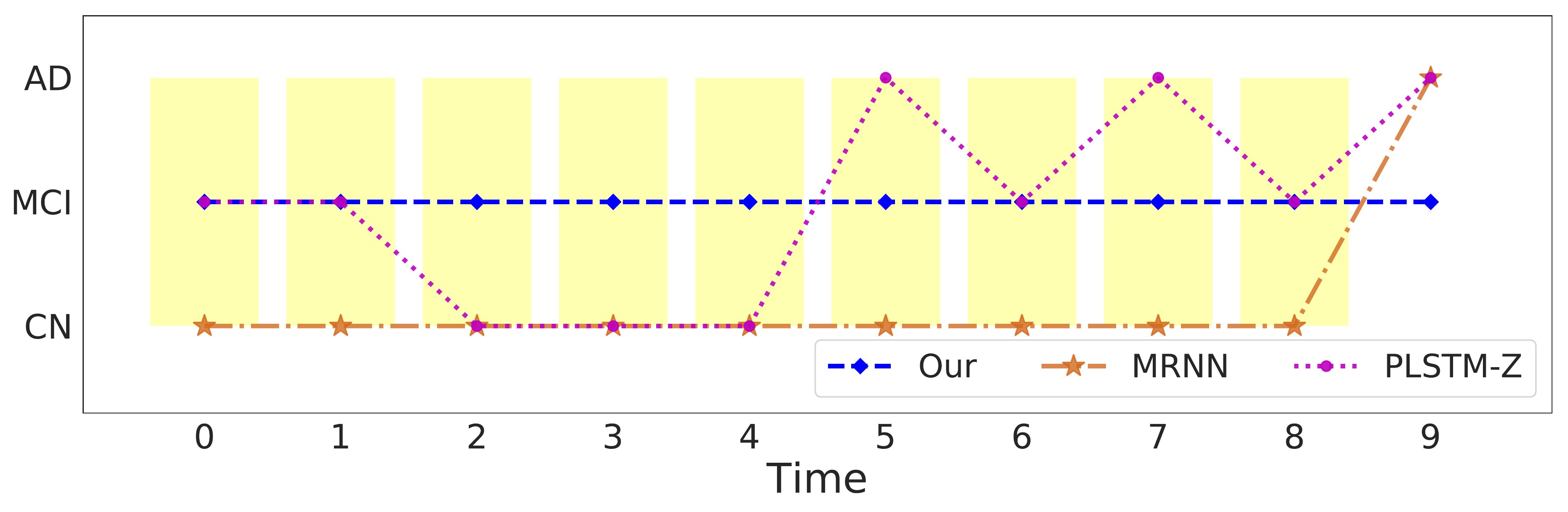}
    \caption{MCI}
    \label{fig:Diagnosis_MCI}
\end{subfigure}
\begin{subfigure}[]{0.49\textwidth}
\centering
    \includegraphics[width=\textwidth]{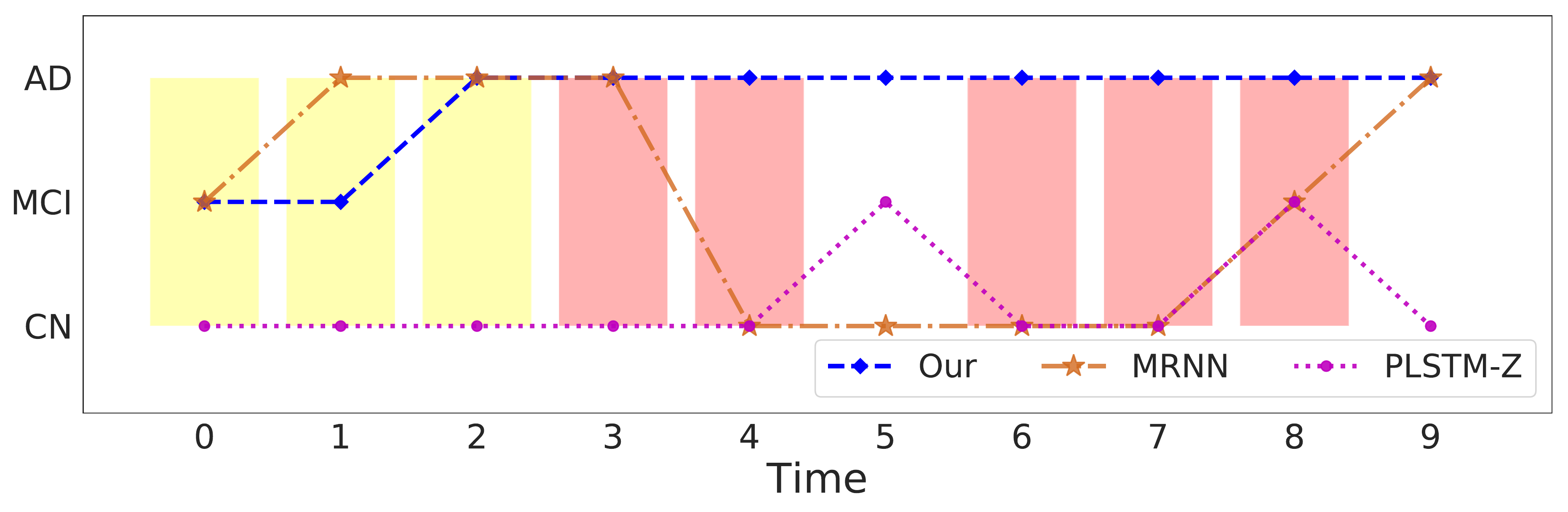}
    \caption{MCI-AD}
    \label{fig:Diagnosis_MCI_AD}
\end{subfigure}
\begin{subfigure}[]{0.49\textwidth}
\centering
    \includegraphics[width=\textwidth]{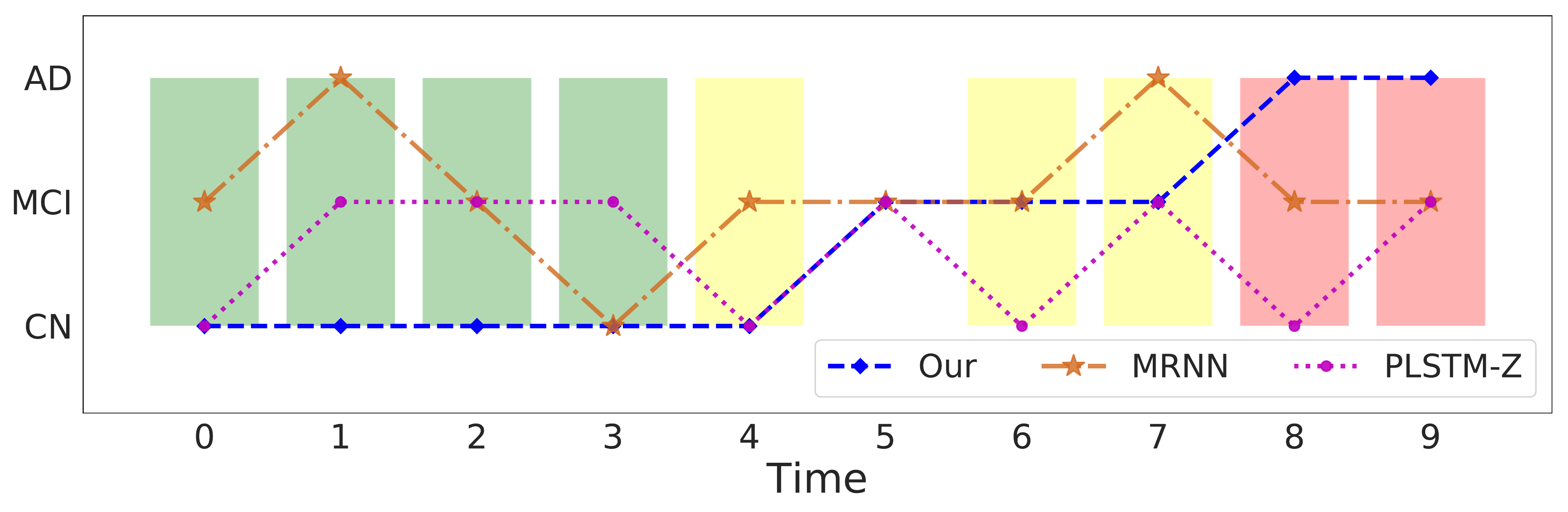}
    \caption{CN-MCI-AD}
    \label{fig:Diagnosis_CN_MCI_AD_1}
\end{subfigure}
\begin{subfigure}[]{0.49\textwidth}
\centering
    \includegraphics[width=\textwidth]{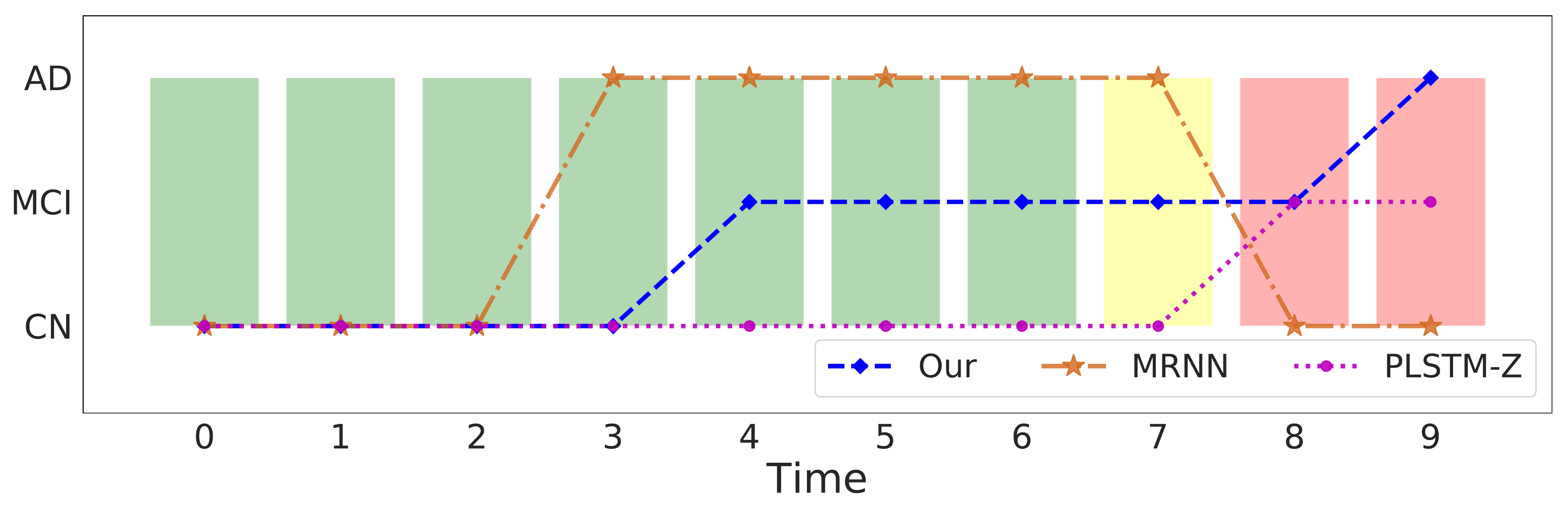}
    \caption{CN-MCI-AD}
    \label{fig:Diagnosis_CN_MCI_AD_2}
\end{subfigure}
\caption{Results of longitudinal status prediction between the proposed method, MRNN, and PLSTM-Z. The plots in each panel represent clinical status conversion from different subjects. Green, yellow, red, and white represent CN, MCI, AD, and missing in the true label, respectively.}
\label{fig:Prognosis_result_individual}
\end{figure}

\subsection{Irreversible Neurodegenerative Alzheimer's Disease}
{\ws The predictive models should also reflect pathological characteristics for the irreversible neurodegeneration of AD. That is, if a subject is under the state of AD at one point, it is not expected for the subject to reverse to MCI or CN in the future. With this consideration, we compared the longitudinal clinical status predictions by MRNN, PLSTM-Z, and our proposed method, shown in Fig. \ref{fig:Prognosis_result_individual}. In the figure, we presented predictions for six different subjects with different progression types. While MRNN and PLSTM-Z made many state-reversing predictions, our proposed method did not show any such non-acceptable predictions. For the subjects with CN (Fig. \ref{fig:Diagnosis_CN}) or MCI (Fig. \ref{fig:Diagnosis_MCI}), our method predicted CN or MCI over the whole period. In Fig. \ref{fig:Diagnosis_CN_MCI} and Fig.~\ref{fig:Diagnosis_MCI_AD}, our method not only predicted earlier disease stage but also not made any state reversing. Moreover, in Fig. \ref{fig:Diagnosis_MCI_AD}, and Fig.~\ref{fig:Diagnosis_CN_MCI_AD_1}, although predictions are misplaced behind by up to three time points, our method did not make any state-reversing predictions.}


\begin{figure}[!t]
\begin{subfigure}[]{0.47\textwidth}
\centering
    \includegraphics[width=1\textwidth]{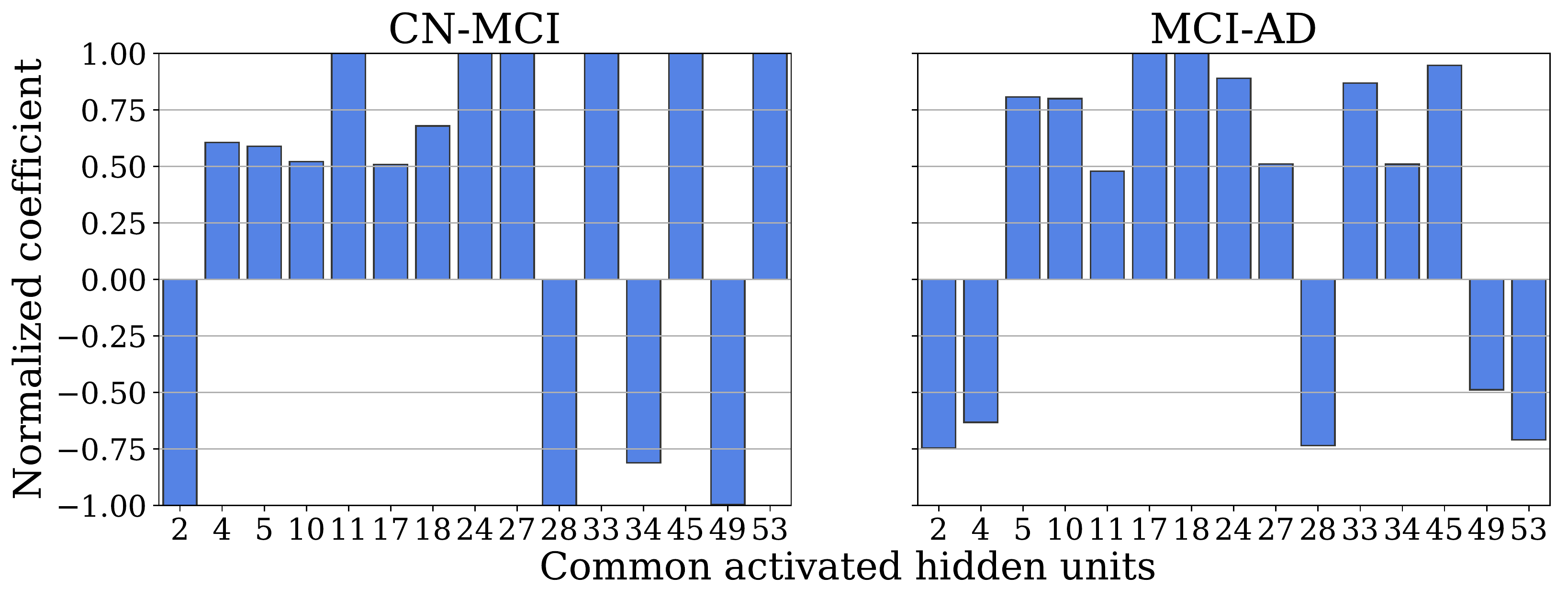}
    \caption{}
\end{subfigure}
\begin{subfigure}[]{0.47\textwidth}
\centering
    \includegraphics[width=\textwidth]{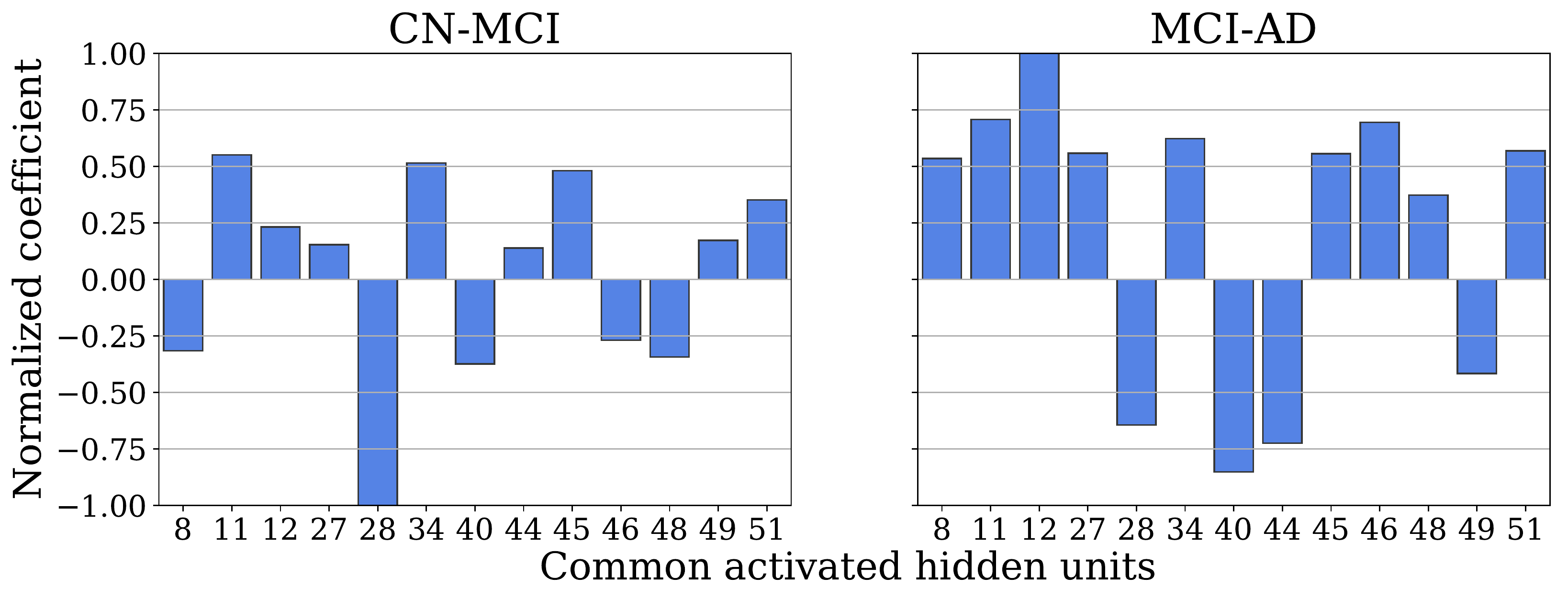}
    \caption{}
\end{subfigure}
\begin{subfigure}[]{.47\textwidth}
\centering
    \includegraphics[width=\textwidth]{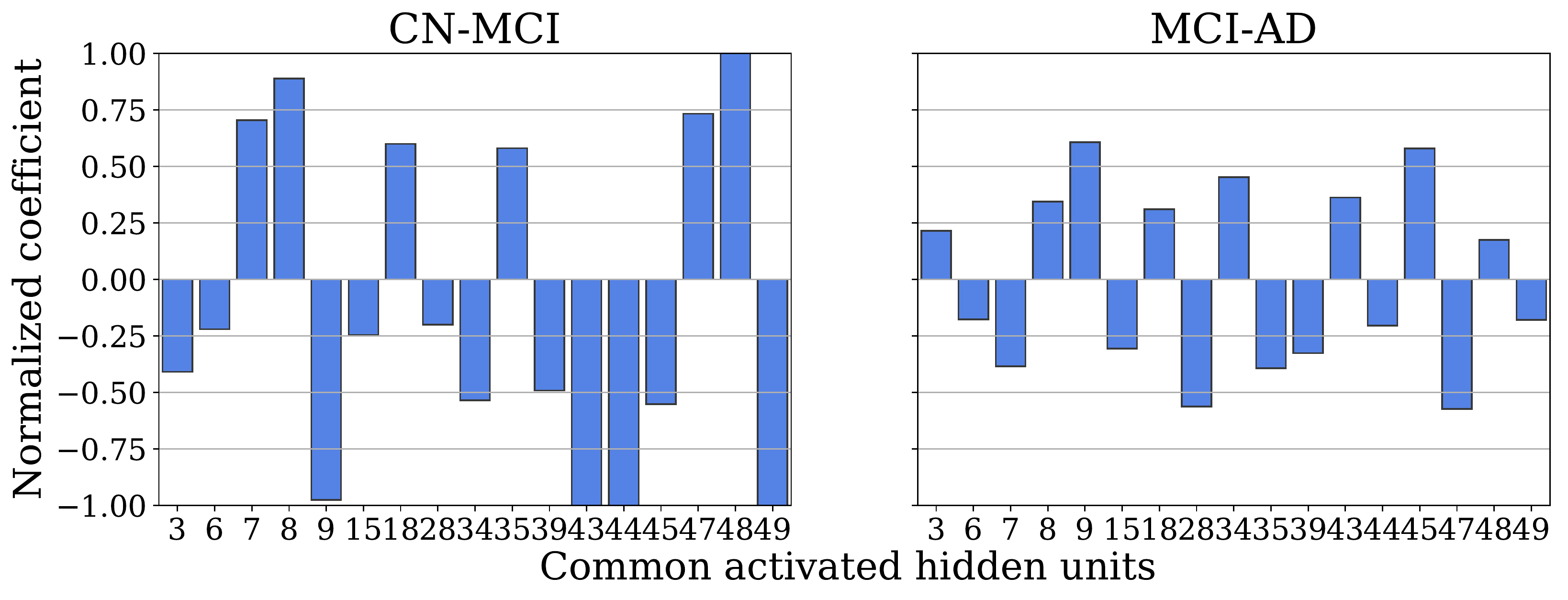}
    \caption{}
\end{subfigure}
\begin{subfigure}[]{0.47\textwidth}
\centering
    \includegraphics[width=\textwidth]{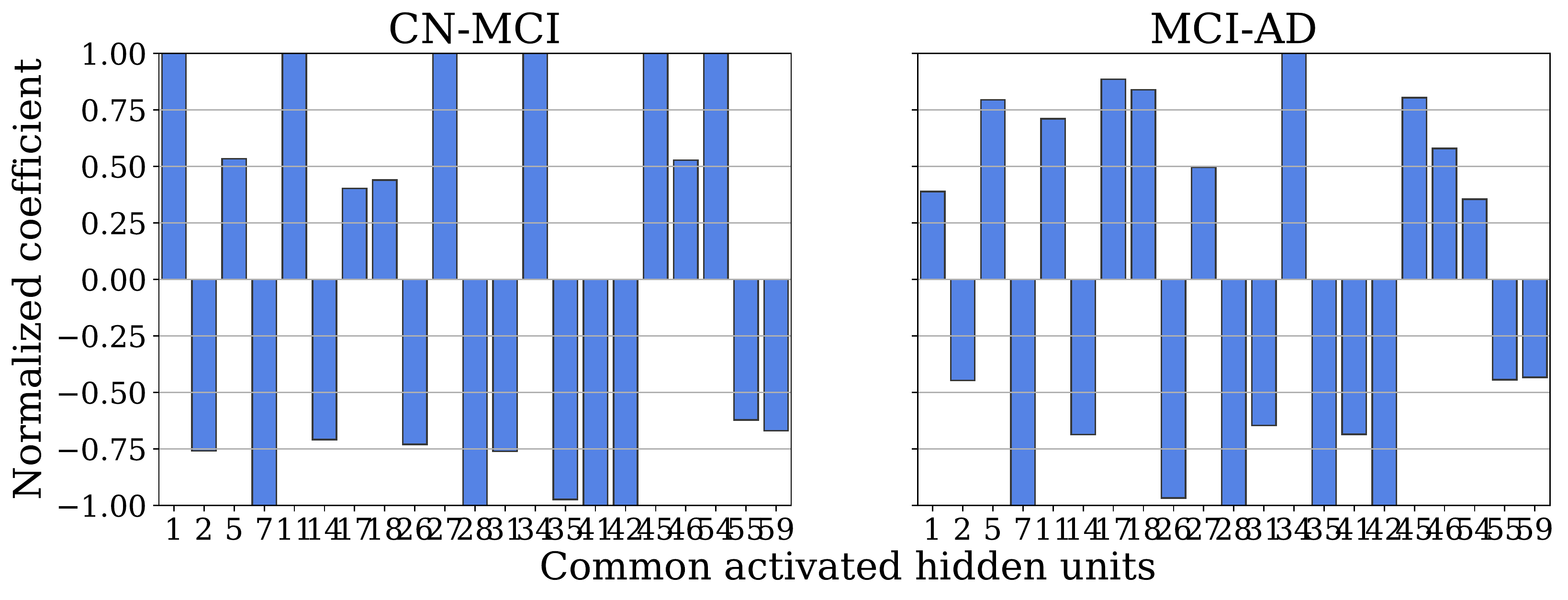}
    \caption{}
\end{subfigure}
\begin{subfigure}[]{0.47\textwidth}
\centering
    \includegraphics[width=\textwidth]{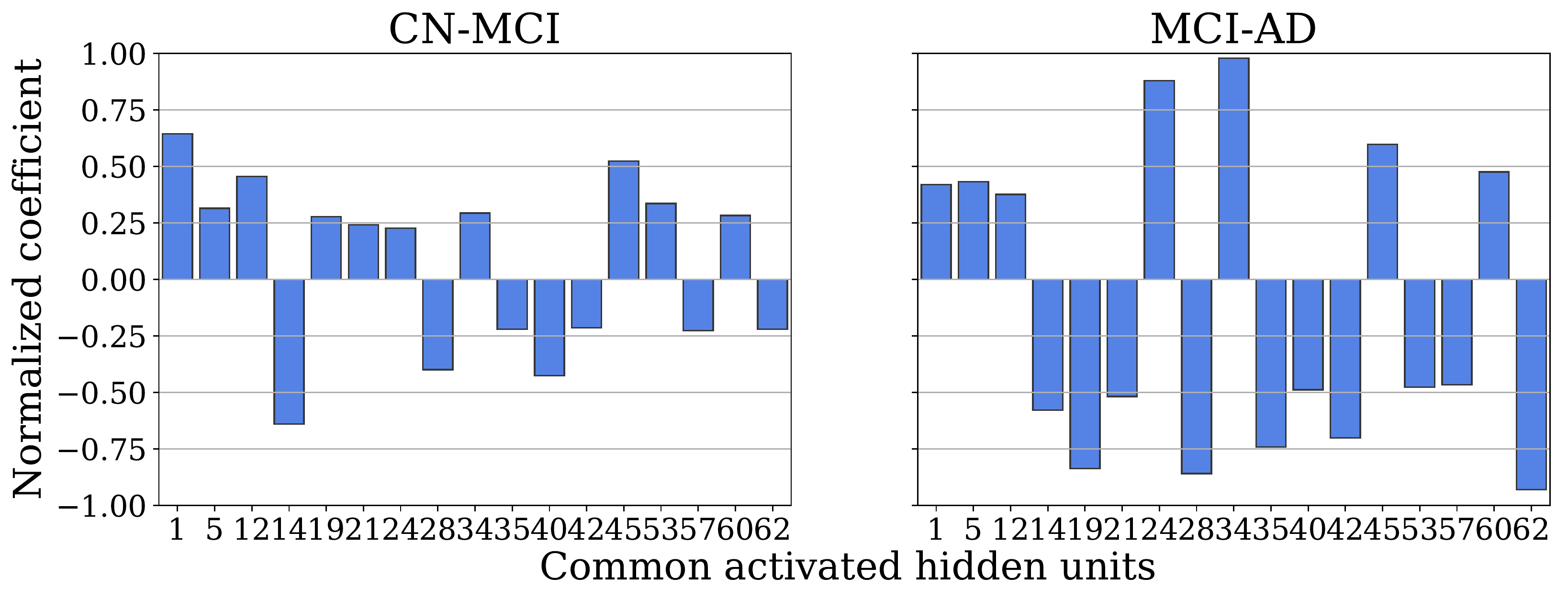}
    \caption{}
\end{subfigure}
\caption{Normalized point biserial correlation coefficients for subjects with conversions, \ie, CN-MCI and MCI-AD, during a period of 10 years from the baseline. Each panel represents the potential cell units for `\emph{model biomarkers},' whose activations were highly correlated with the conversions.}
\label{fig:recurrent_analysis_cog}
\end{figure}

\subsection{Interpreting Cell States}
\label{Deep Recurrent Model Analysis}
While our deep recurrent network achieved relatively superior performance in both MRI biomarker and clinical test scores forecasting, as well as clinical status prediction, it is desirable to understand the dynamics of the internal state representations of our proposed model. In this regard, we devised a novel analysis method of the cell states in our LSTM. To determine which cell states contributed to the identification of the clinical statuses, we estimated the correlation between a patient's disease statuses over time and the corresponding cell state representations. Specifically, we first collected the internal cell representations of $\approx 37$ subjects $(37.0\pm8.8)$\footnote{This value represented the mean and standard deviation over the number of subjects in 5-fold cross-validation.}, who experienced status-conversion during our period of interest, and then measured the correlations between the change of the values in the cell states and a transition of the patient's disease status (0: no conversion and 1: conversion) with point-biserial correlation coefficients.
We grouped all subjects depending on the disease transitions as follows: (i) $\approx 7$ subjects {$(7.8\pm1.7)$} showed transitions from CN to MCI (CN-MCI), (ii) $\approx 26$ subjects {$(26.8\pm8.9)$} showed transitions from MCI to AD (MCI-AD), and (iii) $\approx 2$ subjects {$(2.4\pm1.4)$} showed transitions through all the statuses of CN, MCI, and AD (CN-MCI-AD). Then, we analyzed the calculated point-biserial correlations for CN-MCI and MCI-AD~\footnote{We excluded the CN-MCI-AD group for simplicity in interpretation.}.

Regarding the variation of conversion time among subjects per group, we normalized the point-biserial correlation coefficients with the standard deviation of the coefficients in each group. Then, we sorted the normalized coefficients to identify the significant cell states related to the transitions, as presented in Fig. \ref{fig:cell_state_group}. From the sorted cells, we focused on cell units with the top $25\%$ most extreme coefficients per group. Of the selected cell units, the common $\approx 16$ cell units $(16.6\pm2.7)$ in both groups are presented in Fig.~\ref{fig:recurrent_analysis_cog}, where the subfigures correspond to one fold of our five-fold cross-validation experiments. As these cell states showed high correlations either positively or negatively, we believe that those cell units could be potentially used as `\emph{model biomarkers}' with respect to the clinical status conversion. Interestingly, across the five-fold experiments, a similar number (13$\sim$21) of cell units showed high correlation with the clinical status change. From this observation, we infer the robustness or generalizability of our method over different training sets.

\begin{figure}[h]
\centering
\begin{subfigure}[]{0.48\textwidth}
\centering
    \includegraphics[width=\textwidth]{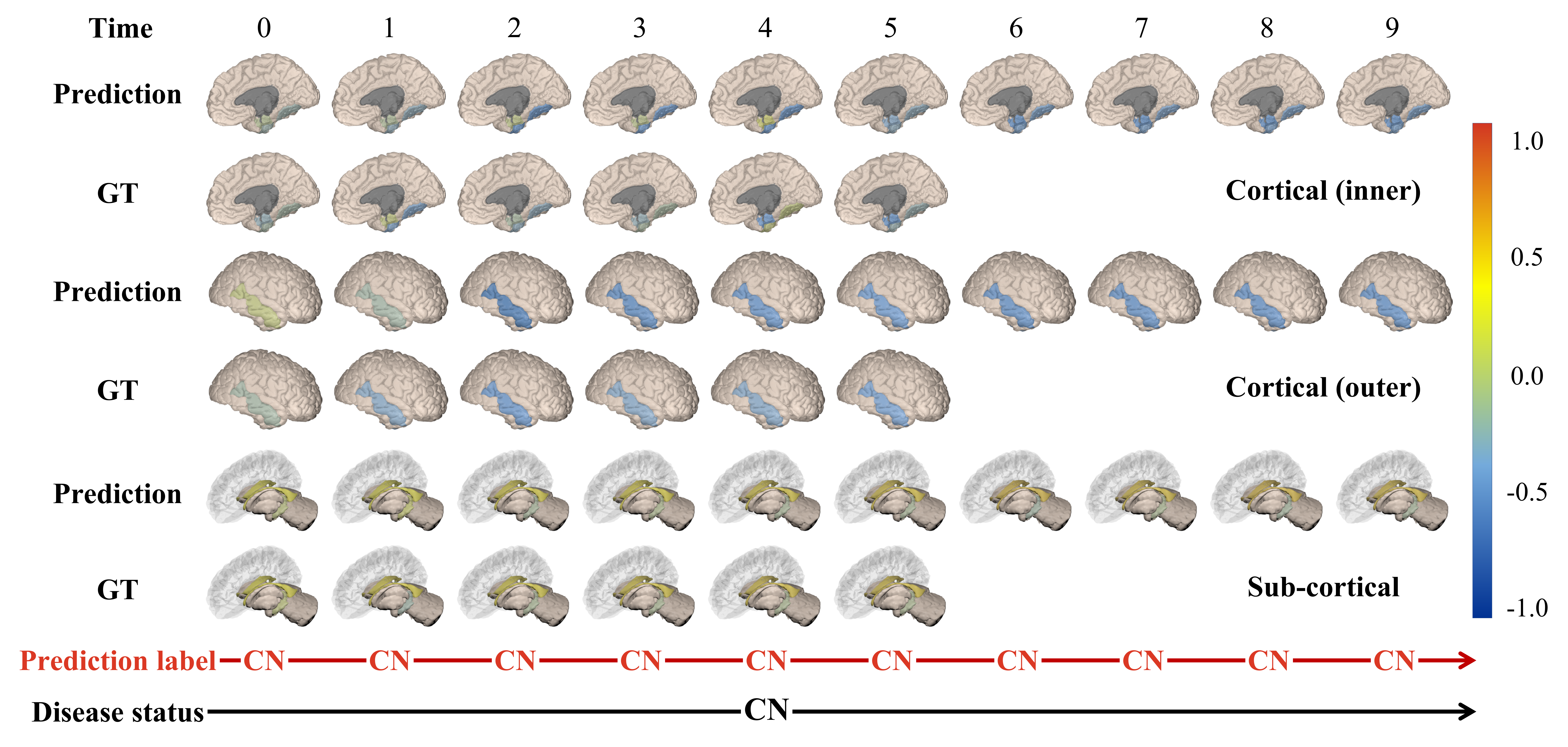}
    \caption{CN}
    \label{fig:Volume_CN}
\end{subfigure}
\begin{subfigure}[]{0.49\textwidth}
\centering
    \includegraphics[width=\textwidth]{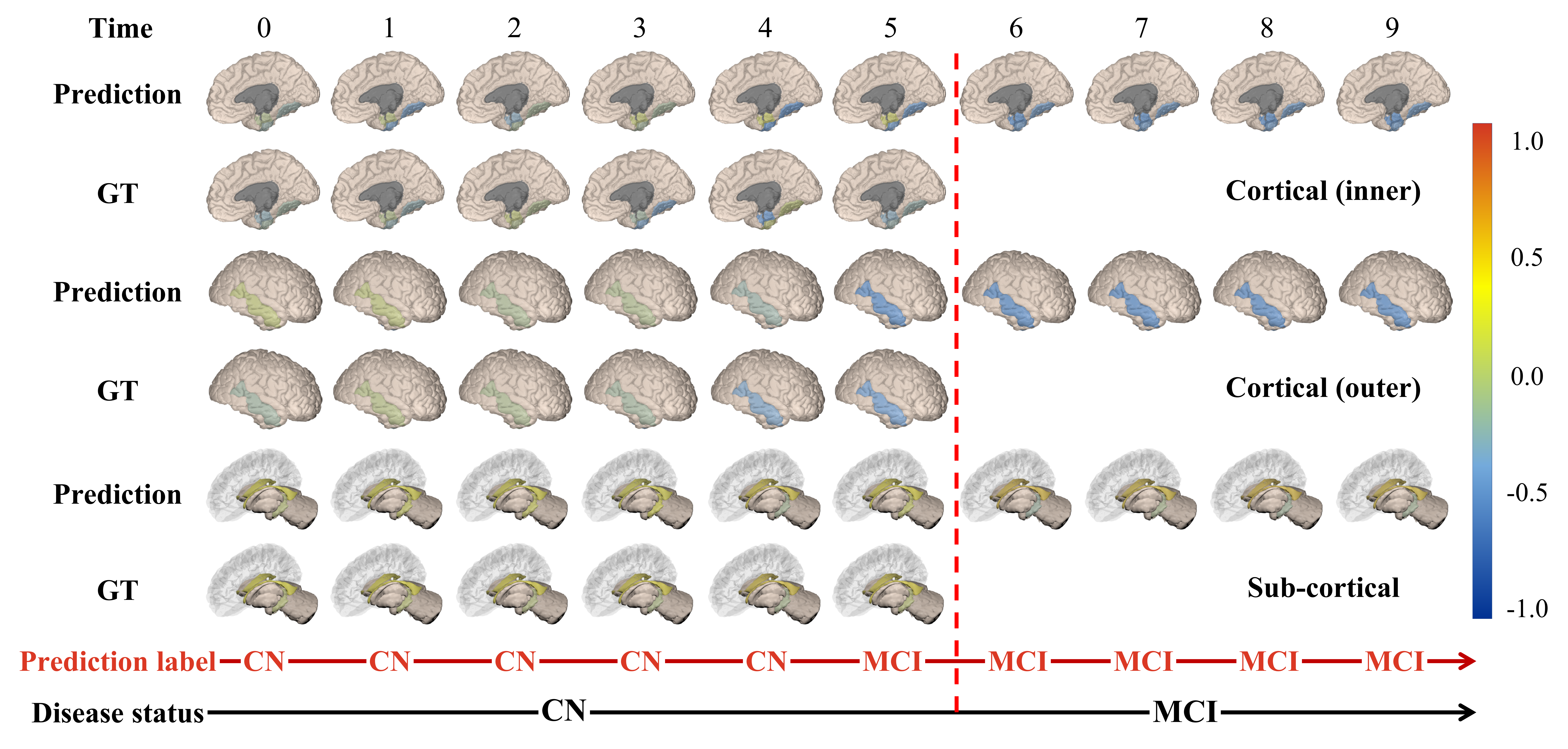}
    \caption{CN-MCI}
    \label{fig:Volume_CN_MCI}
\end{subfigure}
\begin{subfigure}[]{0.48\textwidth}
\centering
    \includegraphics[width=\textwidth]{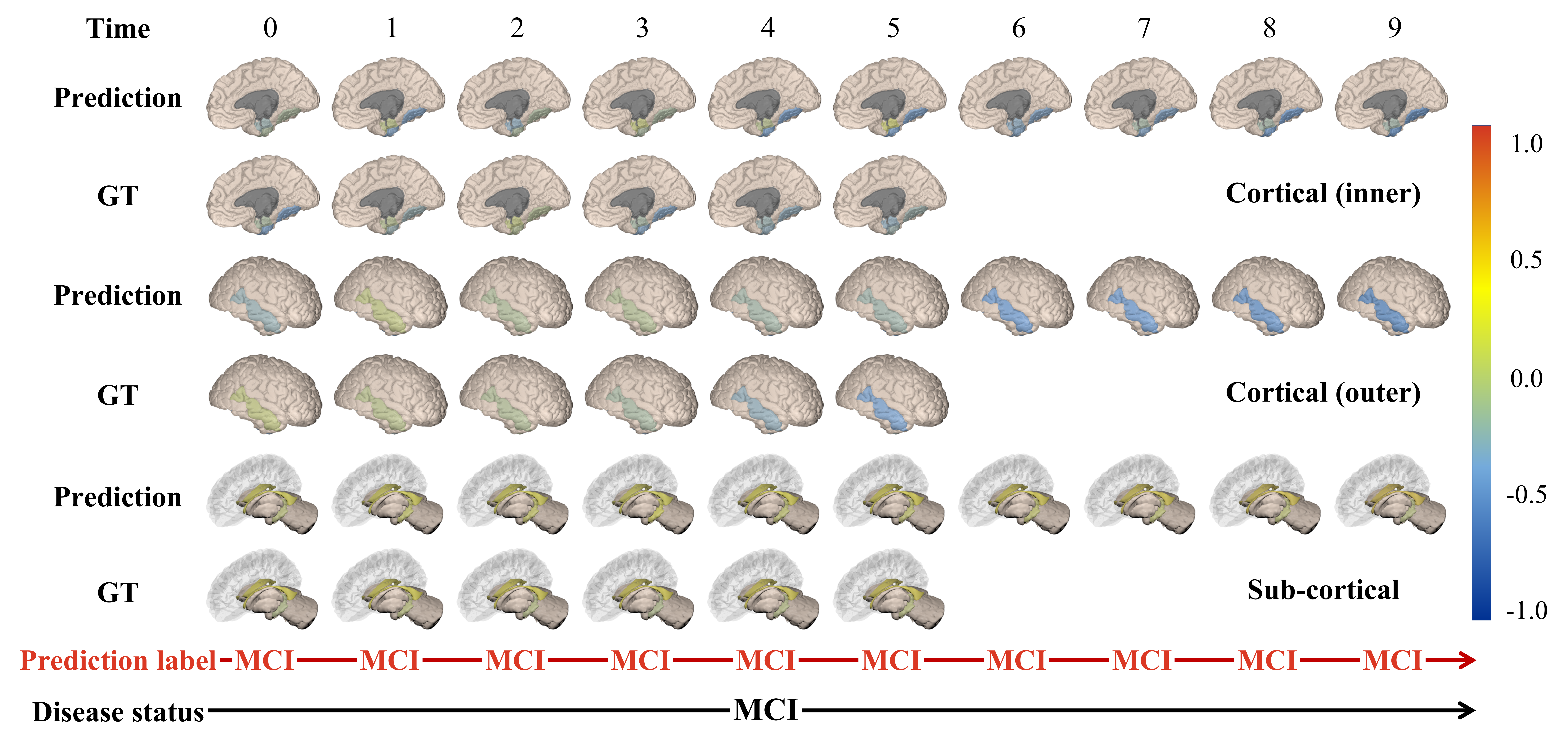}    \caption{MCI}
    \label{fig:Volume_MCI}
\end{subfigure}
\begin{subfigure}[]{0.49\textwidth}
\centering
    \includegraphics[width=\textwidth]{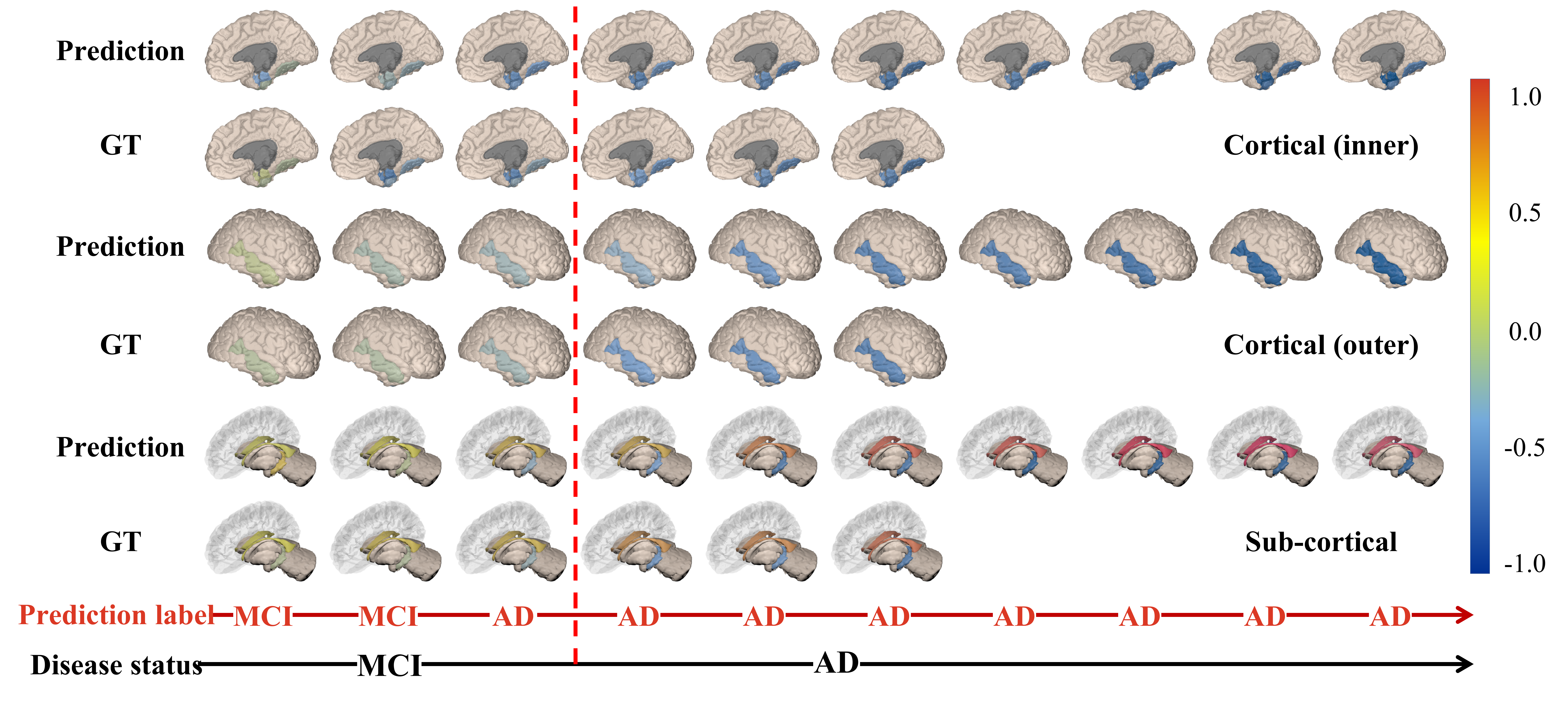}
    \caption{MCI-AD}
    \label{fig:Volume_MCI_AD}
\end{subfigure}
\begin{subfigure}[]{0.48\textwidth}
\centering
    \includegraphics[width=\textwidth]{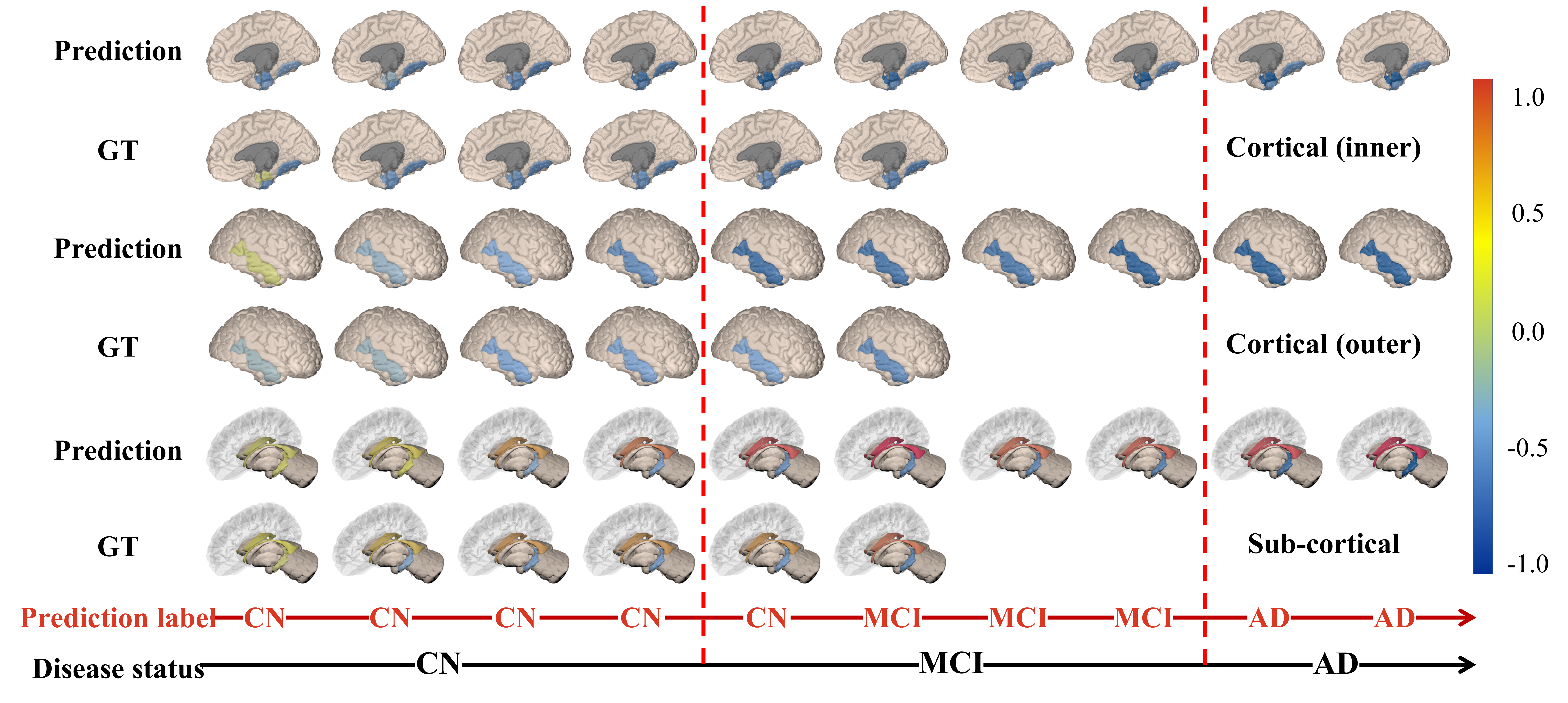}
    \caption{CN-MCI-AD}
    \label{fig:Volume_CN_MCI_AD}
\end{subfigure}
\begin{subfigure}[]{0.49\textwidth}
\centering
    \includegraphics[width=\textwidth]{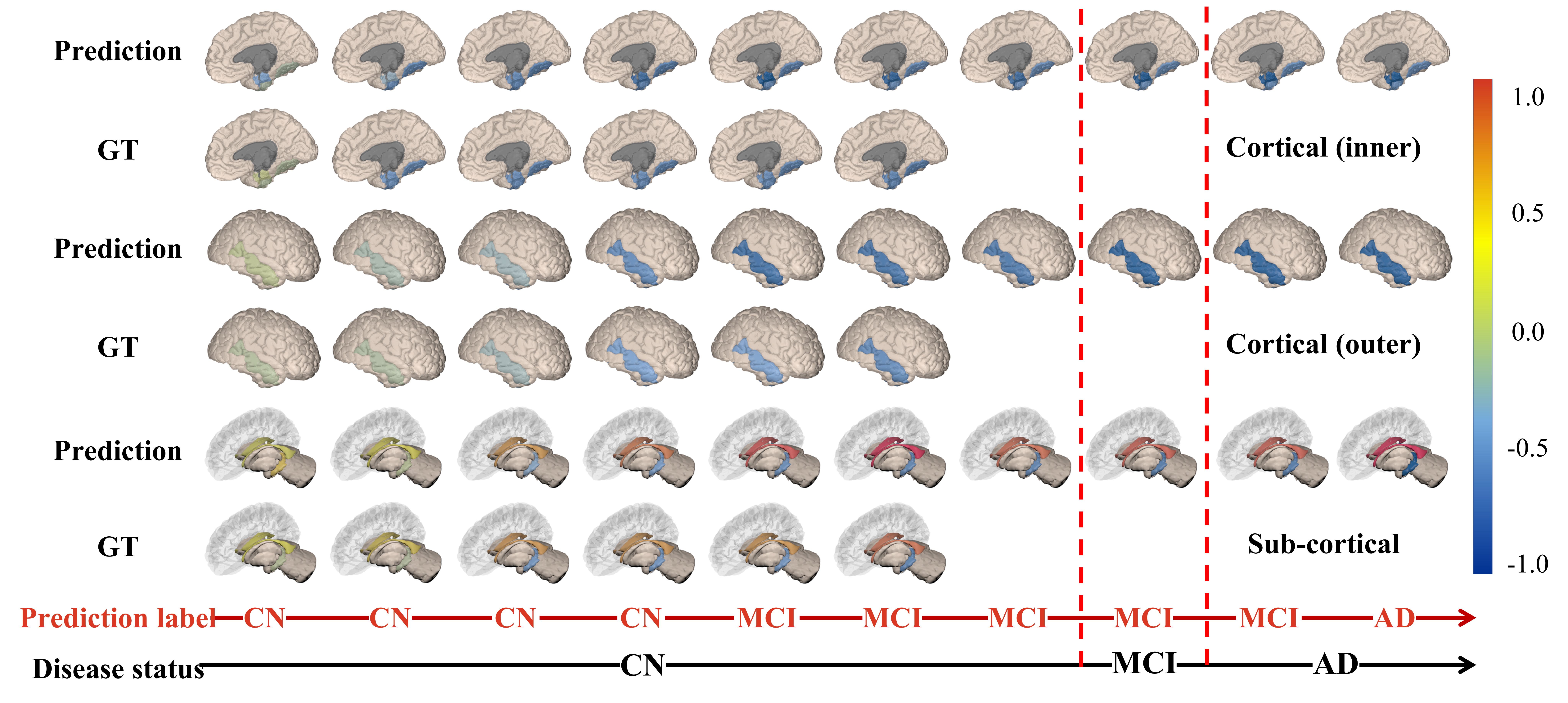}
    \caption{CN-MCI-AD}
    \label{fig:Volume_CN_MCI_AD_2}
\end{subfigure}
\caption{Examples of individual trajectories of MRI biomarkers are the amount of change over time. The color-coded values indicate the normalized relative changes from the respective regions' values at baseline. (GT: ground truth)}
\label{fig:Volume_change_figure}
\end{figure}

\subsection{Individual MRI Biomarker Value Changes over Time}
\label{indi_mri_biomarker_trajectories}
{\ws In regard to the relationship between the MRI biomarker values and the clinical statuses over time, we visualized the trajectories of MRI biomarker change and the corresponding clinical status in Fig. \ref{fig:Volume_change_figure}. We computed the relative changes in the prediction ($p$) and the ground truth ($g$) from the respective regions' values to the baseline as follows:
\begin{equation}
    r_{g,t}^{(i)} = \frac{y_{t}^{(i)}-y_{0}^{(i)}}{y_{0}^{(i)}}, \quad \quad 
    r_{p,t}^{(i)} = \frac{\hat{y}_{t}^{(i)}-y_{0}^{(i)}}{y_{0}^{(i)}} \nonumber
\end{equation}
where $i$ and $t$ denote the indices of MRI biomarkers and time points, respectively. Those relative changes were then normalized with the minimum and maximum values such that the resulting values were in the range of $[-1,1]$, in which the positive and negative values denote increase and decrease, respectively. As a result, comparing our predictions with the ground truths, they showed very similar values over time and across cases. These experimental results validated the our proposed method's ability to forecast the longitudinal MRI biomarkers. Second, unlike the two examples of Fig. \ref{fig:Volume_CN} and Fig. \ref{fig:Volume_MCI}, which presents stable or slowly deteriorating status over the period, the regression to other statuses in the AD spectrum (Fig. \ref{fig:Volume_MCI_AD} and Fig. \ref{fig:Volume_CN_MCI_AD}) showed relatively rapid changes in volumetric measurements, \ie, enlarged ventricles and shrunken other (sub)cortical regions.}

\begin{figure}[h]
\begin{subfigure}[]{0.49\textwidth}
\centering
    \includegraphics[width=\textwidth]{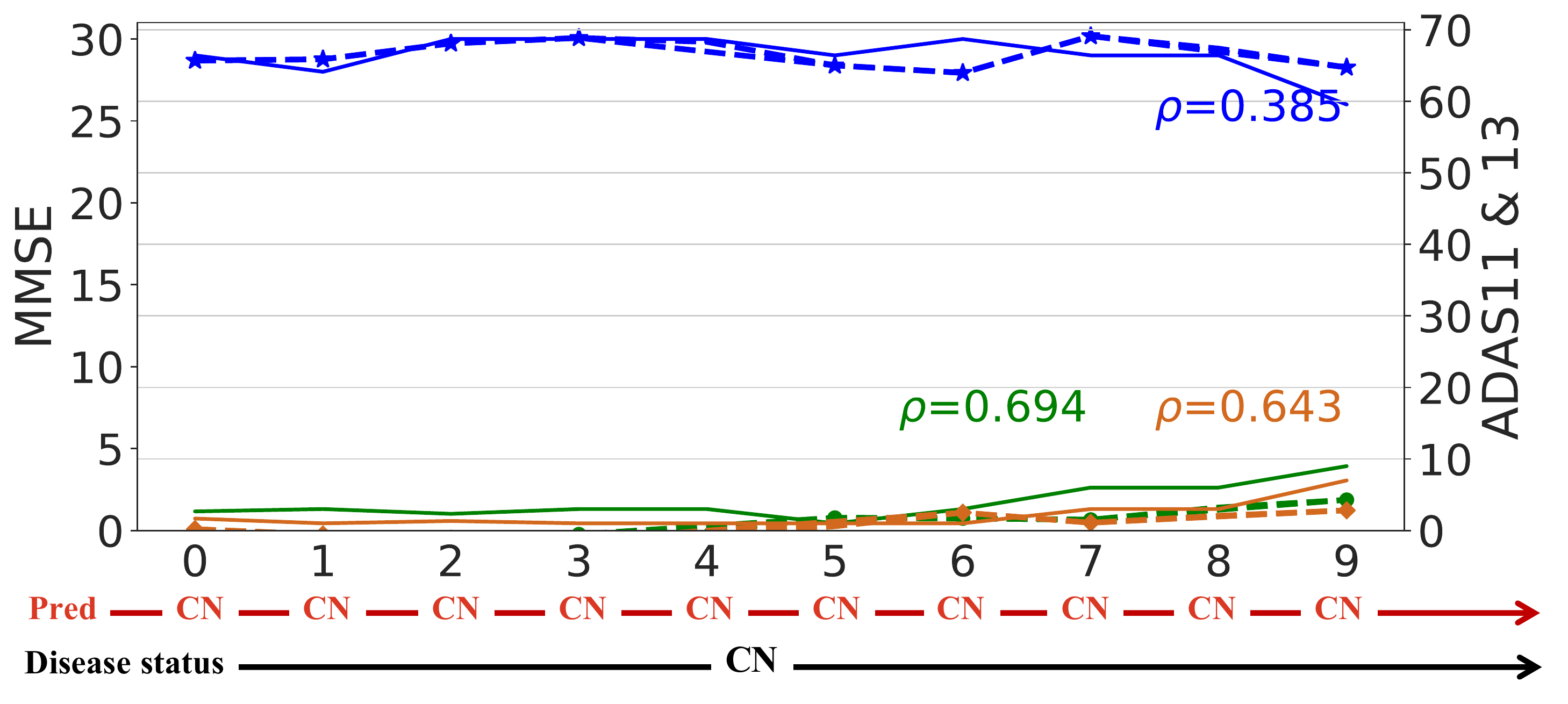}
    \caption{CN}
    \label{fig:cognitive_trajectories_CN}
\end{subfigure}
\begin{subfigure}[]{0.49\textwidth}
\centering
    \includegraphics[width=\textwidth]{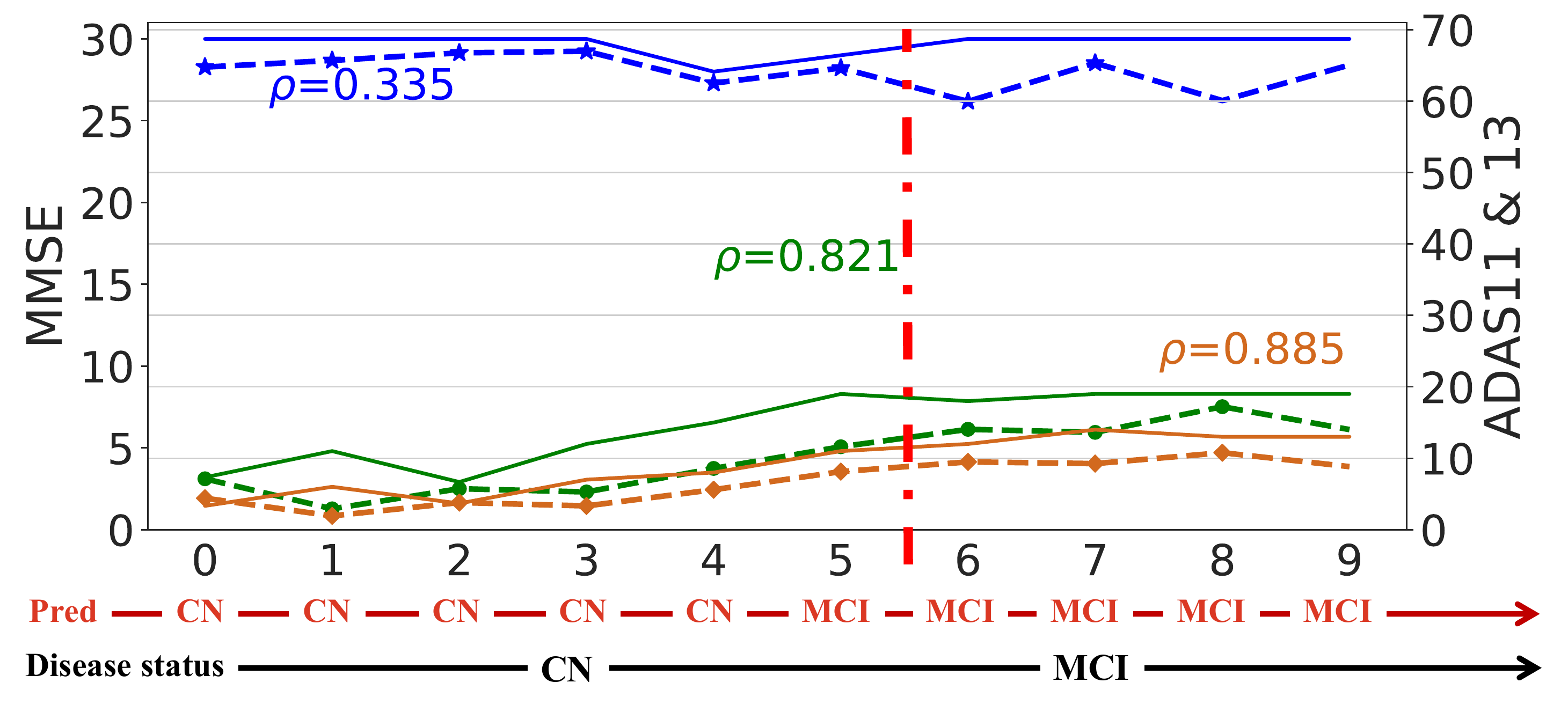}
    \caption{CN-MCI}
    \label{fig:cognitive_trajectories_CN_MCI}
\end{subfigure}
\begin{subfigure}[]{0.49\textwidth}
\centering
    \includegraphics[width=\textwidth]{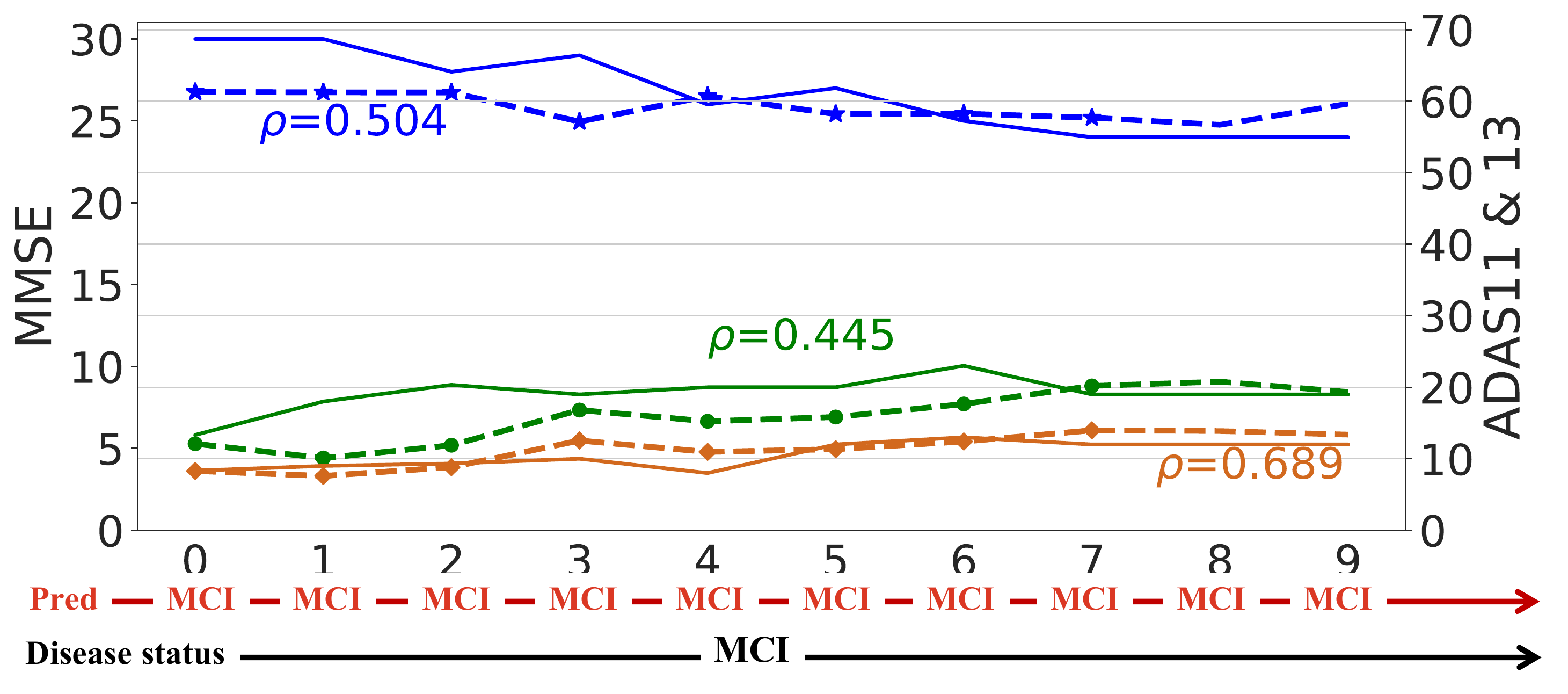}
    \caption{MCI}
    \label{fig:cognitive_trajectories_MCI}
\end{subfigure}
\begin{subfigure}[]{0.49\textwidth}
\centering
    \includegraphics[width=\textwidth]{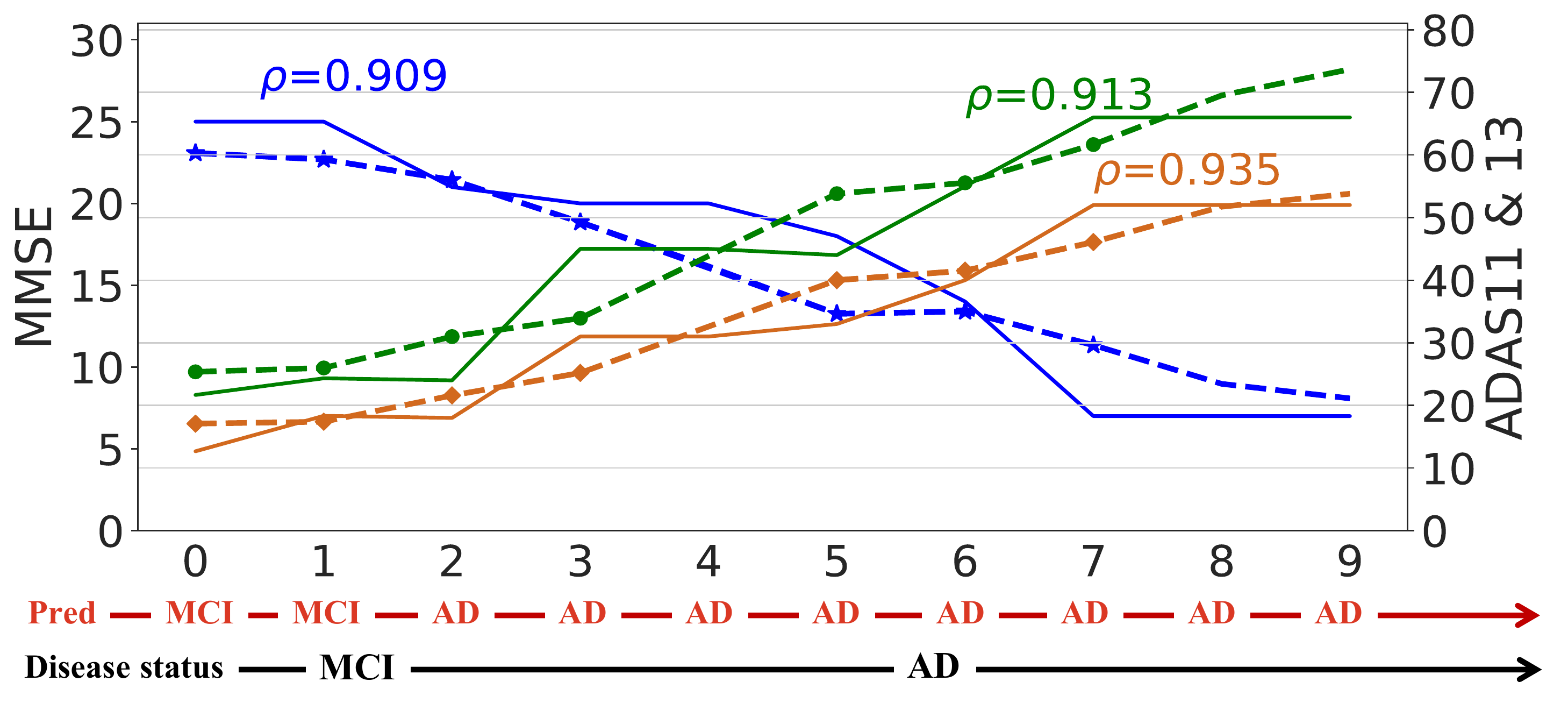}
    \caption{MCI-AD}
    \label{fig:cognitive_trajectories_MCI_AD}
\end{subfigure}
\begin{subfigure}[]{0.49\textwidth}
\centering
    \includegraphics[width=\textwidth]{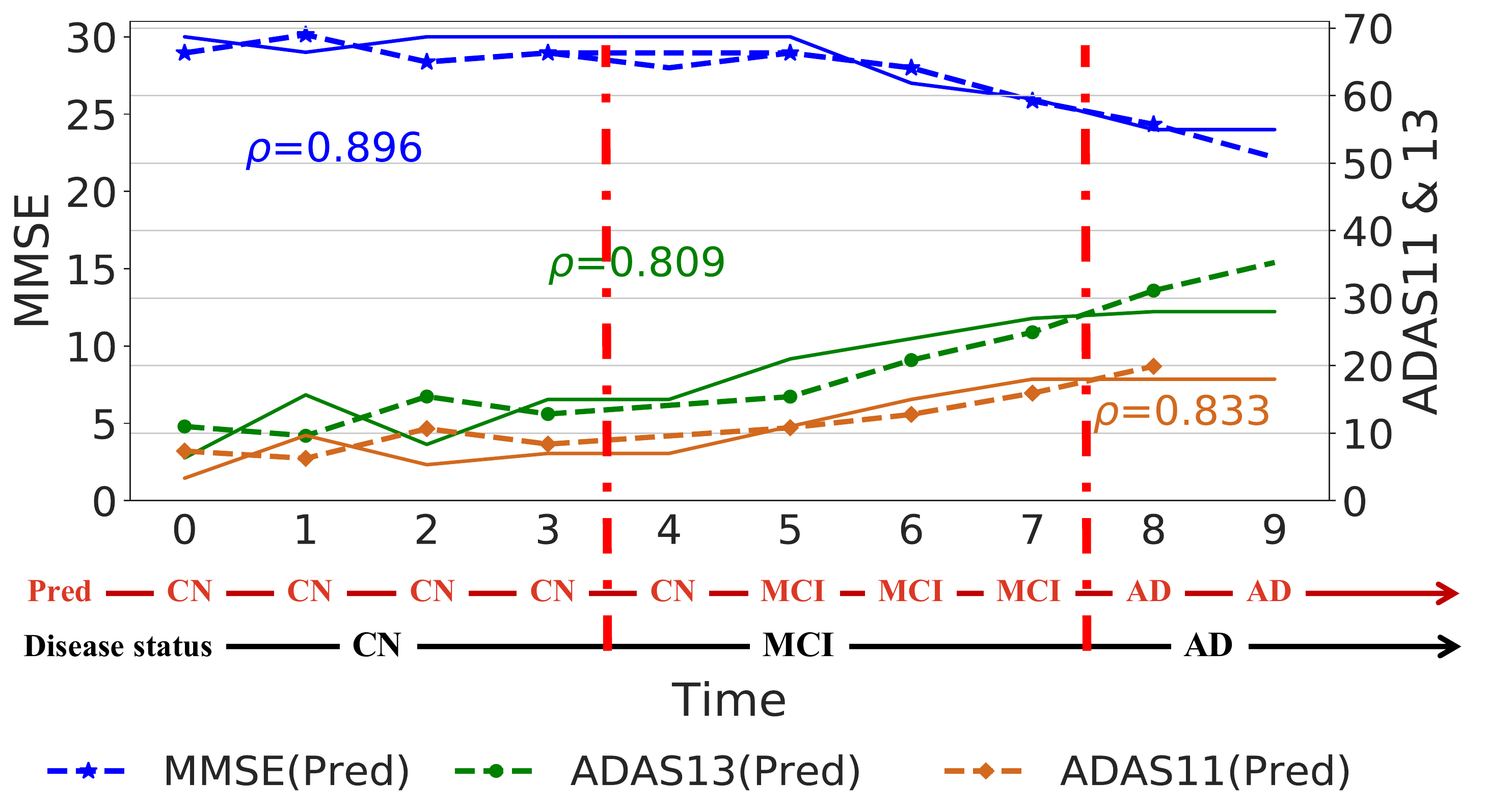}
    \caption{CN-MCI-AD}
    \label{fig:cognitive_trajectories_CN_MCI_AD}
\end{subfigure}
\begin{subfigure}[]{0.49\textwidth}
\centering
    \includegraphics[width=\textwidth]{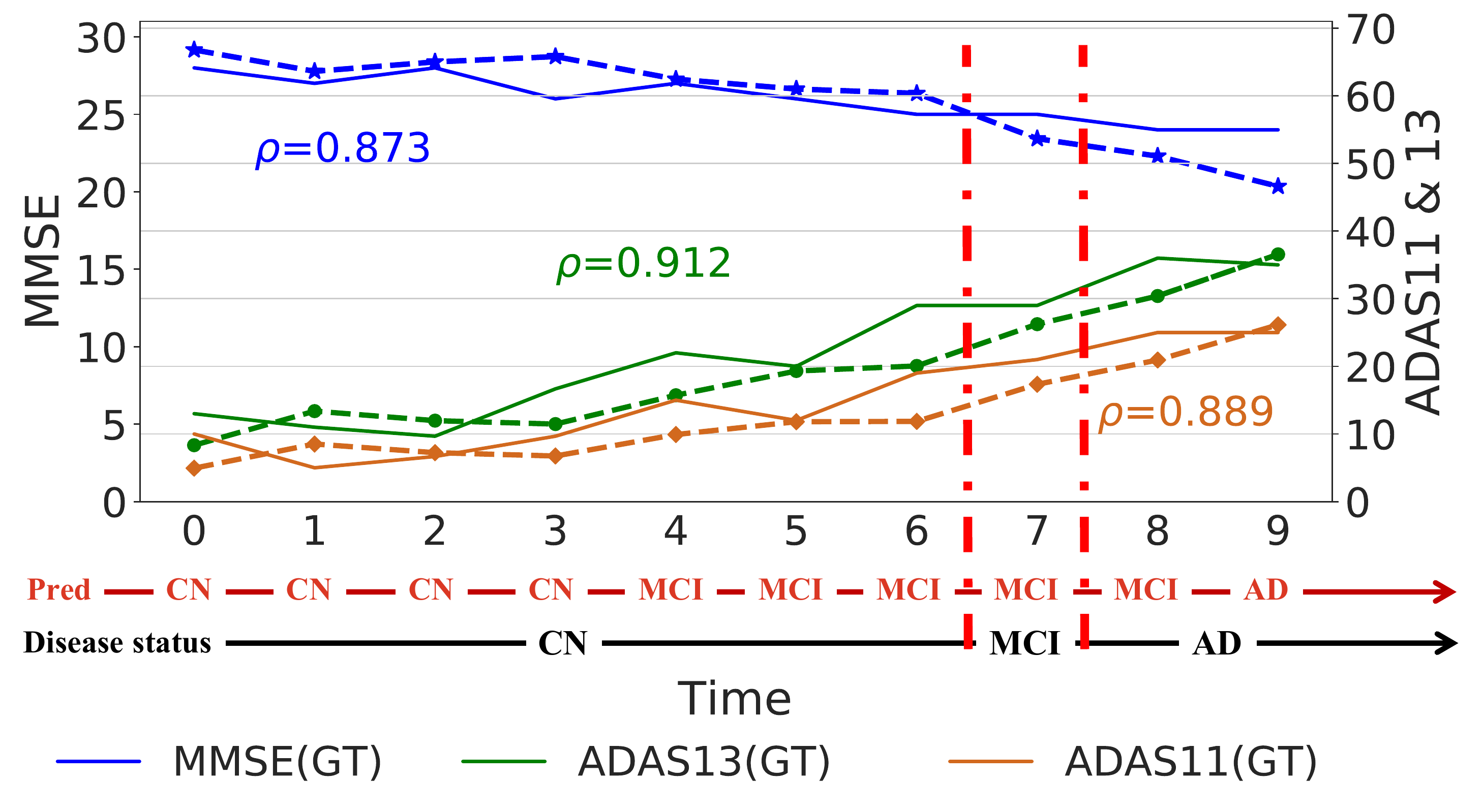}
    \caption{CN-MCI-AD}
    \label{fig:cognitive_trajectories_CN_MCI_AD_2}
\end{subfigure}
\caption{Examples of individual trajectories of the cognitive tests represents the change of scores over time. Blue, magenta, and chocolate denote MMSE, ADAS-cog11, and ADAS-cog13, respectively. ($\rho$: correlation coefficient, Pred: prediction, GT: ground truth)}
\label{fig:cognitive_trajectories_individual}
\end{figure}

\subsection{Individual Cognitive Test Scores Changes over Time}
{\ws Similar to Section~\ref{indi_mri_biomarker_trajectories}, in regard to the relationship between the cognitive test scores and the clinical statuses over time, in Fig. \ref{fig:cognitive_trajectories_individual}, we visualized the trajectories of the cognitive test scores change and the corresponding clinical status for the same subjects in Fig.~\ref{fig:Prognosis_result_individual} and Fig.~\ref{fig:Volume_change_figure}.
Note that, in the figure, the cognitive test scores were normalized to $[0,1]$ in the procedure of model training, and we transformed the predicted scores to their original ranges by multiplying max values of the respective cognitive tests \ie, 30 (MMSE), 70 (ADAS-cog11), and 85 (ADAS-cog13).
In comparisons between estimated cognitive scores and the ground truth, the trajectories of the respective cognitive test scores followed the tendency of the ground truth over time and across cases. In addition, we calculated the Pearson correlation between estimated values and ground truth, denoted as $\rho$ in the figure. All cases achieved higher correlation values ($\rho >0.3)$ in Fig.~\ref{fig:cognitive_trajectories_individual}.
Specifically, we observed high $\rho$ values $(\rho>0.8)$ for disease state changes except for CN-MCI in Fig.~\ref{fig:cognitive_trajectories_CN_MCI} showing relatively low MMSE. In addition, in contrast to the four examples of Fig.~\ref{fig:cognitive_trajectories_CN}, Fig.~\ref{fig:cognitive_trajectories_CN_MCI}, and Fig.~\ref{fig:cognitive_trajectories_MCI}, which showed either relatively stable or slowly decreasing scores over time, the progression to other statuses in the AD spectrum (Fig.~\ref{fig:cognitive_trajectories_MCI_AD}, Fig.~\ref{fig:cognitive_trajectories_CN_MCI_AD}, and Fig.~\ref{fig:cognitive_trajectories_CN_MCI_AD_2}) represented either dramatically decrease or increase in scores, \ie, decrease in MMSE scores and increase in ADAS-cog11 and ADAS-cog13.
As a result, we concluded that our proposed method is capable in capturing the trajectories of the cognitive test scores either stable and slowly degrading the status of the disease or severely deteriorating status.}

\section{Conclusion}
Progressive neurodegenerative AD is of great concern around the world due to its clinical, social, and economic impacts. As there is no treatment or medicine to cure or reverse the progression of AD itself, many researchers in various fields have been devoting their efforts to early diagnosis or prognosis in diverse ways. In particular, the use of neuroimaging data is one of the main areas of study, as it provides quantitative and qualitative measurements on which we can build computational models to predict the risk of AD progression.

Of various strategies for accurate AD diagnosis or prognosis, recent studies on longitudinal datasets validated the significance of using historical observations. In that regard, we formulated the problem of AD progression prognosis, \ie, the tasks of both MRI biomarker and cognitive test score forecasting and clinical status prediction over multiple time points, by means of the disease progression modeling. From a methodological standpoint, we proposed a deep recurrent neural network that is jointly and systematically capable of imputing missing observations, encoding historical inputs into hidden state vector representations, and predicting future MRI biomarker values, cognitive test scores and the respective clinical status. 
We have performed exhaustive experiments and analyses by comparing with other comparative methods in the literature. First, our proposed method was superior to those competing methods in various quantitative metrics. Second, our method resulted in more stable (low-bias and low-variance) outputs in MRI biomarker forecasting and more reasonable clinical status prediction by reflecting the characteristics of irreversible AD (\ie, no status-reverse prediction was made over time). Third, our method represented the reliable trajectories of the cognitive test scores. Lastly, we observed that some internal state vector representations, \ie, the cell states in our LSTM, showed high correlation with the clinical state-conversion, implying the potential use of those as `\emph{model biomarkers}' in predicting clinical status conversion in the near future.

As we considered the six MRI biomarkers and three cognitive test scores in this work, our forthcoming research direction will be to accommodate the MRI biomarkers of the whole brain and to integrate functional and/or genotypic information in multi-modal learning. Further, in order to get new insights about the AD progression, it is also important to adopt interpretable and explainable mechanisms in the model.

\section*{Acknowledgement} 
This work was supported by Institute of Information \& communications Technology Planning \& Evaluation (IITP) grant funded by the Korea government (MSIT) (No. 2019-0-00079, Artificial Intelligence Graduate School Program (Korea University)) and the National Research Foundation of Korea (NRF) grant funded by the Korea government (MSIT) (No. 2019R1A2C1006543).

\section*{Code and Data Availability} 
All codes used in our experiments are available at {`\url{https://github.com/ssikjeong1/Deep_Recurrent_AD}'}.
We used the TADPOLE longitudinal cohort (\url{https://tadpole.grand-challenge.org/Data/}) from the publicly available ADNI database, including ADNI-1, ADNI-2 and ADNI-GO.

\section*{Declaration of Competing Interest}
The authors declare that they have no conflict of interest.

\bibliographystyle{elsarticle-harv}

\bibliography{main}


\appendix
\setcounter{figure}{0}
\setcounter{table}{0}

\clearpage 
\section{Summary of the TADPOLE dataset}
\begin{table}[h]
\caption{Demographic information of subjects in the TADPOLE dataset. Values are reported as a mean $\pm$ standard deviation. Note that the number of subjects is measured at the baseline time point and includes missing observation values and clinical statuses during the subject's visits.}
\label{appendix:tab_demographic}
\renewcommand*{\arraystretch}{0.65}
\centering\scriptsize{\begin{tabular}{lccccccc}
\toprule
\multirow{2}{*}{Clinical Status}&\multicolumn{2}{c}{$\#$ of subjects}&\multicolumn{2}{c}{$\#$ of visits}&\multicolumn{2}{c}{Age (mean $\pm$ std.)}\\
&{Male}&{Female}&{Male} & {Female} & {Male} & {Female} \\\toprule
\multicolumn{1}{l}{Cognitively Normal (CN)}&\multicolumn{1}{c}{$253$}&\multicolumn{1}{c}{$270$}&\multicolumn{1}{c}{$2,126$} & \multicolumn{1}{c}{$2,084$} & \multicolumn{1}{c}{$75.10\pm5.64$} & \multicolumn{1}{c}{$74.42\pm5.38$} \\\midrule
\multicolumn{1}{l}{Mild Cognitive Impairment (MCI)}&\multicolumn{1}{c}{$515$}&\multicolumn{1}{c}{$357$}&\multicolumn{1}{c}{$4,207$} & \multicolumn{1}{c}{$2,756$}&\multicolumn{1}{c}{$73.71\pm7.11$}&\multicolumn{1}{c}{$71.80\pm7.85$} \\\midrule
\multicolumn{1}{l}{Alzheimer's Disease (AD)}&\multicolumn{1}{c}{$189$}&\multicolumn{1}{c}{$153$}&\multicolumn{1}{c}{$884$}&\multicolumn{1}{c}{$684$}&\multicolumn{1}{c}{$75.56\pm7.11$}&\multicolumn{1}{c}{$74.52\pm7.81$} \\\midrule
\multicolumn{1}{l}{All}&\multicolumn{2}{c}{$1,737$}&\multicolumn{2}{c}{$12,741$} & \multicolumn{2}{c}{$73.81\pm6.97$} \\ \bottomrule
\end{tabular}}
\end{table}

\section{Anatomical regions of the 6 MRI Biomarkers}
\begin{figure}[h]
\centering
 \includegraphics[width=\textwidth]{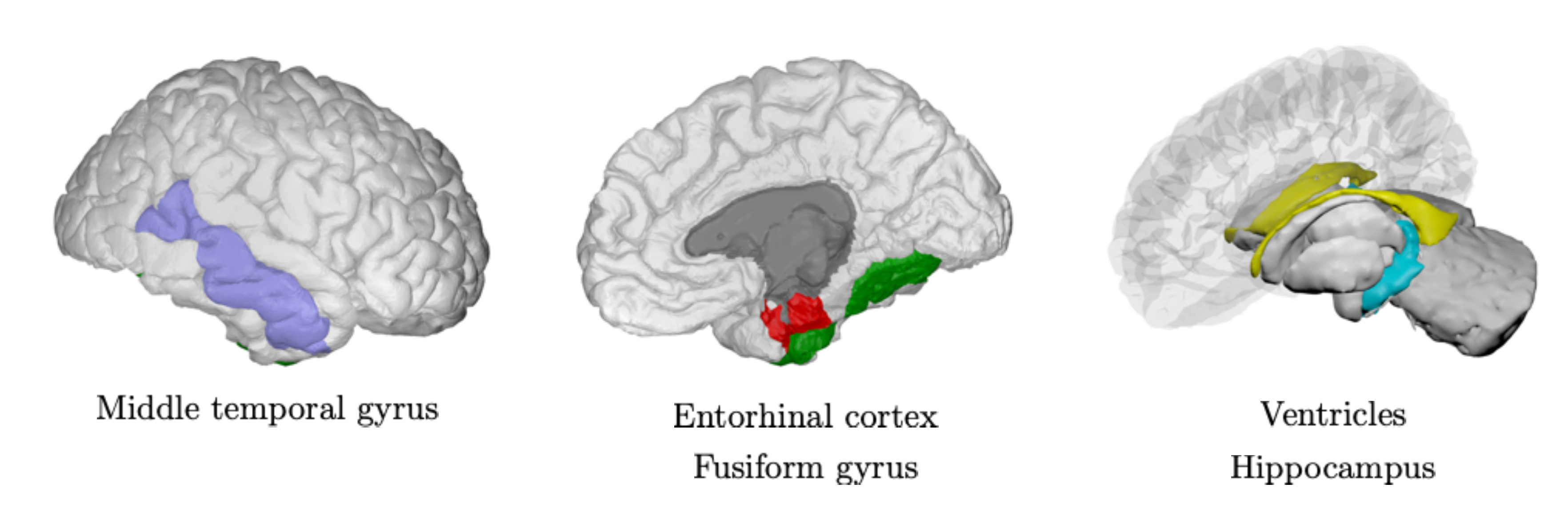}
\caption{Visualization of six MRI volumetric biomarkers that are used in our experiments. This figure was generated by the BrainPainter~\citep{marinescu2019brain}.}
\label{fig:mri_biomarkers}
\end{figure}

\clearpage

\section{Statistics of the Demographic and Cognitive Scores}
\begin{table}[h]
\caption{Statistical information on a demographic and cognitive score for each group every year. Scores on the CDR-SOB, ADAS-cog11, and ADAS-cog13 were higher at severe levels of cognitive dysfunction. Conversely, scores on the MMSE were lower at serious levels of cognitive dysfunction.}
\label{append:demograpic_info}
\centering\scriptsize{
{\begin{threeparttable}
\renewcommand*{\arraystretch}{0.65}
\begin{tabular}{cccccccc}
\toprule
Year               & Groups & Number (Gender)  & Age           & MMSE          & ADAS-cog11    & ADAS-cog13    & CDR-SOB  \\\midrule
\multirow{3}{*}{Baseline}  & CN     & 226 (M/F: 108/118)       & 74.0$\pm$5.2  & 29.0$\pm$1.1  & 5.7$\pm$2.9   & 8.8$\pm$4.1   & 0.0$\pm$0.1 \\ 
                    & MCI    & 334 (M/F: 289/194)       & 72.9$\pm$7.5  & 27.4$\pm$1.8  & 10.6$\pm$4.7   & 17.2$\pm$6.8  & 1.5$\pm$0.8 \\ 
                    & AD     & 131 (M/F: 63/68)           & 73.7$\pm$7.6  & 23.2$\pm$1.9  & 18.7$\pm$6.8  & 28.9$\pm$8.1  & 4.2$\pm$1.6 \\\midrule
\multirow{3}{*}{1} & CN     & 208 (M/F: 103/105)       & 73.9$\pm$5.3  & 29.0$\pm$1.4  & 5.2$\pm$2.9   & 8.1$\pm$4.3   & 0.1$\pm$0.3 \\  
                    & MCI    & 275 (M/F: 166/109)       & 73.5$\pm$7.5  & 27.3$\pm$2.3  & 10.6$\pm$5.2   & 17.0$\pm$7.7  & 1.7$\pm$1.1 \\
                    & AD     & 177 (M/F: 87/90)          & 73.0$\pm$7.7  & 21.9$\pm$4.0  & 21.3$\pm$8.8  & 31.9$\pm$10.3  & 5.1$\pm$2.4 \\ \midrule 
\multirow{3}{*}{2} & CN     & 203 (M/F: 105/98)       & 73.8$\pm$5.5  & 29.2$\pm$1.0  & 5.0$\pm$2.7   & 7.9$\pm$4.0   & 0.1$\pm$0.4 \\ 
                    & MCI    & 200 (M/F: 116/84)       & 73.1$\pm$7.3  & 27.5$\pm$2.2  & 10.1$\pm$4.9   & 16.3$\pm$7.4  & 1.7$\pm$1.1 \\ 
                    & AD     & 178 (M/F: 91/87)         & 73.3$\pm$7.7  & 20.8$\pm$5.0  & 23.4$\pm$10.6  & 34.0$\pm$11.5  & 6.1$\pm$3.1 \\\midrule 
\multirow{3}{*}{3} & CN     & 126 (M/F: 64/62)         & 73.7$\pm$5.3  & 29.1$\pm$1.3  & 4.7$\pm$2.4   & 7.7$\pm$3.6   & 0.1$\pm$0.3 \\
                    & MCI    & 156 (M/F: 95/61)       & 73.0$\pm$7.4  & 27.5$\pm$2.4  & 9.4$\pm$5.0   & 15.4$\pm$7.3  & 1.6$\pm$1.4 \\
                    & AD     & 95 (M/F: 48/47)         & 73.1$\pm$6.8  & 20.9$\pm$5.4  & 22.6$\pm$10.1 & 31.9$\pm$9.2 & 6.2$\pm$3.3 \\\midrule 
\multirow{3}{*}{4} & CN     & 103 (M/F: 53/50)         & 73.9$\pm$5.8  & 29.2$\pm$1.1  & 5.7$\pm$2.6   & 8.5$\pm$3.7   & 0.1$\pm$0.3 \\ 
                    & MCI    & 99 (M/F: 64/35)        & 72.6$\pm$7.3  & 27.9$\pm$1.9  & 8.7$\pm$3.7   & 14.1$\pm$5.9  & 1.4$\pm$1.2 \\ 
                    & AD     & 70 (M/F: 39/31)         & 72.3$\pm$6.7  & 20.1$\pm$5.2  & 24.0$\pm$11.4 & 34.8$\pm$13.0 & 6.4$\pm$2.9 \\\midrule
\multirow{3}{*}{5} & CN     & 68 (M/F: 33/35)          & 74.7$\pm$5.5  & 29.0$\pm$1.3  & 5.8$\pm$3.0   & 9.0$\pm$4.3   & 0.1$\pm$0.3 \\
                    & MCI    & 67 (M/F: 43/24)         & 72.5$\pm$7.0  & 27.5$\pm$2.2  & 9.2$\pm$3.9   & 15.2$\pm$6.5  & 1.6$\pm$1.2 \\ 
                    & AD     & 47 (M/F: 24/23)          & 71.8$\pm$7.0  & 20.3$\pm$5.4  & 24.8$\pm$11.1 & 35.7$\pm$13.2 & 7.0$\pm$3.1 \\ \midrule  
\multirow{3}{*}{6} & CN     & 64 (M/F: 32/32)          & 75.0$\pm$4.7  & 28.9$\pm$1.5  & 5.7$\pm$2.9   & 8.9$\pm$4.1   & 0.2$\pm$0.3 \\ 
                    & MCI    & 56 (M/F: 40/16)          & 74.4$\pm$6.7  & 27.5$\pm$2.1  & 9.9$\pm$4.9   & 15.9$\pm$7.2  & 1.7$\pm$1.5 \\
                    & AD     & 42 (M/F: 22/20)          & 72.6$\pm$6.8  & 18.2$\pm$6.3  & 26.9$\pm$13.9 & 38.5$\pm$15.7 & 8.1$\pm$3.9 \\\midrule 
\multirow{3}{*}{7} & CN     & 49 (M/F: 26/23)          & 74.7$\pm$4.2  & 29.1$\pm$1.3  & 5.1$\pm$2.9   & 8.2$\pm$4.2   & 0.2$\pm$0.4 \\
                    & MCI    & 52 (M/F: 34/18)          & 74.5$\pm$6.3  & 27.7$\pm$1.8  & 9.8$\pm$4.5   & 15.8$\pm$7.1  & 1.4$\pm$1.1 \\
                    & AD     & 30 (M/F: 15/15)          & 71.4$\pm$6.4  & 18.2$\pm$5.7  & 28.4$\pm$12.4 & 40.4$\pm$13.8 & 8.8$\pm$3.8 \\\midrule
\multirow{3}{*}{8} & CN     & 34 (M/F: 16/18)          & 75.1$\pm$4.7  & 28.9$\pm$1.2  & 5.6$\pm$2.7   & 9.3$\pm$3.9   & 0.2$\pm$0.4 \\
                    & MCI    & 34 (M/F: 24/10)          & 74.1$\pm$6.5  & 27.6$\pm$2.5  & 10.3$\pm$4.8  & 16.2$\pm$6.9  & 1.5$\pm$1.0 \\
                    & AD     & 28 (M/F: 17/11)          & 73.0$\pm$6.9  & 19.2$\pm$6.6  & 25.5$\pm$12.4 & 37.4$\pm$15.0 & 7.3$\pm$3.8 \\\midrule
\multirow{3}{*}{9}& CN     & 22 (M/F: 13/9)          & 73.5 $\pm$3.6 & 29.2$\pm$1.3  & 5.1$\pm$2.4   & 7.8$\pm$3.5   & 0.1$\pm$0.3 \\
                    & MCI    & 25 (M/F: 15/10)          & 73.4$\pm$5.8  & 28.2$\pm$2.1  & 8.9$\pm$3.5   & 14.1$\pm$5.6  & 1.2$\pm$1.2 \\
                    & AD     & 23 (M/F: 14/9)          & 74.1$\pm$6.1  & 19.4$\pm$5.0  & 22.6$\pm$8.3  & 33.8$\pm$10.5 & 6.7$\pm$4.1 \\\midrule
\multirow{3}{*}{10}& CN     & 13 (M/F: 4/9)          & 74.9$\pm$4.0  & 28.8$\pm$1.2  & 5.5$\pm$3.0   & 9.4$\pm$4.6   & 0.1$\pm$0.2 \\
                    & MCI    & 19 (M/F: 14/5)           & 75.7$\pm$5.8  & 27.3$\pm$2.1  & 11.1$\pm$5.3  & 17.1$\pm$7.8  & 1.8$\pm$1.5 \\ 
                    & AD     & 10 (M/F: 6/4)            & 72.5$\pm$4.6  & 20.9$\pm$7.0  & 26.0$\pm$12.6 & 36.9$\pm$15.5 & 7.2$\pm$2.8
\\\bottomrule
\end{tabular}
\begin{tablenotes}\scriptsize
    \item $\star$ Age, MMSE, CDR-SB, ADAS-cog11, and ADAS-cog13 are listed as (Mean $\pm$ Standard deviation).
    \item $\star$ Abbreviation: AD, Alzheimer's disease; ADAS-cog, Alzheimer's Disease Assessment Scale-cognitive subscale; CDR-SOB, Clinical Dementia Rating scale-Sum of the Boxes; CN, Cognitively Normal; F, Female; M, Male; MCI, Mild Cognitive Impairment; MMSE, Mini-Mental State Examination.
\end{tablenotes}
\end{threeparttable}}}
\end{table}

\clearpage 

\section{Performance across Time}
\begin{table}[h]
\caption{Longitudinal classification and regression results for each year in each of the specific volumes for the brain and the cognitive test scores. We observed that the data imbalance was caused over time but the performance did not change significantly. Additionally, the MAE of cognitive test values decreased and the accuracy of classification increased over time. Lastly, we observed the patterns between cognitive score decline and clinical status prediction accuracy.}
\label{tb:longitudinal_performance_time_points_cog}
\centering\scriptsize{\begin{threeparttable}
\renewcommand*{\arraystretch}{0.65}
\begin{tabular}{ccccccccccc}
\toprule
Year & 0 & 1 & 2 & 3 & 4 & 5 & 6 & 7 & 8 & 9\\\midrule
\multicolumn{5}{l}{\textbf{\textit{$\#$ of subjects}}}&&&&&&\\
CN & 226 & 208 & 203 & 126 & 103 & 68 & 64 & 49 & 34 & 22\\
MCI& 334 & 275 & 200 & 156 & 99  & 67 & 56 & 52 & 34 & 25 \\
AD & 131 & 177 & 178 & 95  & 70  & 47 & 42 & 30 & 28 & 23 \\\midrule
\multicolumn{5}{l}{\textbf{\textit{MRI Biomarker Regression (MAE)}}}&&&&&&\\
Ventricles            & 1.972 &1.952&	2.257 &2.427 &2.906 &2.874 &2.282 &2.566 &3.590 &2.961\\
Hippocampus           & 0.160 &0.170&	0.159 &0.182 &0.178 &0.177 &0.236 &0.259 &0.154 &0.325\\
Whole Brain           & 11.489 &12.656&	11.437 &12.987 &17.796 &17.238 &21.905 &21.995 &8.815 &39.245\\
Entorhinal Cortex     & 0.205 &0.196&	0.208 &0.223 &0.247 &0.224 &0.254 &0.180 &0.081 &0.622\\
Fusiform Gyrus        & 0.477 &0.477&	0.495 &0.509 &0.630 &0.560 &0.470 &0.444 &0.559 &0.399\\
Middle Temporal Gyrus & 0.452 &0.465&	0.511 &0.519 &0.571 &0.577 &0.569 &0.537 &0.354 &0.649\\\midrule
\multicolumn{5}{l}{\textbf{\textit{Cognitive Score Regression (RMSE)}}}&&&&&&\\
MMSE        & 2.421 &2.409&	2.483 &2.411 &2.640 &2.793 &2.344 &2.303 &2.546 &2.044\\
ADAS-cog11        & 4.676 &4.535&	4.306 &5.065 &4.989 &5.415 &4.488
&4.384 &4.007 &4.328\\
ADAS-cog13 & 6.108 & 5.700 & 5.435 &6.353 &6.532 &6.412 &5.816 &5.269 &5.567 &5.357\\\midrule
\multicolumn{5}{l}{\textbf{\textit{AD vs. MCI vs. CN}}}&&&&&&\\
mAUC      & 0.846 & 0.859 & 0.868 & 0.886 & 0.886 & 0.872 & 0.891 & 0.884 & 0.844 & 0.853\\
Precision & 0.652 & 0.678 & 0.706 & 0.735 & 0.724 & 0.722 & 0.757 & 0.708 & 0.694 & 0.677\\
Recall    & 0.689 & 0.695 & 0.693 & 0.750 & 0.792 & 0.754 & 0.736 & 0.710 & 0.639 & 0.739\\
\bottomrule
\end{tabular}
\end{threeparttable}}
\end{table}

\begin{figure}[h]
\centering
    \includegraphics[width=.84\textwidth]{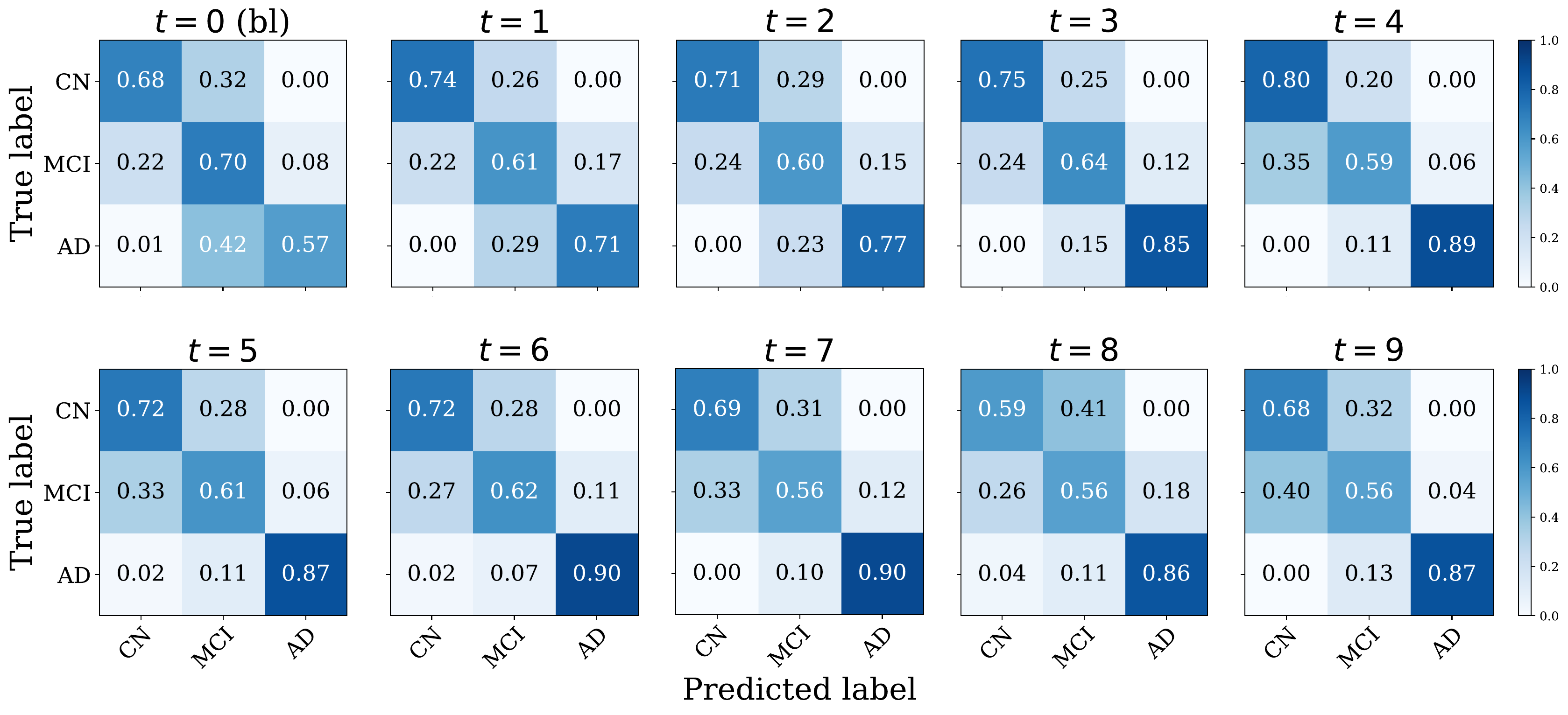}
    \caption{Yearly computed multi-class classification confusion matrices with normalization using our proposed method. The diagonal elements are the number of points where the predicted labels were the same as the real labels, excluding missing labels, whereas the non-diagonal elements were misclassified by our proposed method. The higher the diagonal value and the darker the shade of blue, the more accurate is the diagnosis of the clinical status.}
    \label{confusion_matrix_cog}
\end{figure}

\clearpage
\section{Interpreting Cell States}

\begin{figure}[h]
\centering
\begin{subfigure}[]{0.47\textwidth}
\centering
    \includegraphics[width=\textwidth]{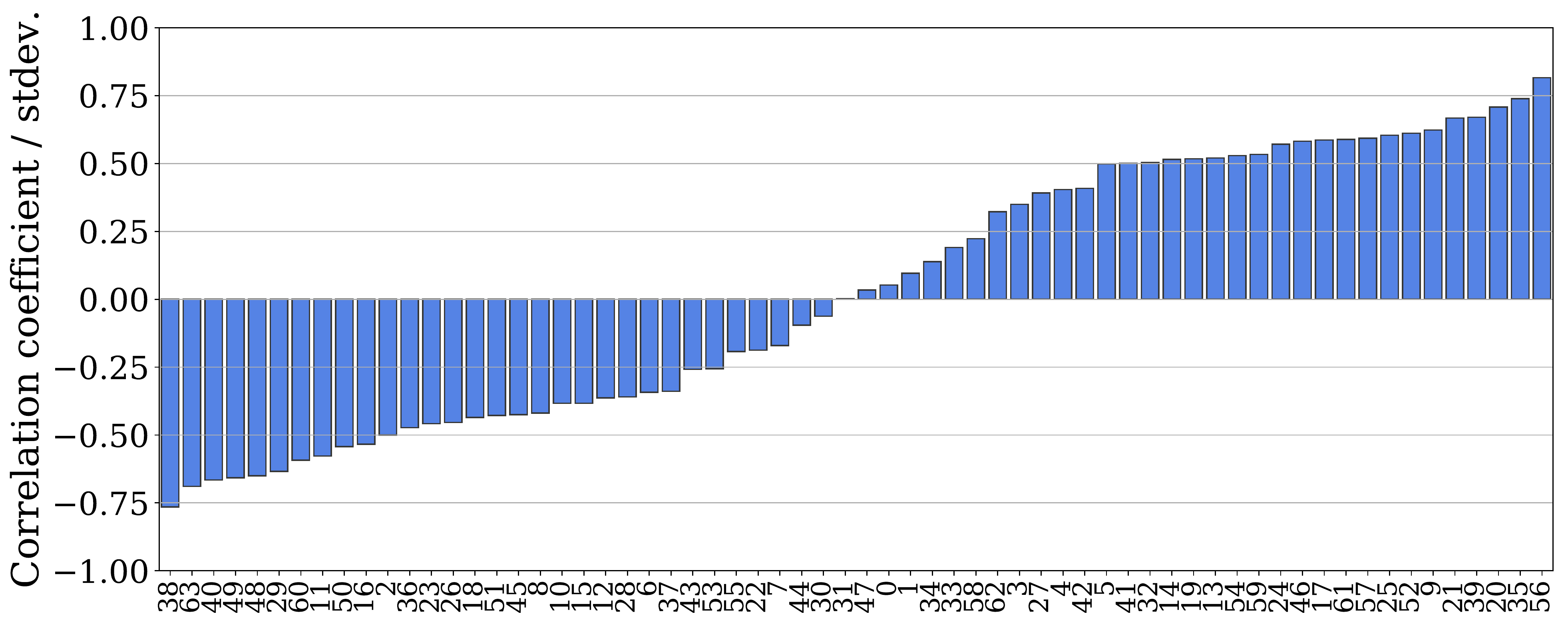}
\end{subfigure}
\begin{subfigure}[]{0.47\textwidth}
\centering
    \includegraphics[width=\textwidth]{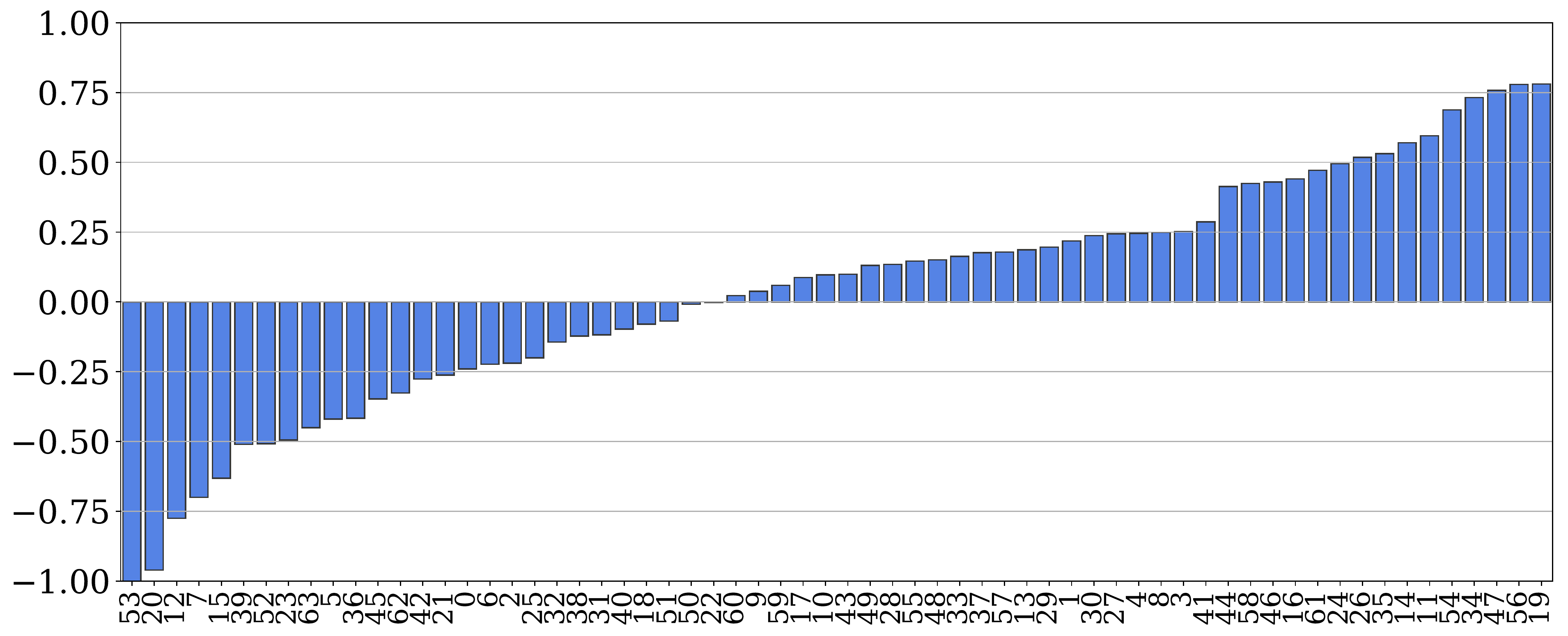}
\end{subfigure}
\begin{subfigure}[]{0.47\textwidth}
\centering
    \includegraphics[width=\textwidth]{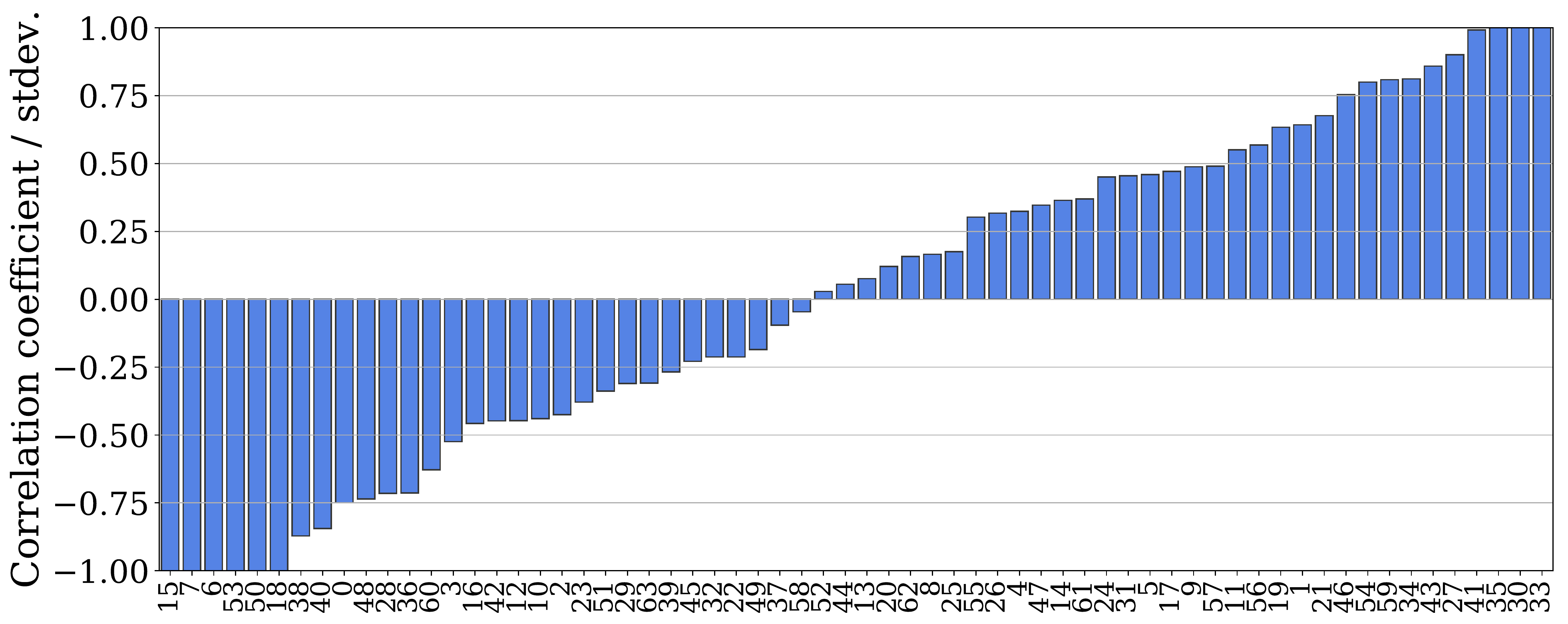}
\end{subfigure}
\begin{subfigure}[]{0.47\textwidth}
\centering
    \includegraphics[width=\textwidth]{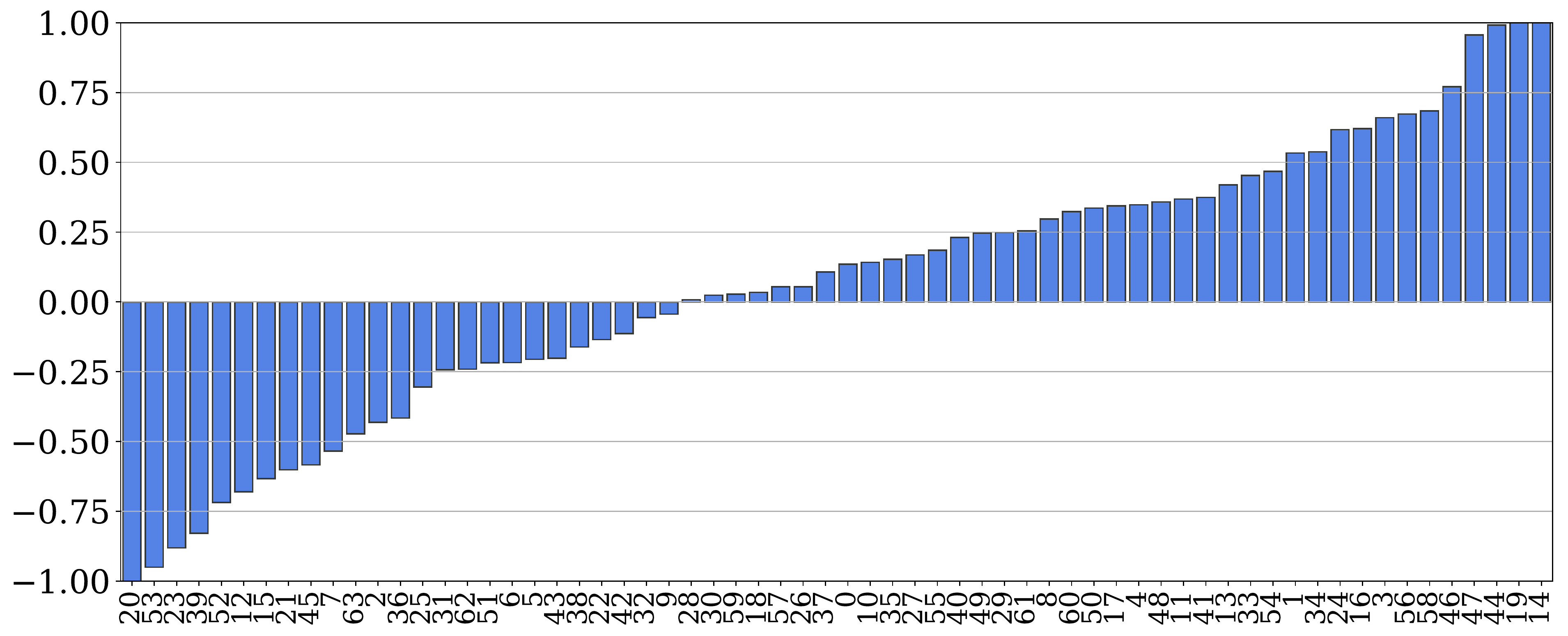}
\end{subfigure}
\begin{subfigure}[]{0.47\textwidth}
\centering
    \includegraphics[width=\textwidth]{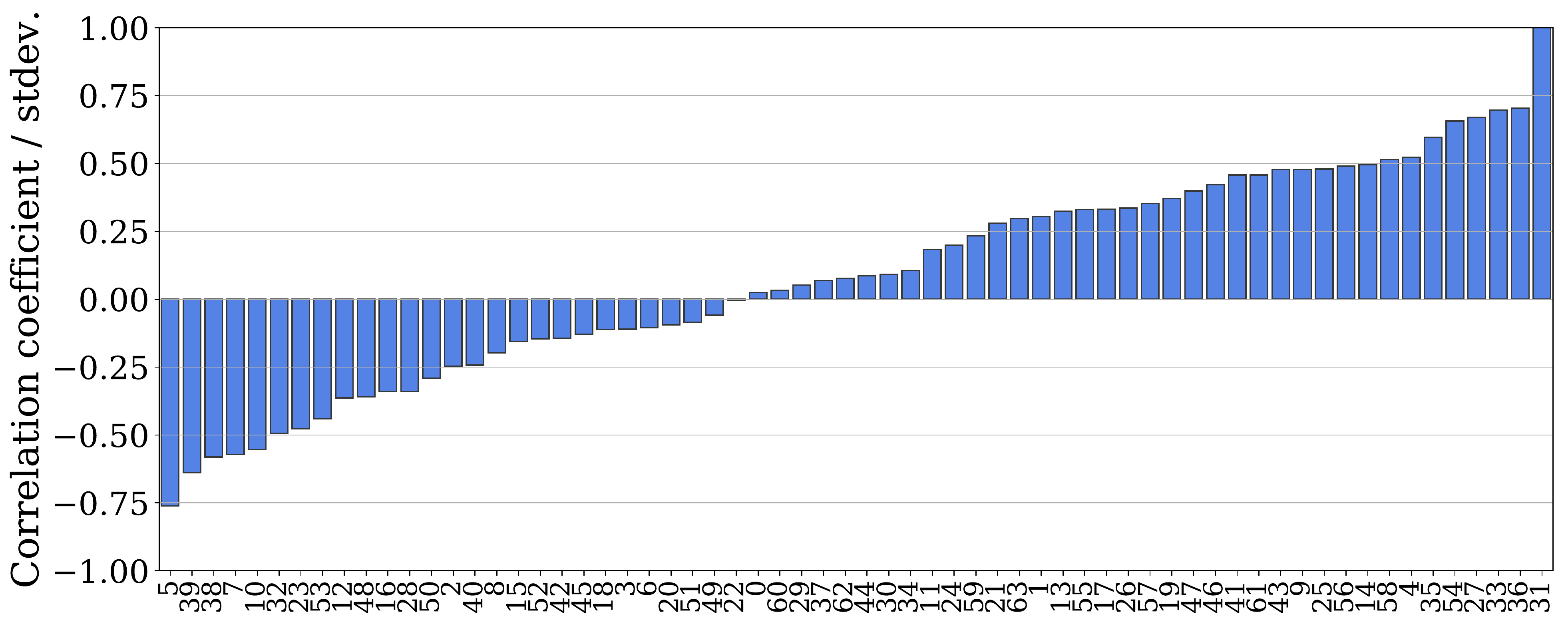}
\end{subfigure}
\begin{subfigure}[]{0.47\textwidth}
\centering
    \includegraphics[width=\textwidth]{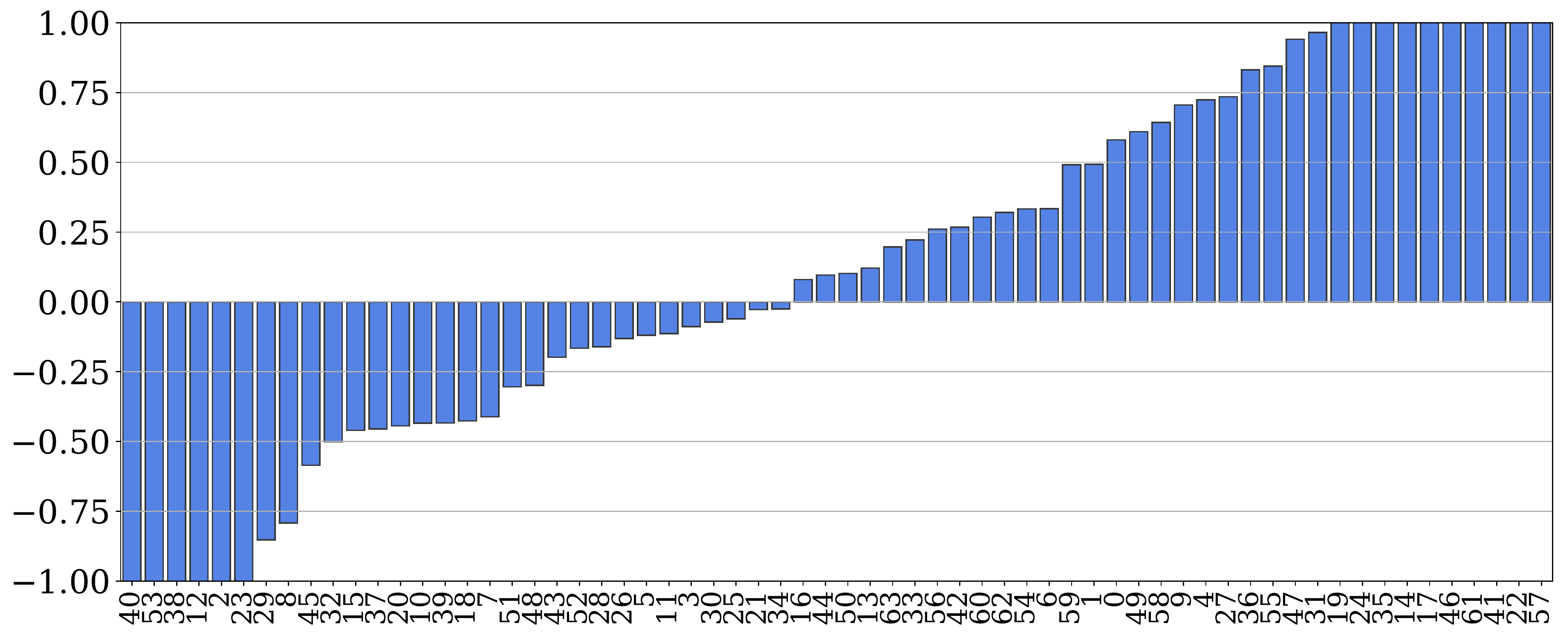}
\end{subfigure}
\begin{subfigure}[]{0.47\textwidth}
\centering
    \includegraphics[width=\textwidth]{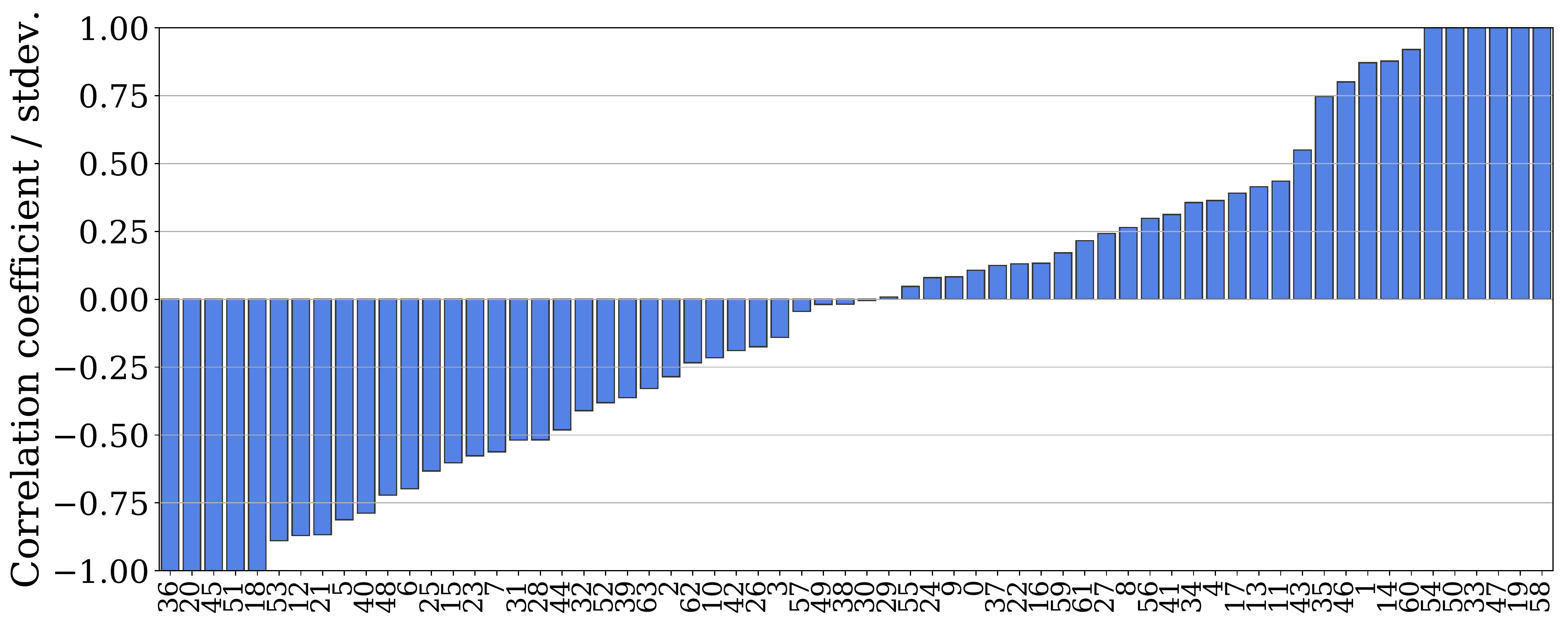}
\end{subfigure}
\begin{subfigure}[]{0.47\textwidth}
\centering
    \includegraphics[width=\textwidth]{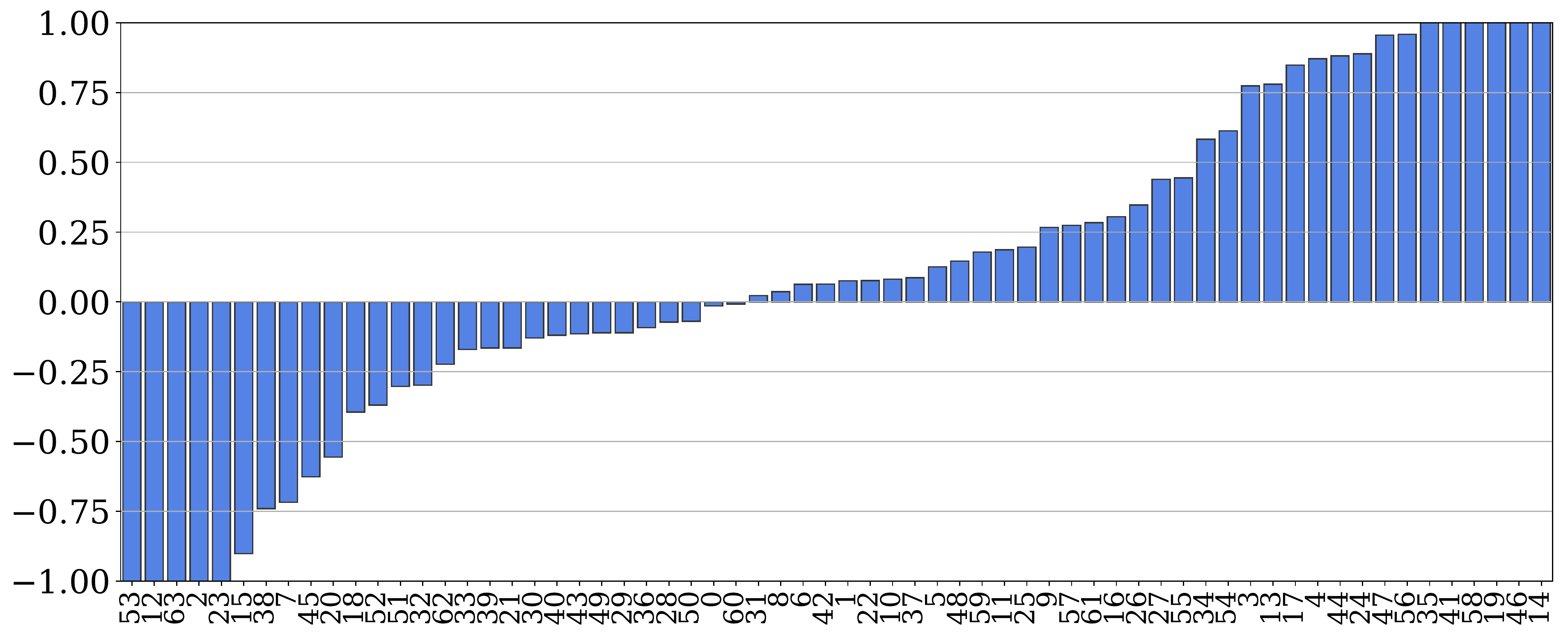}
\end{subfigure}
\begin{subfigure}[]{0.47\textwidth}
\centering
    \includegraphics[width=\textwidth]{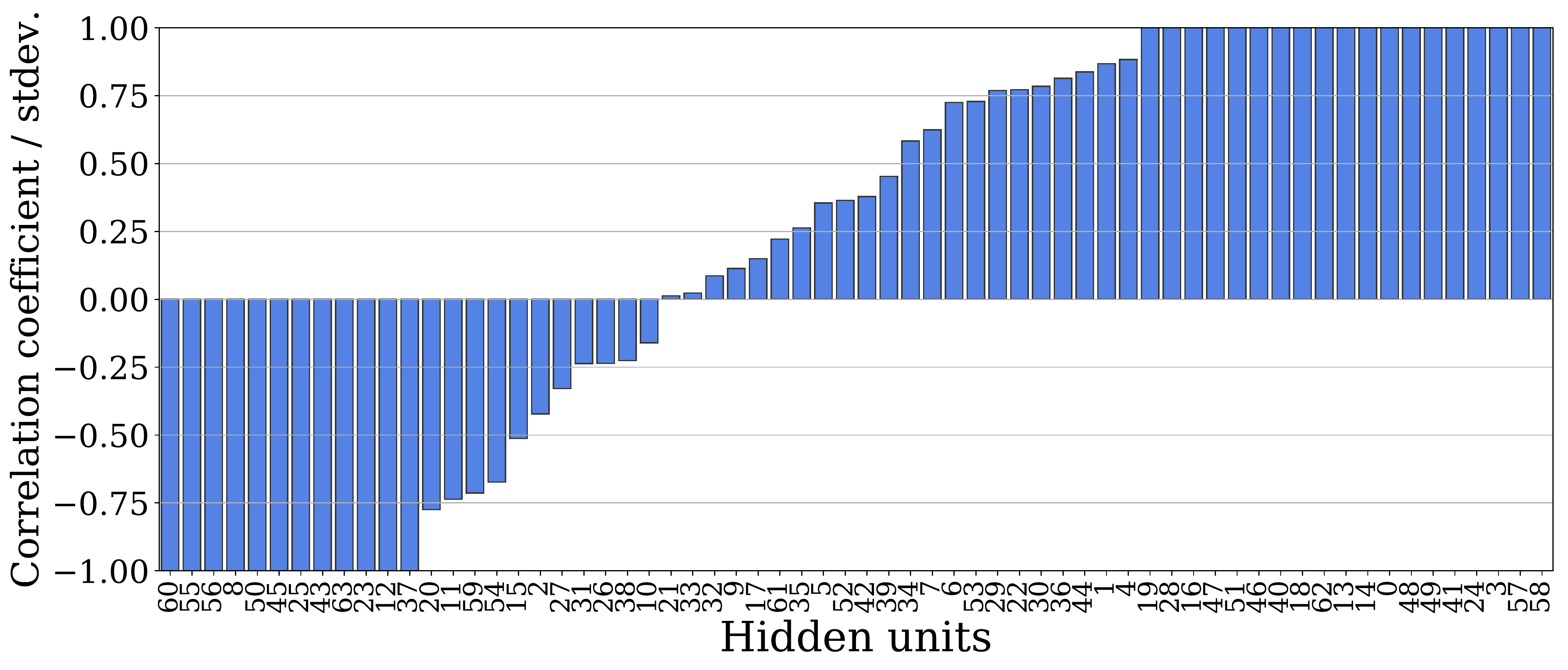}
    \caption{CN-MCI}
\end{subfigure}
\begin{subfigure}[]{0.47\textwidth}
\centering
    \includegraphics[width=\textwidth]{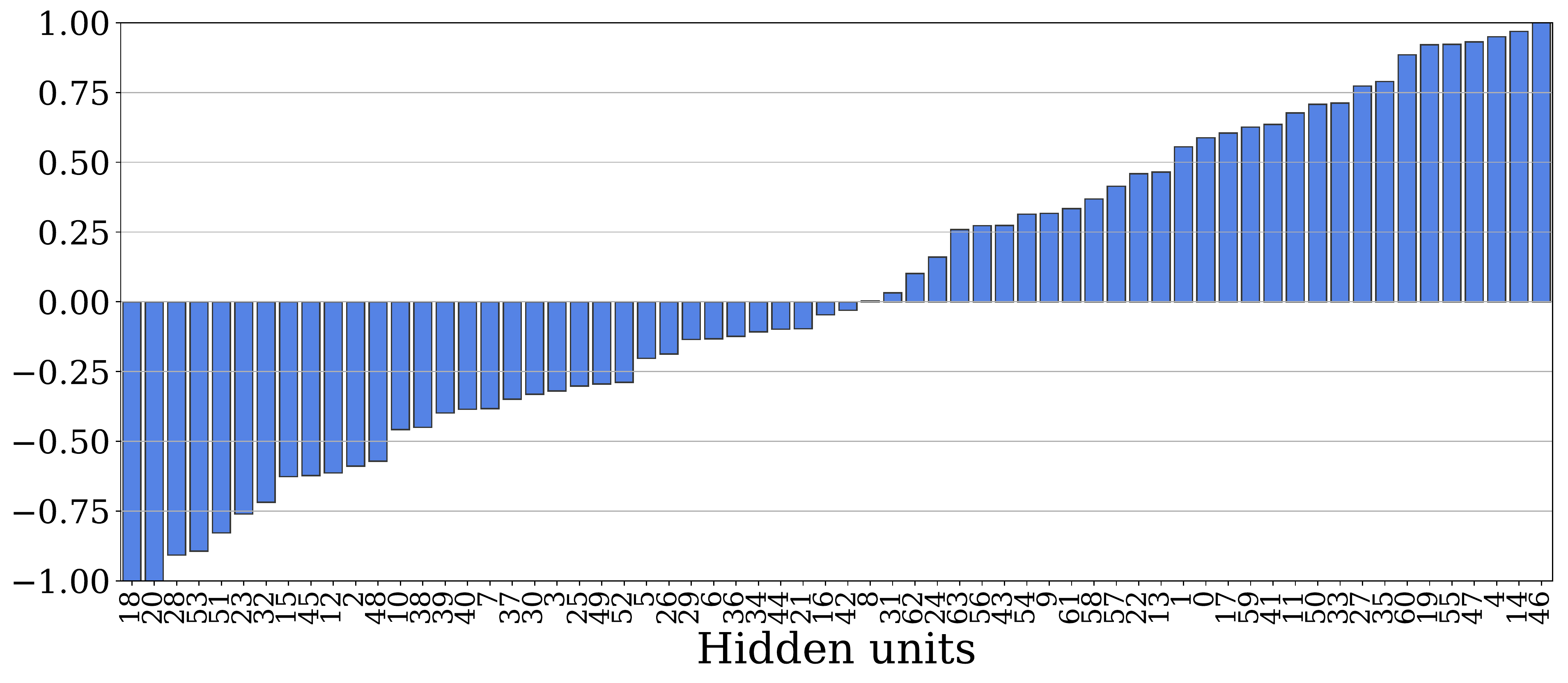}
    \caption{MCI-AD}
\end{subfigure}
\caption{Normalized point biserial correlation coefficients for subjects with conversion from CN to MCI (left) and from MCI to AD (right) during a period of 10 years from the baseline.}
\label{fig:cell_state_group}
\end{figure}
\end{document}